\documentclass{article}
\usepackage{appendix}
\usepackage{graphicx,float}
\usepackage{amsmath,mathpazo,mathtools}
\usepackage{xcolor}
\usepackage{multirow}
\usepackage{biblatex}
\usepackage{geometry}[margin=1in]
\usepackage{hyperref}
\usepackage{authblk}

\hypersetup{
    colorlinks,
    citecolor=black,
    filecolor=black,
    linkcolor=black,
    urlcolor=black
}

\usepackage{amssymb}
\usepackage{listings}

\title{What do physics-informed DeepONets learn? Understanding and improving training for scientific computing applications}

\author[1,2]{Emily Williams}
\author[1]{Amanda Howard}
\author[3]{Brek Meuris}
\author[1]{Panos Stinis}

\affil[1]{Advanced Computing, Mathematics and Data Division, Pacific Northwest National Laboratory, WA 99354}
\affil[2]{Massachusetts Institute of Technology, MA 02139}
\affil[3]{Sandia National Laboratories, NM 87123}

\begin{document}

\maketitle

\begin{abstract}
    Physics-informed deep operator networks (DeepONets) have emerged as a promising approach toward numerically approximating the solution of partial differential equations (PDEs). In this work, we aim to develop further understanding of what is being learned by physics-informed DeepONets by assessing the universality of the extracted basis functions and demonstrating their potential toward model reduction with spectral methods. Results provide clarity about measuring the performance of a physics-informed DeepONet through the decays of singular values and expansion coefficients. In addition, we propose a transfer learning approach for improving training for physics-informed DeepONets between parameters of the same PDE as well as across different, but related, PDEs where these models struggle to train well. This approach results in significant error reduction and  learned basis functions that are more effective in representing the solution of a PDE.
\end{abstract}

\section{Introduction}

In recent years, physics-informed deep learning has emerged as a viable approach for the numerical solution of partial differential equations (PDEs) \cite{meuris_machine-learning-based_2023, karniadakis_physics-informed_2021, alber_integrating_2019}. The deep operator network (DeepONet) consists of a deep neural network (DNN) for encoding the discrete input function space (branch net) and another DNN for encoding the domain of the output functions (trunk net) \cite{lu_learning_2021}. Physics-informed DeepONets leverage automatic differentiation to impose the underlying physical laws during model training \cite{wang_learning_2021}. The use of the neural tangent kernel (NTK) to assign weights dynamically for the terms in the loss function used to train the DeepONet has demonstrated improved predictions compared with using fixed predetermined weights \cite{qadeer_efficient_2024, wang_improved_2022}. However, further investigation is warranted toward understanding and improving training for physics-informed DeepONets \cite{wang_improved_2022}.

Further, we want to investigate the benefits in exploiting physics-informed DeepONets toward improving traditional scientific computing methods. DeepONets can be used to identify custom-made candidate basis functions on which to expand the solution of PDEs for evolution using a spectral approach \cite{meuris_machine-learning-based_2023}. In this work, we assess the applicability of the basis functions extracted from data-driven and physics-informed DeepONets. Then, we propose a transfer learning approach to initializing physics-informed models for PDEs where these models fail to train well.

\section{Technical Approach}

\subsection{Data-driven and physics-informed DeepONets}

Consider a time-dependent PDE
\begin{align}
    s_t + \mathcal{N}[s] = 0
\end{align}
with appropriate boundary conditions and initial condition $s(x,0) = u(x)$. The goal is to learn the continuous, possibly nonlinear, solution operator $\mathcal{G} : u(x) \mapsto s(x,t)$. The universal approximation theorem guarantees the existence of a pair of 2-layer neural networks such that the inner product of their outputs can approximate the action of a continuous nonlinear operator to arbitrary accuracy \cite{lu_learning_2021, chen_universal_1995}. 
\begin{align}
    \mathcal{G}(u)(y) \approx \sum_{k=1}^w \underbrace{b_k(u(x_1),u(x_2),\dots,u(x_m))}_{\textrm{branch}} \underbrace{\gamma_k(y)}_{\textrm{trunk}}
\end{align}
A DeepONet $\mathcal{G}_{NN}(u)(y)$ is a deep neural architecture designed to approximate $\mathcal{G}$ evaluated at $y = (x,t)$. It takes as inputs a discrete representation $u = (u(x_j))_{1 \leq j \leq m}$, where $x_1, \dots, x_m$ are pre-selected sensor points, and an output location $y = (x,t)$ in the spatiotemporal domain. The DeepONet consists of deep branch and trunk networks $\{b_k\}_{1 \leq k \leq w}$ and $\{\gamma_k\}_{1 \leq k \leq w}$ merged together in a dot product layer 
\begin{align}
    \mathcal{G}_{NN}^\theta (u)(y) = \sum_{k=1}^w b_k(u) \gamma_k(y)
\end{align}
where $\theta$ denotes the trainable parameters. Given $N$ input-output function pairs $\{(u^{(j)},s^{(j)})\}_{1 \leq j \leq N}$, where $s^{(j)} = \mathcal{G}(u^{(j)})$, and $P$ corresponding evaluation points $\{{y}_i^{(j)}\}_{1 \leq i \leq P, 1 \leq j \leq N}$, this architecture is trained by minimizing the mean-squared error as the loss function
\begin{align}
    \mathcal{L}(\theta) = \frac{1}{N P} \sum_{j=1}^{N} \sum_{i=1}^{P} \left( s^{(j)} \left({y}_i^{(j)}\right) - \mathcal{G}_{NN}^\theta \left(u^{(j)}\right)\left({y}_i^{(j)}\right) \right)^2
\end{align}
For each input function $u$, the same number and location of the sensor points is required. There are no constraints on the number or locations for the evaluation of the output functions.

\begin{figure}[!htb]
    \centering
    \includegraphics[width=0.49\textwidth]{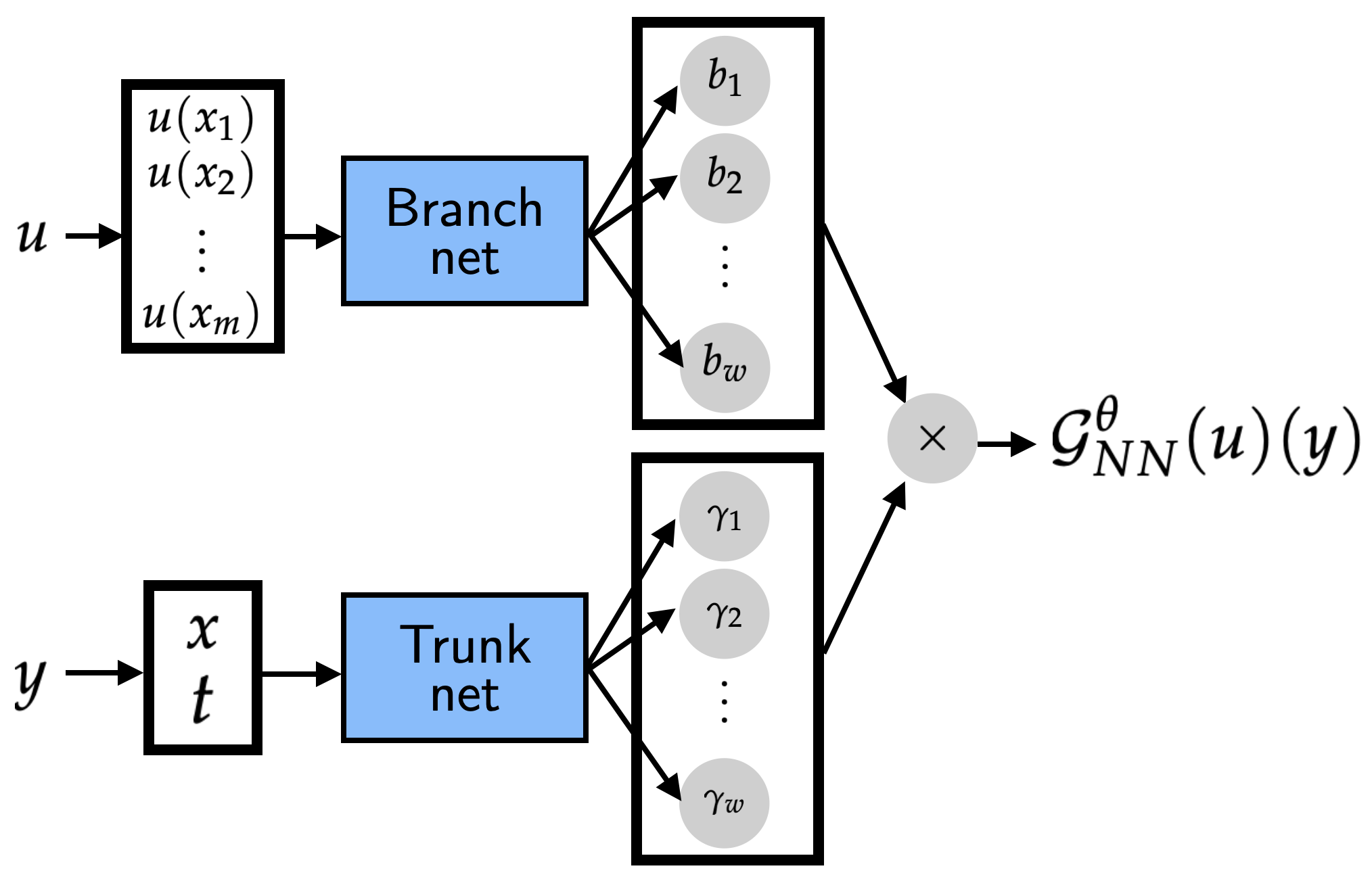}
    \hfill
    \includegraphics[width=0.49\textwidth]{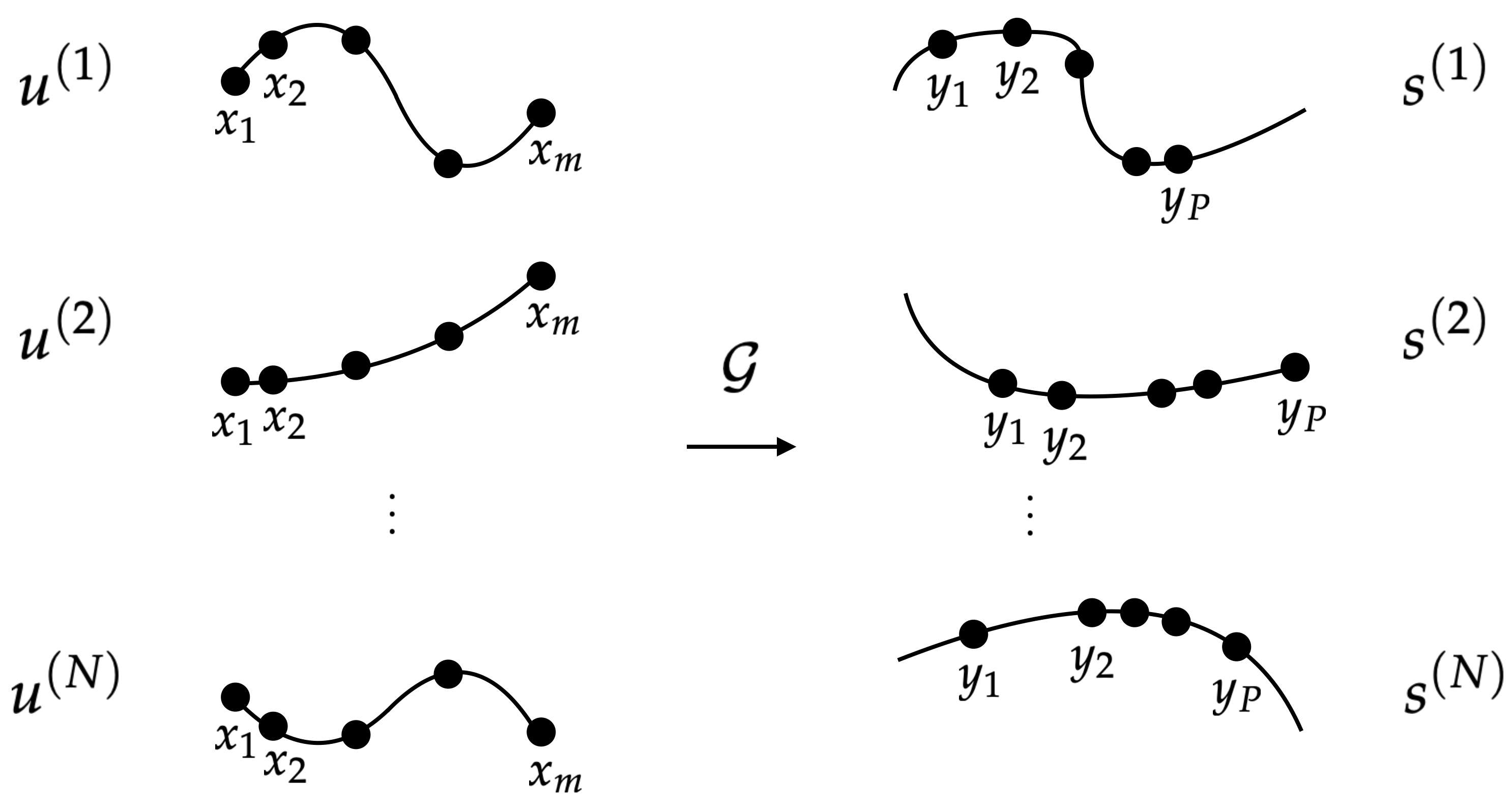}
    \caption{DeepONet architecture and training (adapted from \cite{lu_learning_2021}).}
    \label{fig:training}
\end{figure}

Physics-informed DeepONets recognize that the outputs of a DeepONet model are differentiable with respect to their input coordinates. This enables the use of automatic differentiation to formulate an appropriate regularization mechanism for biasing the target output functions to satisfy the underlying physical constraints. Then, we have a procedure for training physics-informed DeepONet models without requiring any training data for the latent output functions, except for the appropriate initial and boundary conditions of a given PDE system \cite{wang_learning_2021}.

\begin{figure}[!htb]
    \centering
    \includegraphics[width=\textwidth]{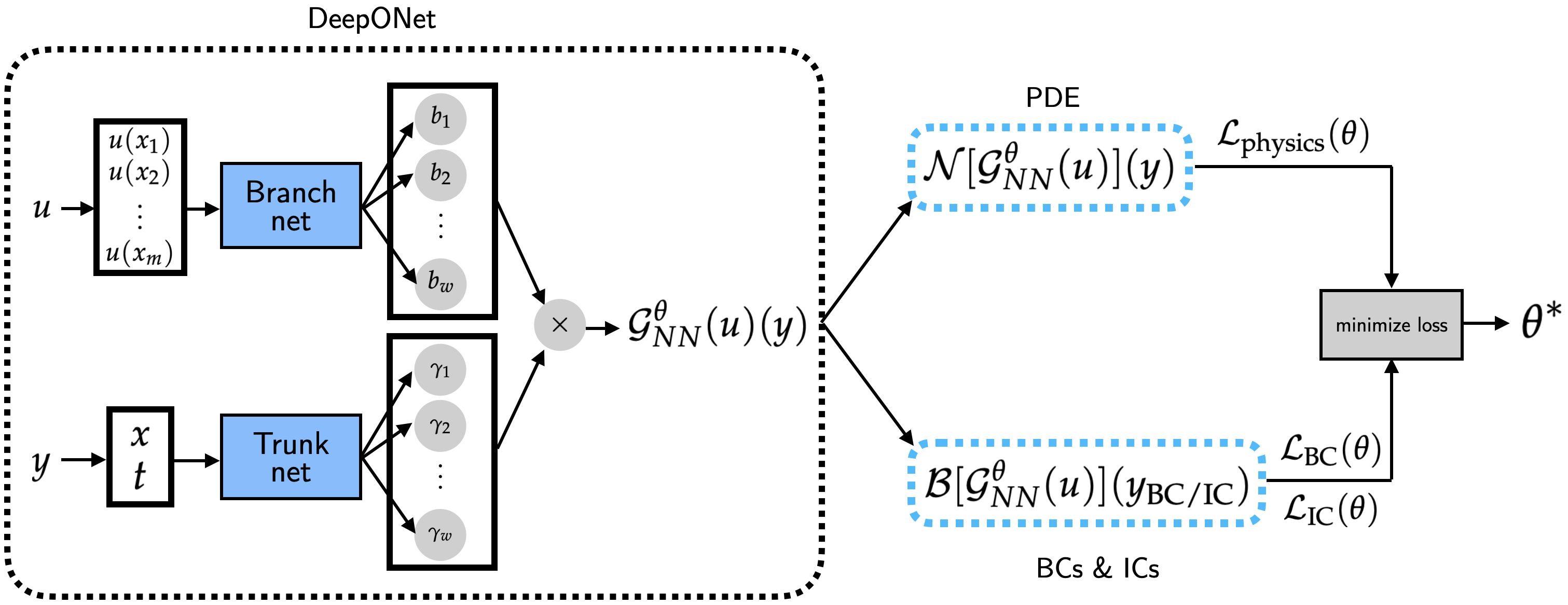}
    \caption{Physics-informed DeepONet (adapted from \cite{wang_learning_2021}).}
    \label{fig:pi-arch}
\end{figure}

\subsection{Transfer learning for initialization}

Transfer learning refers to exploiting what has been learned in one problem to improve generalization in a different, but related, problem \cite{goodfellow_deep_2016, bozinovski_influence_1976, do_transfer_2005}. Transfer learning has been shown to decrease the training time for a DNN while also resulting in lower generalization error for tasks in computer vision and natural language processing, for example, that require lengthy training times with large amounts of training data \cite{shin_deep_2016, conneau_supervised_2018, liu_exploring_2019, iman_review_2023, venkatasubbu_harnessing_2023}. Initialization of DNNs can have a substantial impact on convergence during training \cite{glorot_understanding_2010}. Initializing a new model with weights from a previously trained model is called fine-tuning and has been shown to result in improved performance compared to random initialization \cite{yosinski_how_2014, zhang2024deeponetmultioperatorextrapolationmodel, xu2024multi}. Transfer learning has been increasingly used in operator learning models. Recent work has included fine-tuning DeepONets to result in lower training times for transferring and evolving knowledge \cite{xu_transfer_2023, goswami_deep_2022, wang_long-time_2021}. The capability of DeepONets have been extended to train for multiple tasks on multiple domains and geometries with the aid of transfer learning \cite{kumar_synergistic_2024, kahana_geometry_2023, hu_operator_2024}. However, the choice of initialization in particular and the tasks that benefit from transfer initialization by weights is still not well understood, especially for physics-informed problems \cite{goodfellow_deep_2016, wu_fine-tuning_2024, goswami_deep_2022}. We use transfer learning for the initialization of the weights and biases for physics-informed DeepONets from previously trained models. This is demonstrated for the same PDE across different parameters, as well as across different, but related, PDEs.

\subsection{Neural tangent kernel}

The neural tangent kernel (NTK) is a framework for better convergence and generalization guarantees for DNNs
\cite{jacot_neural_2018, du_gradient_2019, allen-zhu_convergence_2019}. For physics-informed DeepONets, the NTK is used to assign weights dynamically for each term in the loss function (e.g., ICs/BCs, PDE) for each evaluation point and for each training iteration \cite{wang_improved_2022, qadeer_efficient_2024}. As above and following \cite{wang_improved_2022}, a physics-informed DeepONet is trained by minimizing
\begin{align}
    \mathcal{L}(\theta) &= \mathcal{L}_{\mathrm{BC/IC}} (\theta) + \mathcal{L}_{\mathrm{PDE}}(\theta) = \frac{2}{N^*} \sum_{k=1}^{N^*} \left(\mathcal{T}^{(k)} \left(u^{(k)}(x_k), \mathcal{G}_{NN}^\theta \left(u^{(k)}\right)(y_k)\right)\right)^2
\end{align}
where $N^* = 2 N P$. For every $k$, $\mathcal{T}^{(k)} (\cdot,\cdot)$ can be the identity operator, differential operator $\mathcal{N}$, or boundary condition $\mathcal{B}$. This form of the loss function is ``fully-decoupled'' \cite{wang_improved_2022}. Given this loss function, the NTK is a matrix $\mathbf{H}(\theta) \in \mathbb{R}^{N^* \times N^*}$ with entries
\begin{align}
    \mathbf{H}_{ij} (\theta) = \left\langle \frac{\mathcal{T}^{(i)} \left(u^{(i)}(x_i), \mathcal{G}_{NN}^\theta \left(u^{(i)}\right)(y_i)\right)}{d\theta}, \frac{\mathcal{T}^{(j)} \left(u^{(j)}(x_j), \mathcal{G}_{NN}^\theta \left(u^{(j)}\right)(y_j)\right)}{d\theta} \right\rangle
\end{align}
for $i,j = 1, 2, \dots, N^*$. For NTK-guided weights, consider the weighted loss function
\begin{align}
    \mathcal{L}(\theta) = \frac{2}{N^*} \sum_{k=1}^{N^*} \lambda_k \left(\mathcal{T}^{(k)} \left(u^{(k)}(x_k), \mathcal{G}_{NN}^\theta \left(u^{(k)}\right)(y_k)\right)\right)^2
\end{align}
All weights $\{\lambda_k\}_{k=1}^{N^*}$ are initialized to 1 and updated during training by
\begin{align}
    \lambda_k = \left( \frac{||\mathbf{H}(\theta)||_\infty}{\mathbf{H}_{kk}(\theta_n)} \right)^\alpha
\end{align}
where $\theta_n$ denotes the DeepONet parameters at step $n$. Here, $\alpha$ is a hyper-parameter that determines the magnitude of each weight. From \cite{wang_improved_2022}, $\alpha = 1$ refers to local NTK weights while $\alpha = 1/2$ refers to moderate local NTK weights. Using NTK-guided weights for the loss function was shown to result in higher prediction accuracy for the physics-informed DeepONet when compared to fixed predetermined weights.

A disadvantage of the NTK is that it can greatly increase the computational time for training a physics-informed DeepONet, and in some cases the NTK is intractable due to having to propogate through a large network. The conjugate kernel (CK) is the kernel induced by the Jacobian only with respect to the parameters $\theta$ in the last layer \cite{qadeer_efficient_2024, fan_spectra_2020}. The CK can be thought of as a zeroth-order approximation to the NTK \cite{qadeer_efficient_2024}. Using CK-guided weights has been shown to result in increased efficiency and accuracy for training physics-informed DeepONets for many applications, although the NTK can have better convergence for some PDEs \cite{howard2024conjugate}.

\subsection{Basis functions for spectral methods}

The custom basis functions are extracted from the trunk net function space using a singular value decomposition (SVD) based method. Following \cite{meuris_machine-learning-based_2023}, we denote the collection of ``frozen-in-time'' trunk net functions by $\{\tau_k\}_{1 \leq k \leq p}$, e.g., by evaluating the trunk net functions $\{\gamma_k\}$ at $t=0$ (so that $p = w$, where $w$ is the width of the trunk net output used in the DeepONet representation).

Denote by $\langle{\cdot,\cdot}\rangle$ the $L^2$ inner product on a spatial domain $\Omega$ and let $\{(x_i,\omega_i)\}_{1 \leq i \leq M}$ be a quadrature rule on $\Omega$ so that $\langle{h_1,h_2}\rangle \approx  \sum_{i = 1}^M \overline{h_1(x_i)}h_2(x_i)\omega_i$. The eigenfunctions $\{\phi_k\}_{1 \leq k \leq p}$ of the covariance operator
\begin{align}
\mathcal{C} = \sum_{k = 1}^p \tau_k \otimes \tau_k = \sum_{k = 1}^p \tau_k\langle{\tau_k,\cdot}\rangle,
\end{align}
ordered by decreasing eigenvalues, form an orthonormal basis for $\mathcal{S} = \text{span}(\{\tau_k\}_{1 \leq k \leq p})$. Discretizing $\mathcal{C}$ and performing its eigendecomposition to compute the basis functions is computationally infeasible. Instead, define the $M \times p$ matrix $B$ by $B_{ik} = \omega_i^{1/2}\tau_k(x_i)$ and perform its SVD $B = QSV^*$. Letting $W = \text{diag}(\omega_1, \dots, \omega_M)$, the entries of $W^{-1/2} Q$ provide the values of $\{\phi_k\}$ at the quadrature points via
\begin{align}
\phi_k(x_i) = (W^{-1/2}Q)_{ik} \ \text{ for } 1\leq i \leq M \text{ and } 1 \leq k \leq p
\end{align}

The functional forms of the basis functions need to be recovered to allow for their usage in a spectral method, such as through an orthogonal polynomial expansion. For any $\widetilde{M} < M$, let $\{L_j\}_{0\leq j \leq \widetilde{M}}$ be the orthonormal Legendre polynomials on $\Omega$ and define the functions $\{\tilde \phi_k\}_{1 \leq k \leq p}$ by
\begin{align}
\tilde \phi_k = \sum_{j = 0}^{\widetilde{M}} \left(\sum_{i = 1}^M L_j(x_i)\phi_k(x_i) \omega_i\right) L_j
\end{align}

This procedure proposed by \cite{meuris_machine-learning-based_2023} only uses well-conditioned operations to obtain a projection enabling the evaluation of the basis functions. By choosing a sufficiently large $\widetilde{M}$, the $\{\tilde \phi_k\}$ serve as good approximations to $\{\phi_k\}$. The singular values $S = \text{diag}(\sigma_1,\dots,\sigma_p)$ signify the contribution of each basis function to $\mathcal{S}$. Only using basis functions corresponding to singular values greater than a specified cutoff, typically $10^{-13}$, leads to more robust solutions, since the noisy functions are weeded out, for cheaper computational cost in a spectral approach \cite{meuris_machine-learning-based_2023}. Consider the weak formulation for a time-dependent PDE
\begin{align}
    \langle \tilde{\phi}_k, s_t + \mathcal{N}[s] \rangle = 0
\end{align}
for $1 \leq k \leq p$. Express the solution as a linear combination
\begin{align}
    \hat{s} = \sum_{i=1}^p a_i (t) \tilde{\phi}_i (x)
\end{align}
Substituting this approximation into the weak formulation results in a system of ordinary differential equations (ODEs) for the expansion coefficients
\begin{align}
    a'_k(t) &= -\left\langle \tilde{\phi}_k, \mathcal{N}\left[\sum_{i=1}^p a_i(t) \tilde{\phi}_i \right] \right\rangle \\
    a_k (0) &= \langle \tilde{\phi}_k, u \rangle = \sum_{j=1}^M \tilde{\phi}_k (x_j) u (x_j) \omega_j
\end{align}

\section{Results}

\subsection{Advection-diffusion}

We first demonstrate the use of physics-informed basis functions in a spectral method, comparing to the effectiveness of the data-driven custom basis functions \cite{meuris_machine-learning-based_2023}. Consider the advection-diffusion equation 
\begin{align}
    \frac{\partial s}{\partial t} + \alpha \frac{\partial s}{\partial x} - \nu \frac{\partial^2 s}{\partial x^2} = 0
\end{align}
on $(x,t) \in (0,2\pi) \times (0,1)$ with periodic boundary conditions.  The advection coefficient is $\alpha = 4$ and the diffusion coefficient is $\nu = 0.01$. The goal is to learn the solution operator $\mathcal{G} : u(x) \mapsto s(x,t)$, where the input to the branch network is the initial condition $s(x,0) = u(x) \sim $ Gaussian random field (GRF). 

\begin{figure}[!htb]
    \centering
    \includegraphics[width=0.32\textwidth]{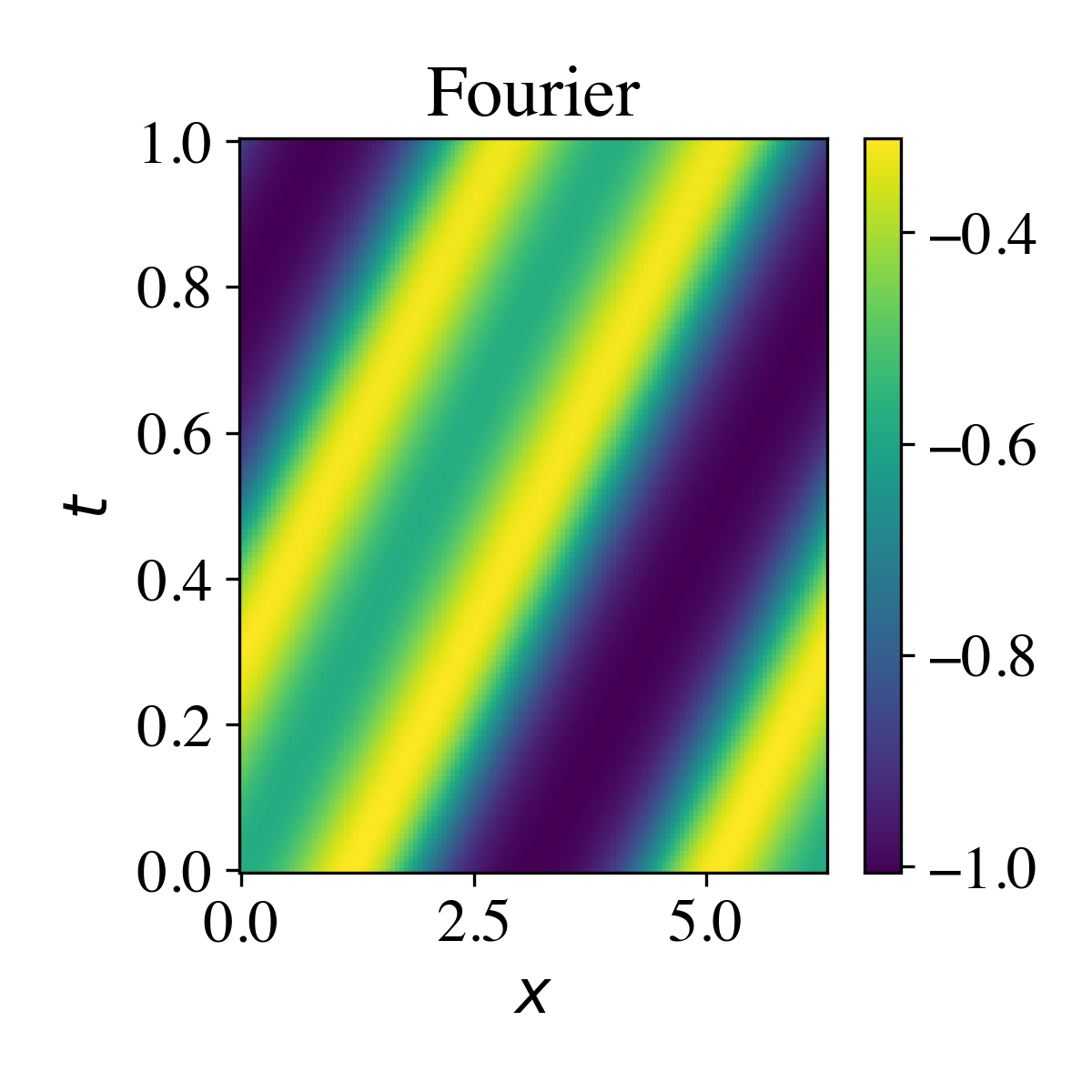}
    \includegraphics[width=0.32\textwidth]{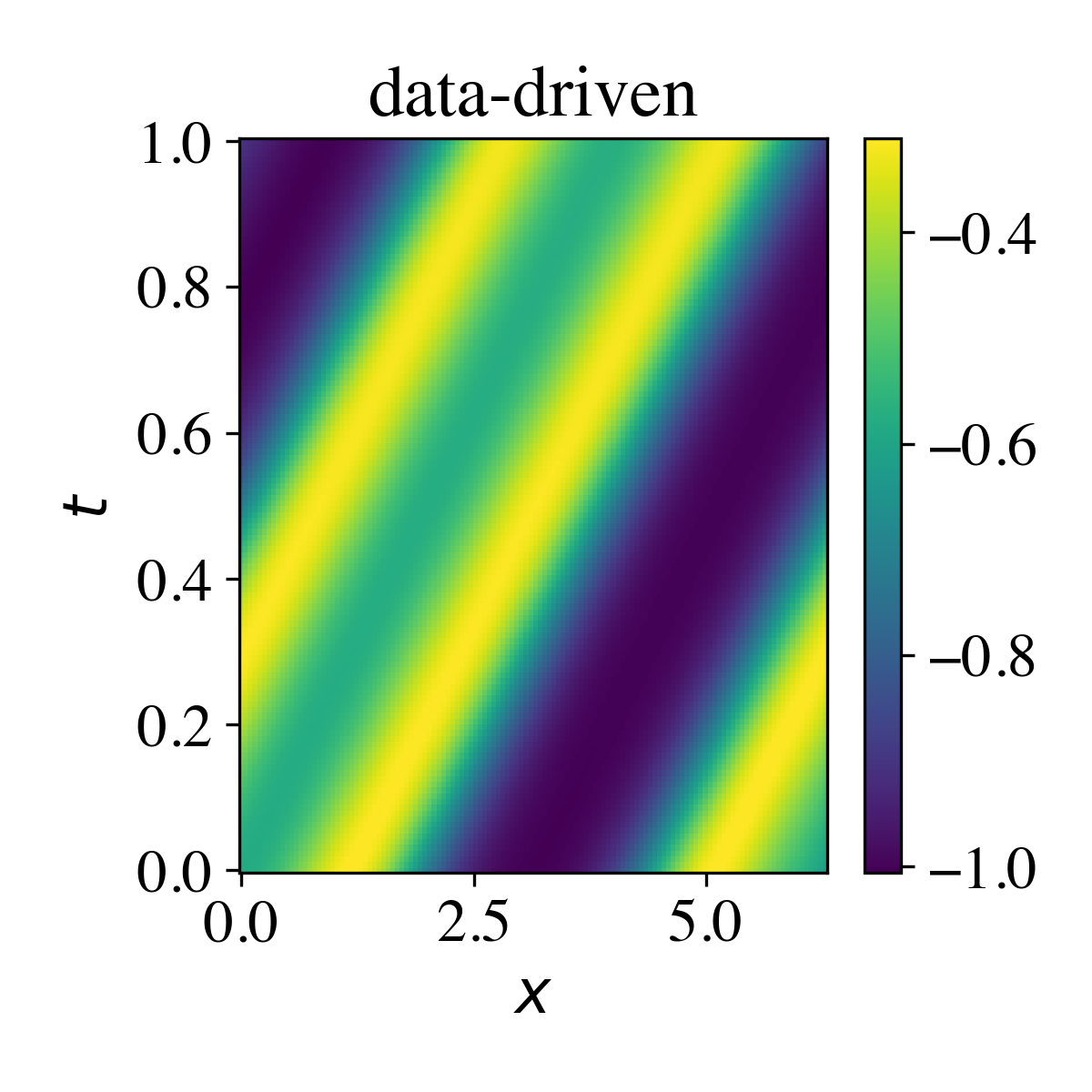}
    \includegraphics[width=0.32\textwidth]{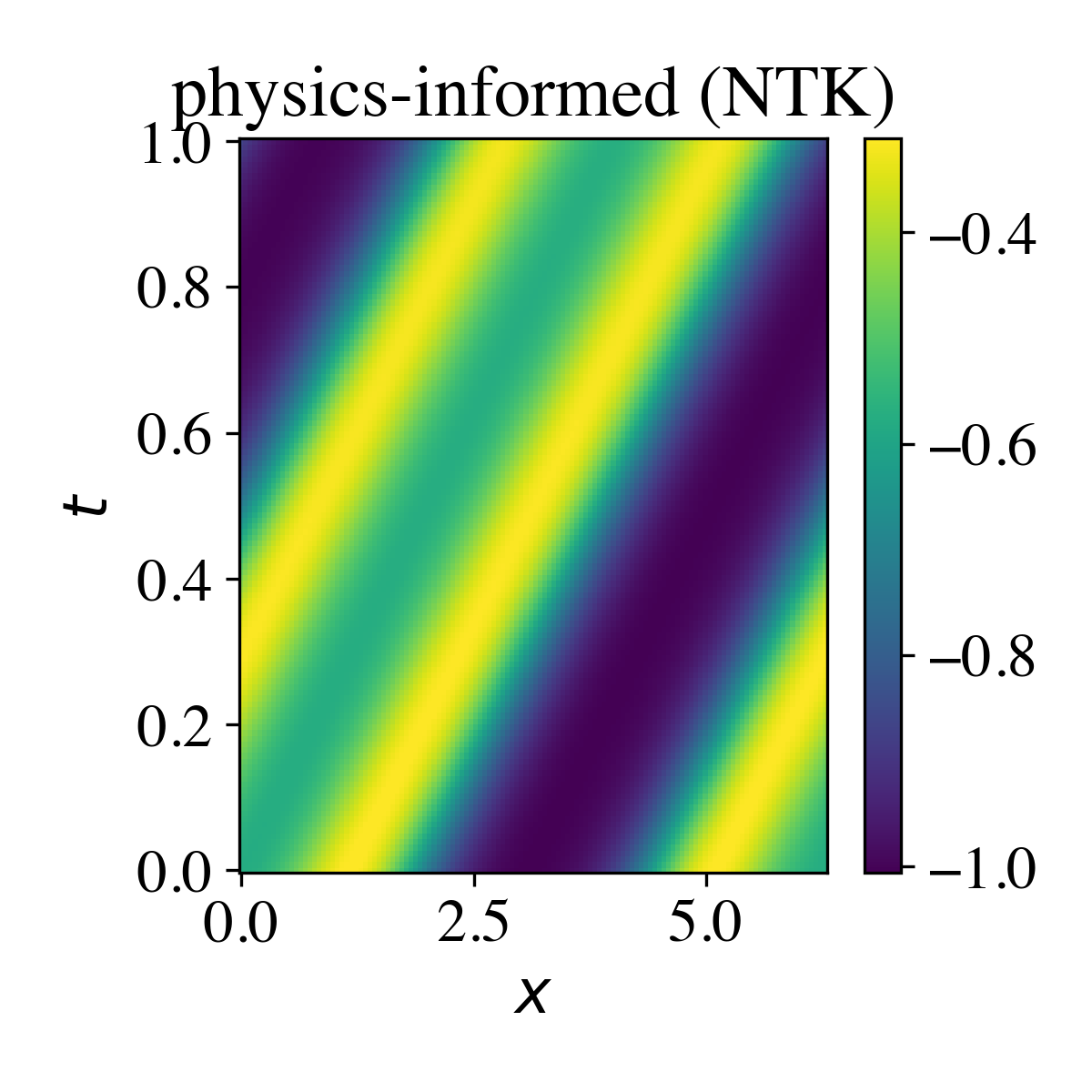}\\
    \hfill
    \includegraphics[width=0.32\textwidth]{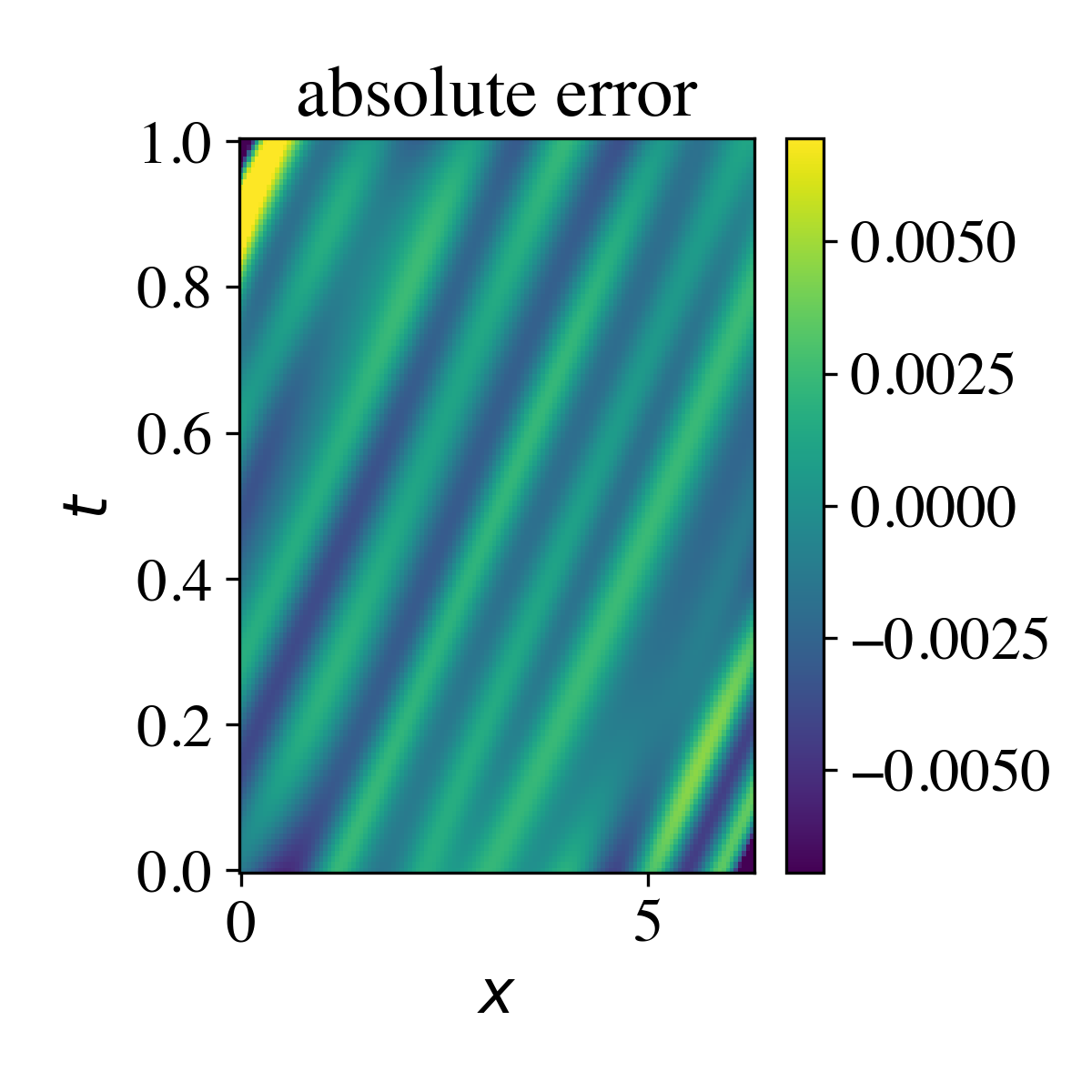}
    \includegraphics[width=0.32\textwidth]{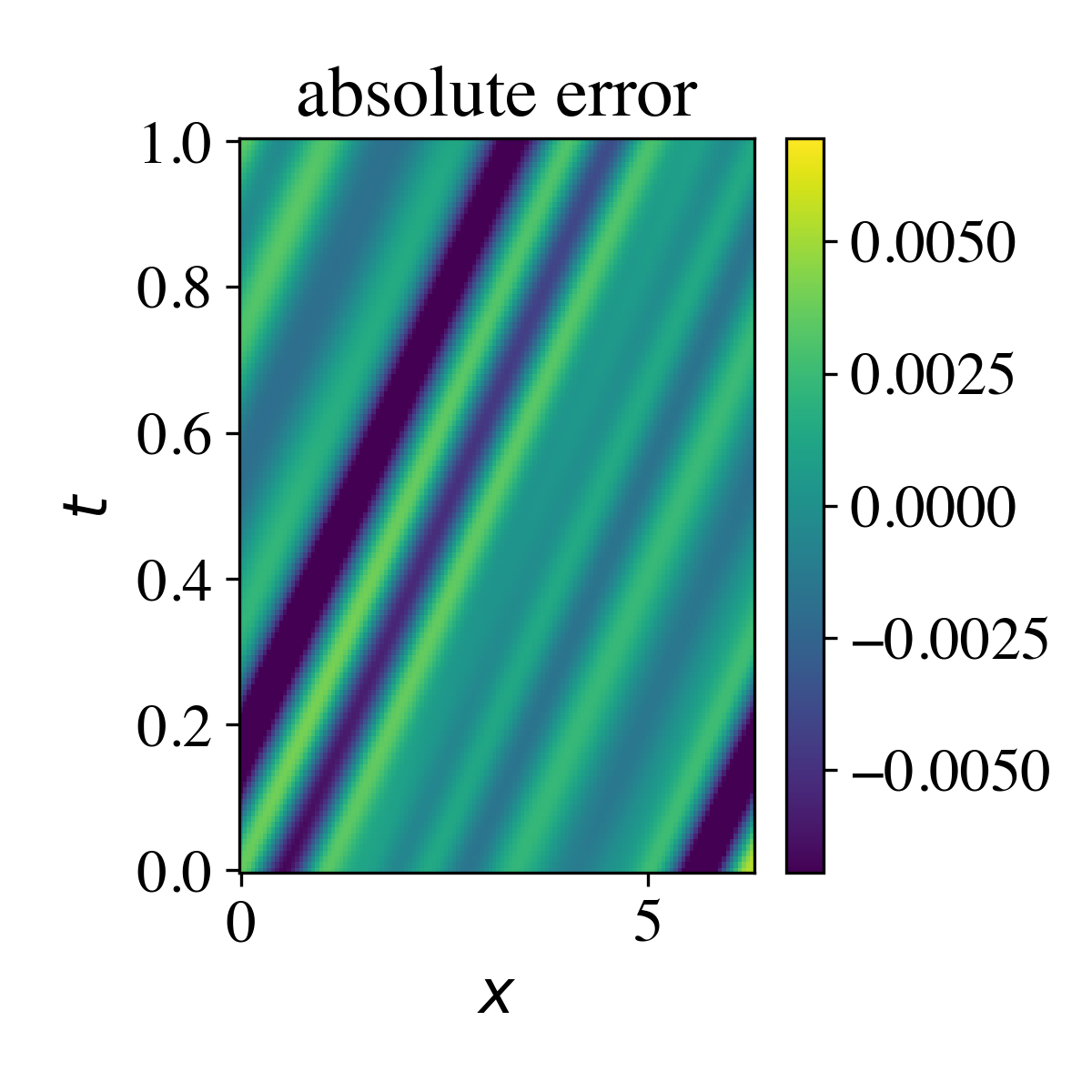}
    \caption{(Left) Fourier reference solution field, (middle) data-driven, (right) physics-informed DeepONet predictions for advection-diffusion. The absolute error is calculated as $\hat{s} - s$.}
    \label{fig:adv-diff-fourier}
\end{figure}

To show the difference between data-driven and physics-informed training, we train both with data from a solution field computed using a Fourier pseudo-spectral method and with a physics-informed loss function, which does not use data. The data-driven and physics-informed DeepONets with fixed weights use the MLP architecture as in \cite{meuris_machine-learning-based_2023}. The physics-informed DeepONet trained with local NTK adaptive weights uses the modified DeepONet architecture with an additional hidden layer \cite{wang_improved_2022}. Figure \ref{fig:adv-diff-fourier} shows the reference solution field generated by the Fourier method for one independent test sample initial condition and the predictions from the data-driven and physics-informed (NTK) DeepONets. Qualitatively, the approximations show agreement with the solution. Table \ref{tab:adv-diff-average-errors} quantifies the average relative $\ell_2$ errors for the data-driven, physics-informed with fixed weights, and physics-informed with local NTK adaptive weights compared with the reference solution. The relative $\ell_2$ error is calculated as in Equation \ref{eq:relL2}, where $\hat{s}$ is the approximation and $s$ is the reference. This error is integrated over time to calculate the average relative $\ell_2$ error (see Appendix \ref{sec:error}). All models approximate the solution well, within 1\% error. Importantly, for this problem, the physics-informed training is almost as accurate as data-driven training.

\begin{align}\label{eq:relL2}
    E(t_n) = \frac{||\hat{s}(x,t_n) - s(x,t_n)||_2}{||s(x,t_n)||_2}
\end{align}

\begin{table}[!htb]
    \centering
    \begin{tabular}{|c|c|c|}\hline
         & $w$ & Average relative $\ell_2$ error \\ \hline\hline
    data-driven & 128 & 0.44\% $\pm$ 0.32\% \\ \hline
    physics-informed (fixed) & 128 & 0.48\% $\pm$ 0.41\% \\ \hline
    physics-informed (NTK) & 128 & 0.82\% $\pm$ 0.54\% \\ \hline
    \end{tabular}
    \caption{Average errors across 100 test samples for advection-diffusion.}
    \label{tab:adv-diff-average-errors}
\end{table}

\begin{figure}[!htb]
    \centering
    \includegraphics[width=0.45\textwidth]{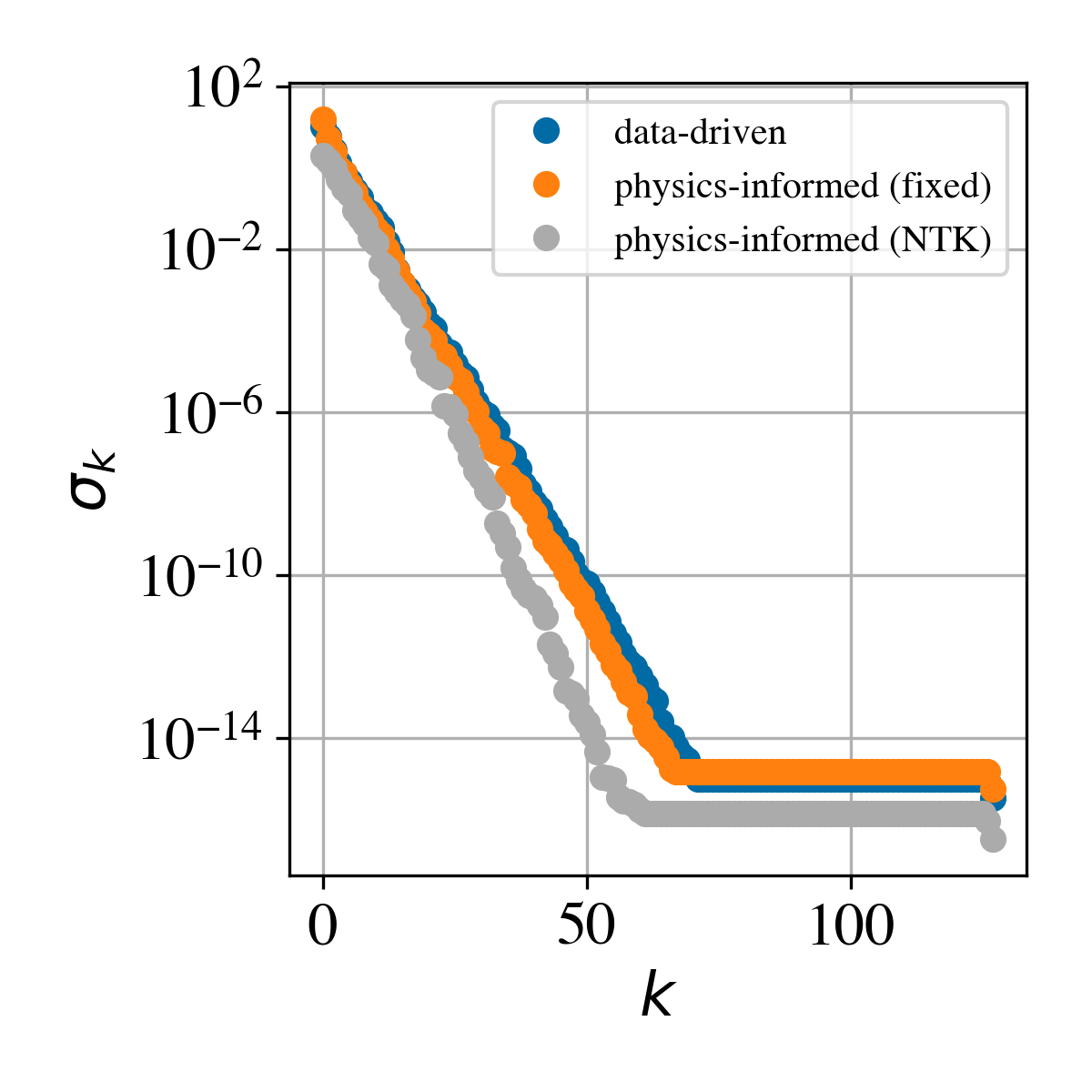}
    \includegraphics[width=0.45\textwidth]{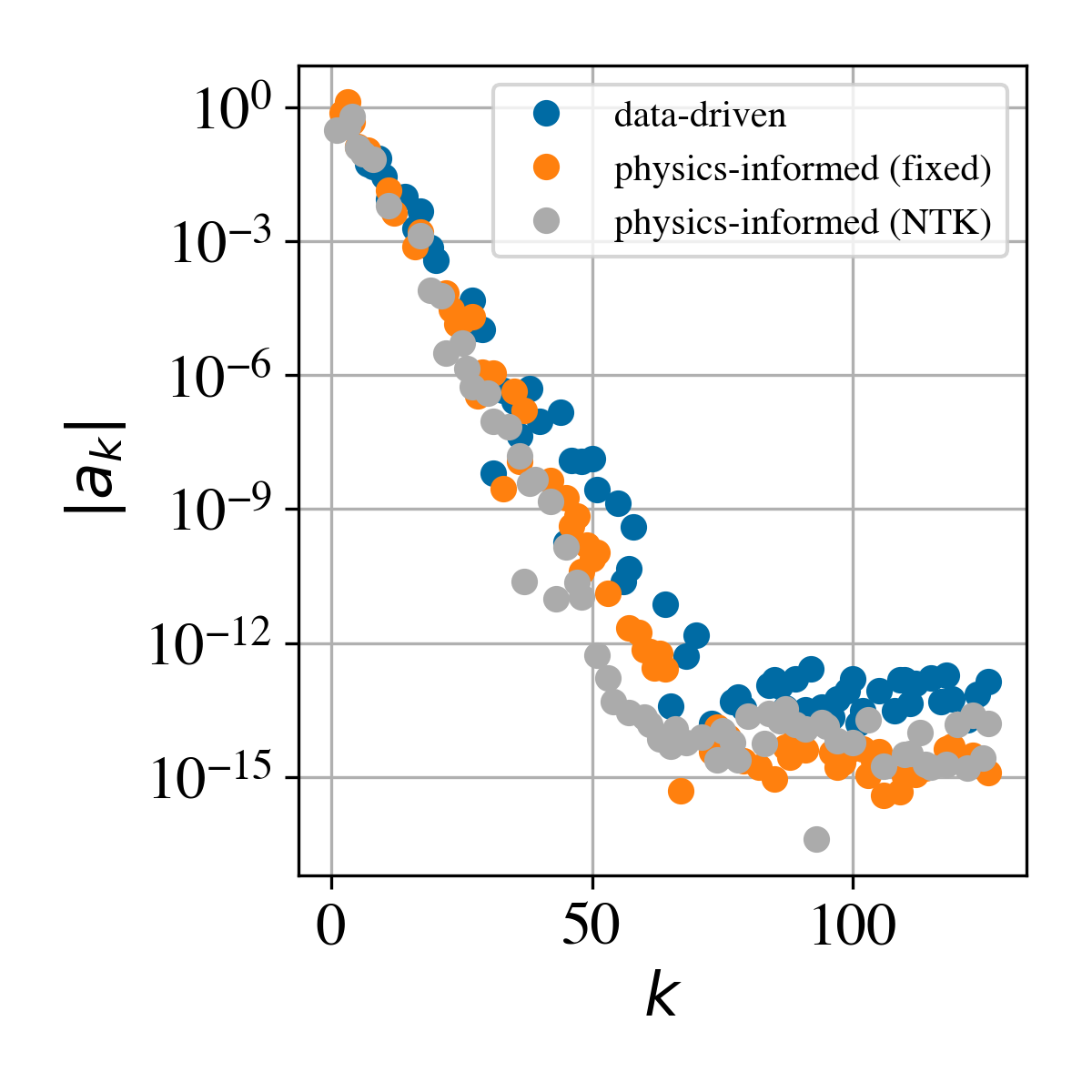}
    \caption{ (Left) Singular values and (right) expansion coefficients $e^{\sin(x)}$ for data-driven and physics-informed DeepONets for advection-diffusion.}
    \label{fig:adv-diff-B-a-k}
\end{figure}

\begin{figure}[!htb]
    \centering
    \includegraphics[width=\textwidth]{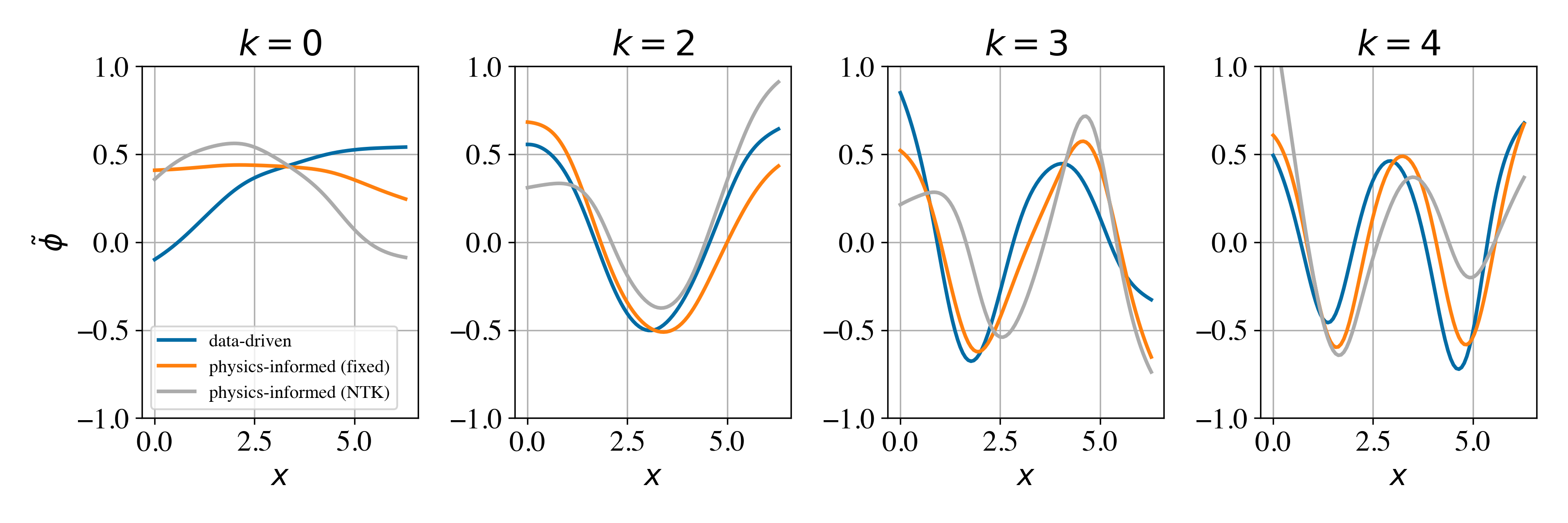}
    \caption{Custom basis functions, plotting with consistent boundaries $\Tilde{\phi}(x = 0 \text{ or } 2\pi)$ for advection-diffusion.}
    \label{fig:adv-diff-basis}
\end{figure}

The singular values accompanying each basis function provide a measure of the contribution of each function to the trunk net function space \cite{meuris_machine-learning-based_2023}. The decay of the expansion coefficients, calculated by $a_k = \langle \tilde{\phi}, f \rangle$ for function $f = e^{\sin(x)}$, relates to the accuracy of the trained model. Figure \ref{fig:adv-diff-B-a-k} shows the decay of the singular values and the expansion coefficients for the DeepONets trained for advection-diffusion. The expansion coefficients of all the trained models decay to machine precision. The singular values of the basis functions extracted from the physics-informed model trained with NTK adaptive weighting show faster decay than those from the physics-informed DeepONet model with fixed weights and the data-driven DeepONet model.  When DeepONets train well, evident through the decay of the expansion coefficients, then the decay of the singular values shows the degrees of freedom that are important in representing the solution space. A sample of the custom basis functions is shown in Figure \ref{fig:adv-diff-basis}. The basis functions (except for $k=0$) show good agreement across all the models (data-driven and physics-informed), demonstrating potential universality in what is learned by DeepONets during the training process. The first 10 custom basis functions can be found in Appendix \ref{sec:adv-diff-basis}.

\begin{figure}[!htb]
    \centering
    \includegraphics[width=0.32\textwidth]{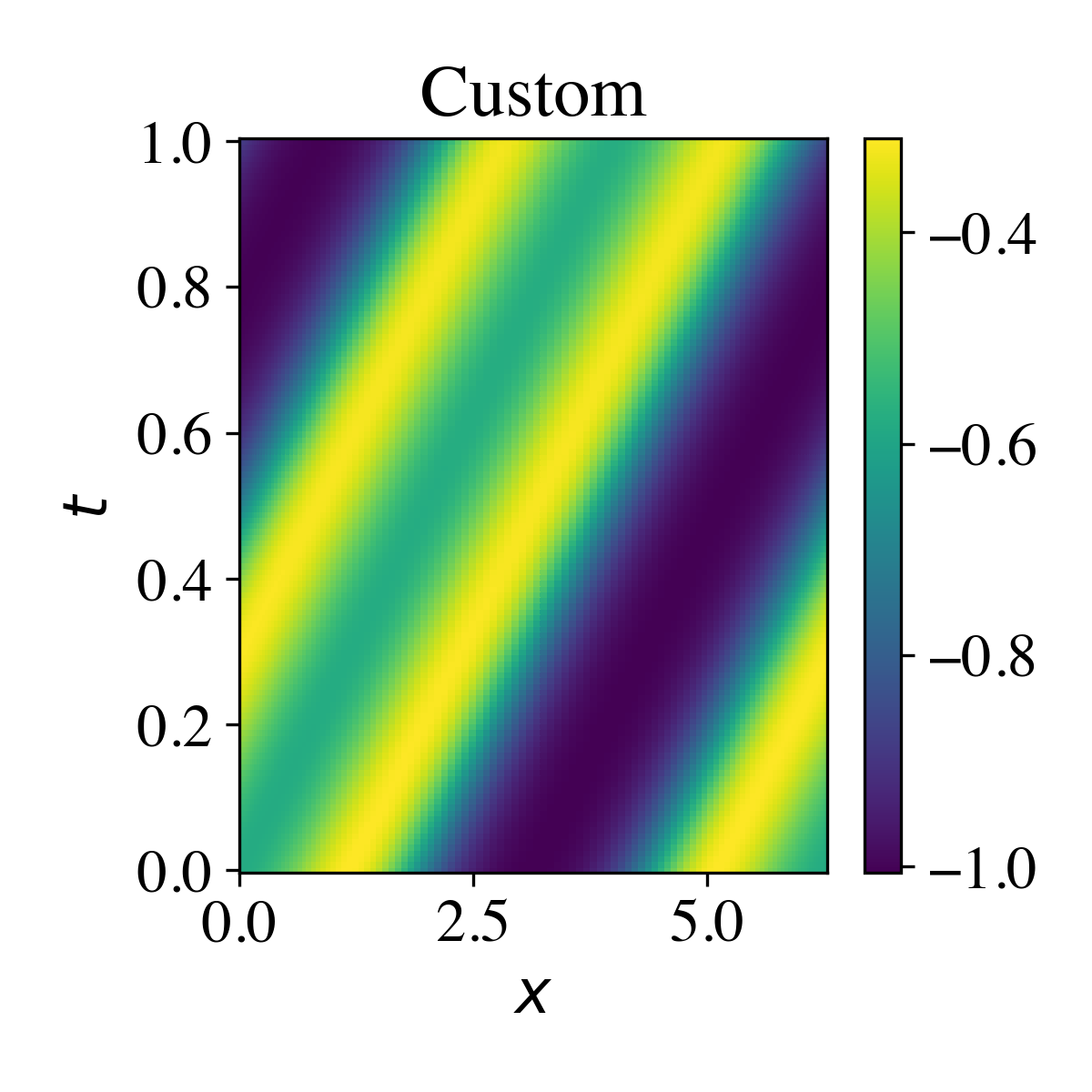}
    \includegraphics[width=0.32\textwidth]{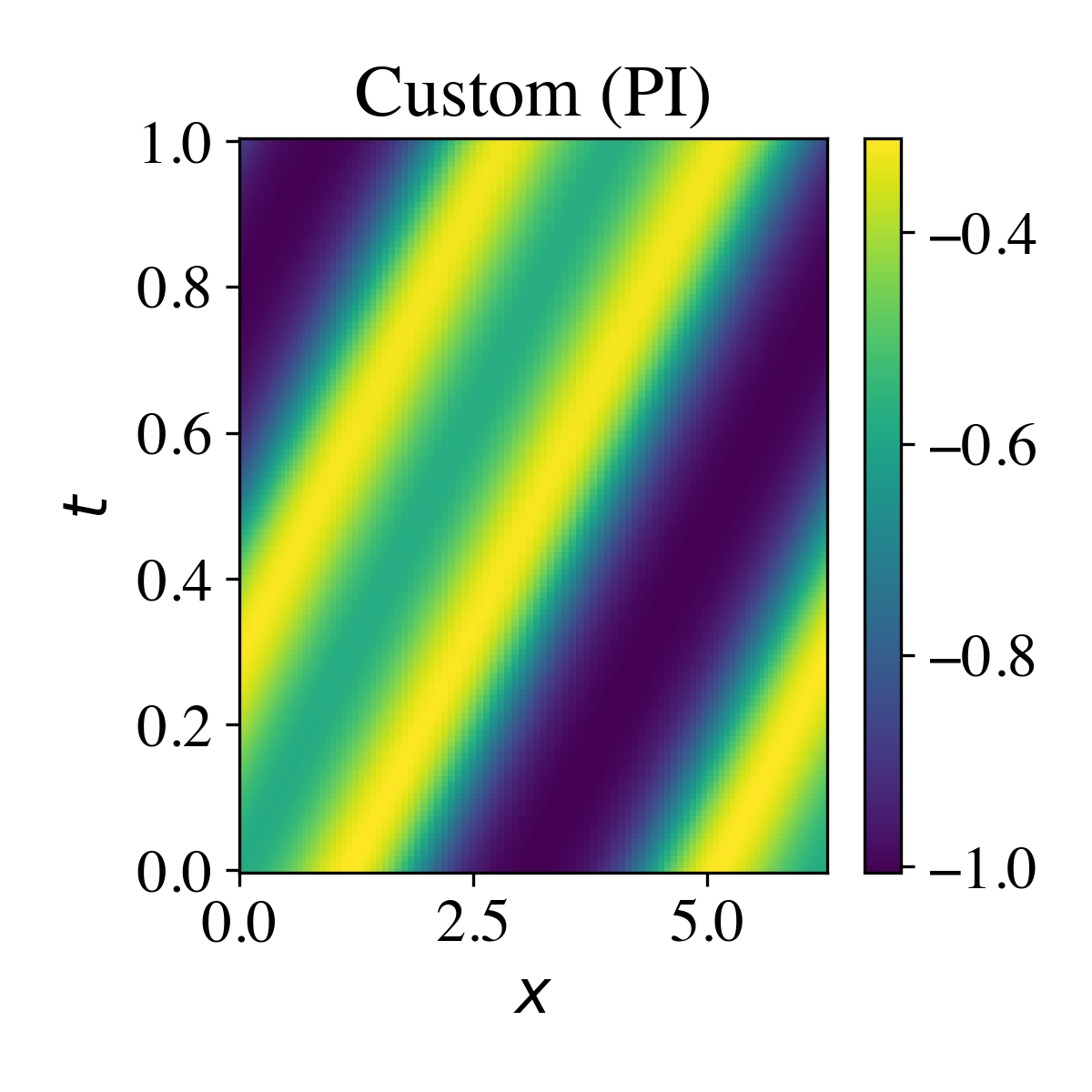}
    \caption{Spectral evolution with (left) 62 data-driven and (right) 47 physics-informed custom basis functions for advection-diffusion.}
    \label{fig:adv-diff-spectral}
\end{figure}

\begin{figure}[H]
    \centering
    \includegraphics[width=0.45\textwidth]{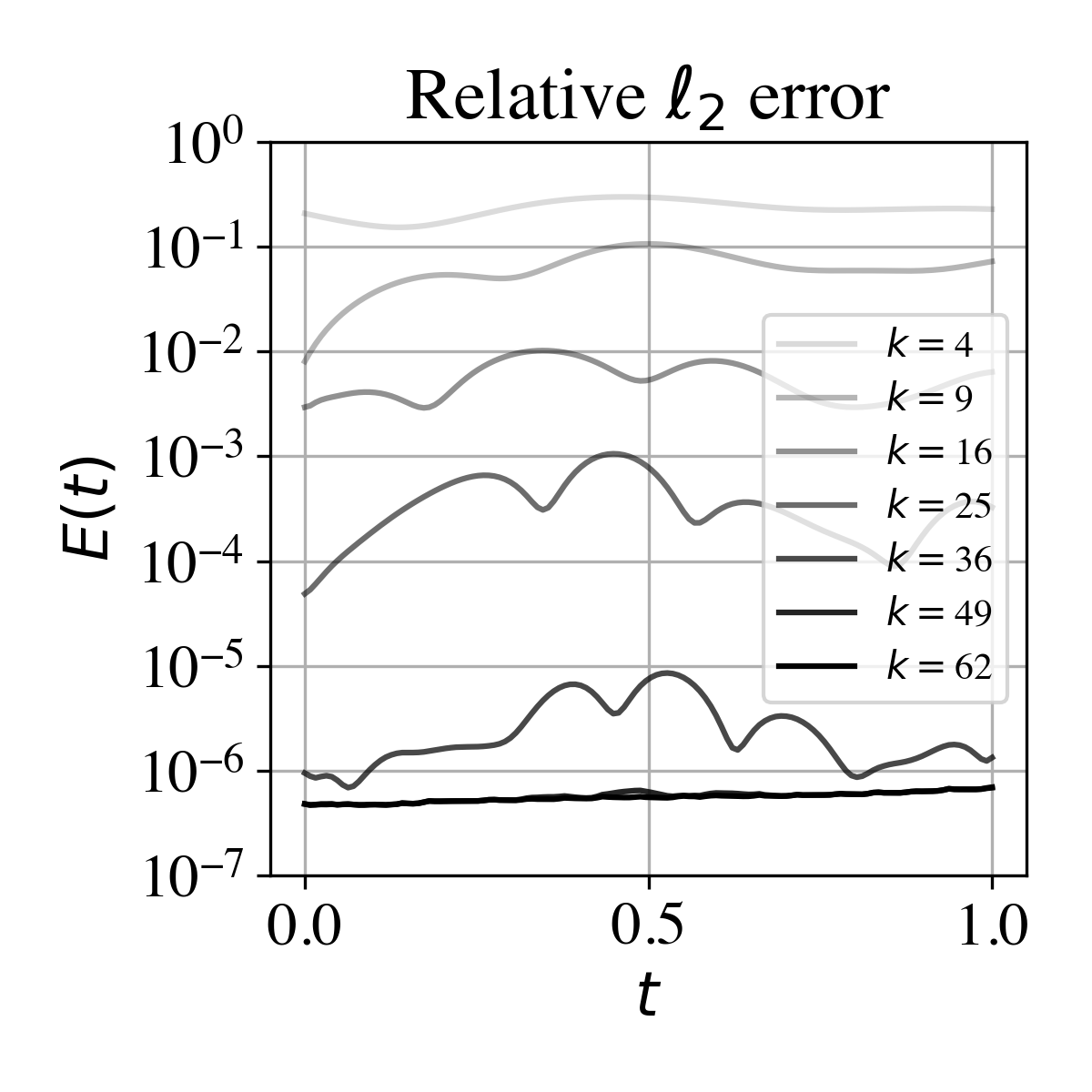}
    \includegraphics[width=0.45\textwidth]{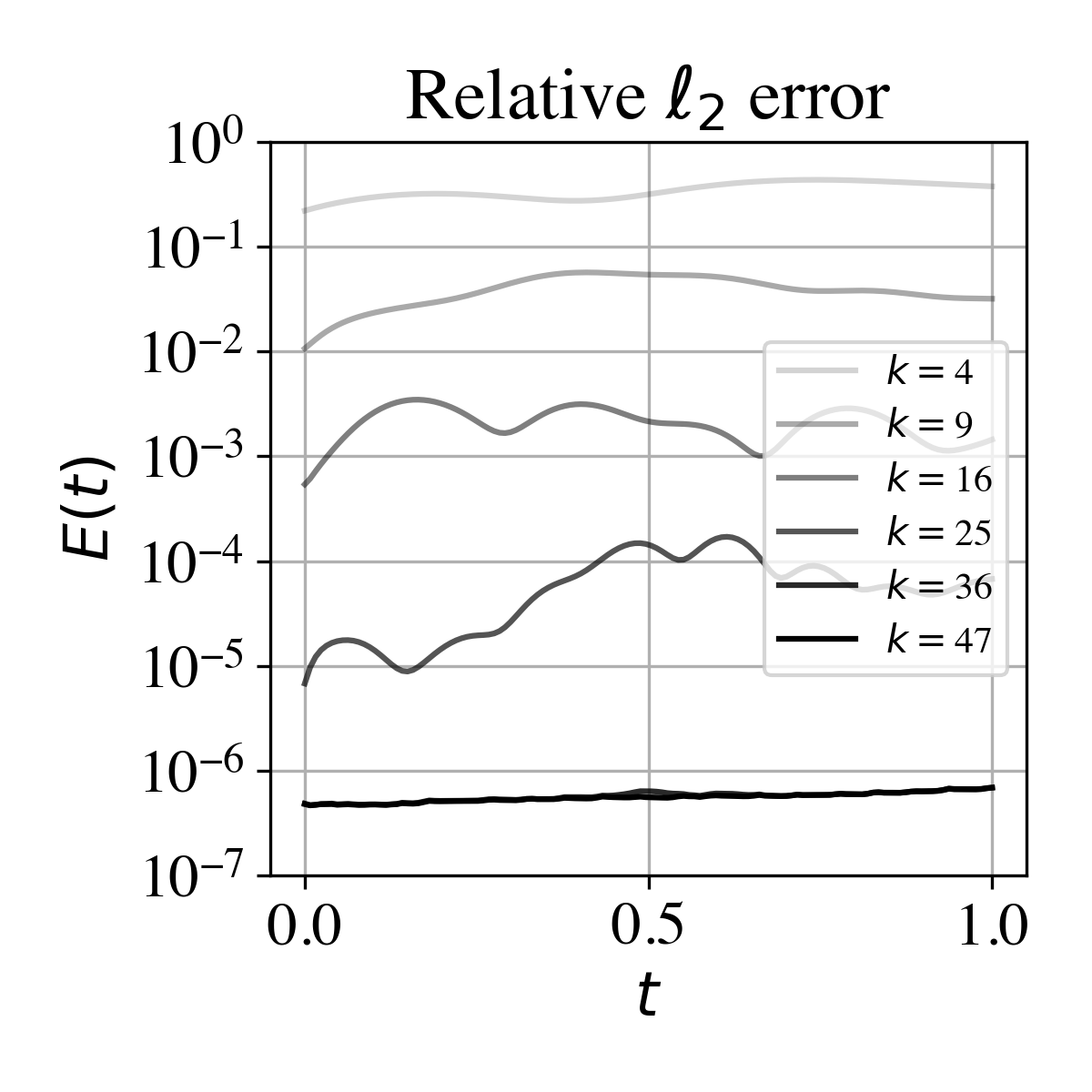}
    \includegraphics[width=0.45\textwidth]{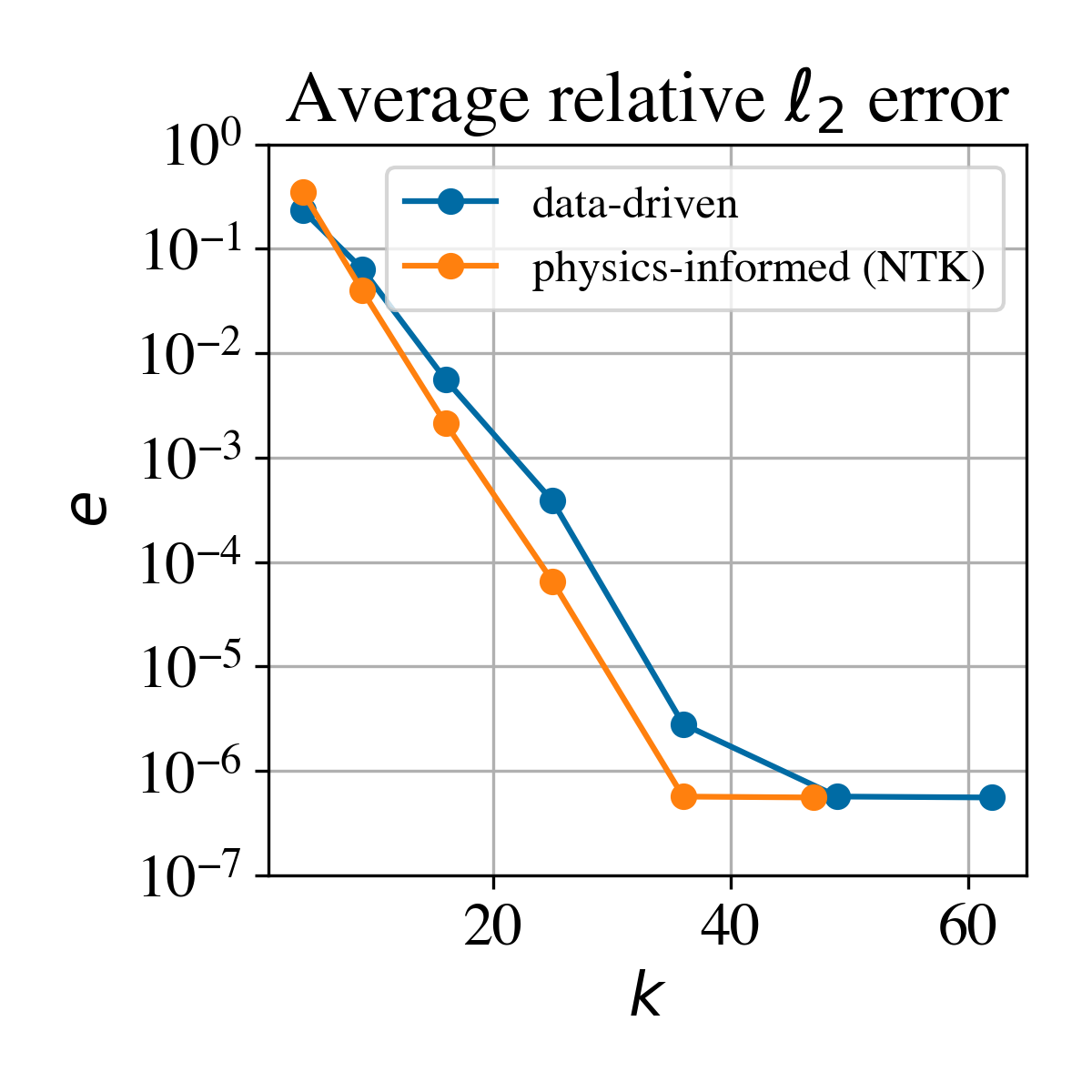}
    \caption{(Top left) Relative data-driven, (top right) relative physics-informed, and (bottom) average relative $\ell_2$ test sample errors for random test sample initial condition with $k$ custom basis functions for advection-diffusion.}
    \label{fig:adv-diff-error-plots}
\end{figure}

Then, following \cite{meuris_machine-learning-based_2023}, we can use the basis functions from the physics-informed DeepONet with NTK-guided weights in evolving the approximation using a spectral approach. Corresponding to singular values greater than the specified cutoff of $10^{-13}$, we use 62 data-driven custom basis functions to approximate the solution. For this same cutoff, we only require 47 physics-informed custom basis functions. The approximation fields from the test initial condition are shown in Figure \ref{fig:adv-diff-spectral}. The top plots of Figure \ref{fig:adv-diff-error-plots} show the relative $\ell_2$ errors as we use more custom basis functions in generating the solution field, approaching each respective cutoff, for data-driven and physics-informed. The errors in time are relatively consistent and decrease overall as we use more basis functions. As the cutoff is approached, the errors saturate. From the bottom plot, using the custom basis functions from the physics-informed DeepONet results in lower error. Shown in Table \ref{tab:adv-diff-spectral-error}, we reach the same magnitude of the average relative $\ell_2$ error of $10^{-7}$ for both data-driven and physics-informed custom spectral approximations. However, we only require 47 physics-informed custom basis functions to reach this error compared to 62 data-driven custom basis functions. From this, we can see that data-driven and physics-informed DeepONets train to a similar relative error for this equation. However, when considering using the learned DeepONet basis functions in a spectral method, the physics-informed DeepONet representation used 25\% fewer basis functions to reach the same relative error. This suggests that the physics-informed basis functions are more effective in representing the solution space for this linear PDE.

\begin{table}[!htb]
    \centering
    \begin{tabular}{|c|c|c|}\hline
         & $p$ & Average relative $\ell_2$ error \\ \hline\hline
    data-driven & 62 & $5.767 \times 10^{-7}$ \\ \hline
    physics-informed (NTK) & 47 & $5.576 \times 10^{-7}$ \\ \hline
    \end{tabular}
    \caption{Average errors for independent test sample for advection-diffusion.}
    \label{tab:adv-diff-spectral-error}
\end{table}

\subsection{Viscous Burgers}
The viscous Burgers equation has become a standard benchmark for physics-informed DeepONets \cite{wang_learning_2021, wang_improved_2022}. While physics-informed DeepONets train well for high viscosities ($\nu = 0.1, 0.01$), they can struggle to train for low viscosities  ($\nu = 0.0001$), making this benchmark an interesting case to understand physics-informed DeepONet training. See Appendix \ref{sec:cost} for a cost assessment for training with different weighting schemes for the viscous Burgers benchmark case.

\subsubsection{$\nu = 0.1$}

\begin{figure}[!htb]
    \centering
    \includegraphics[width=0.32\textwidth]{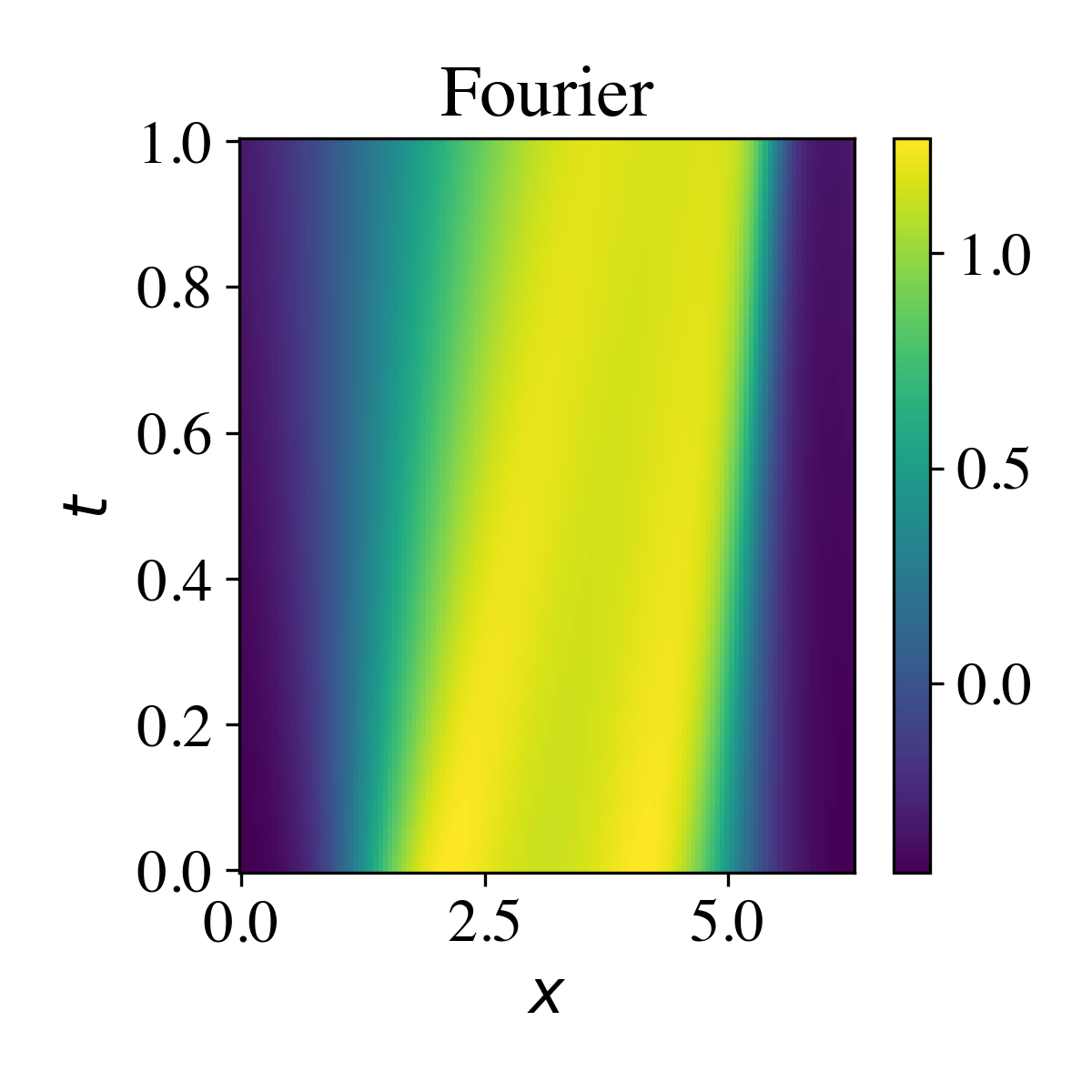}
    \includegraphics[width=0.32\textwidth]{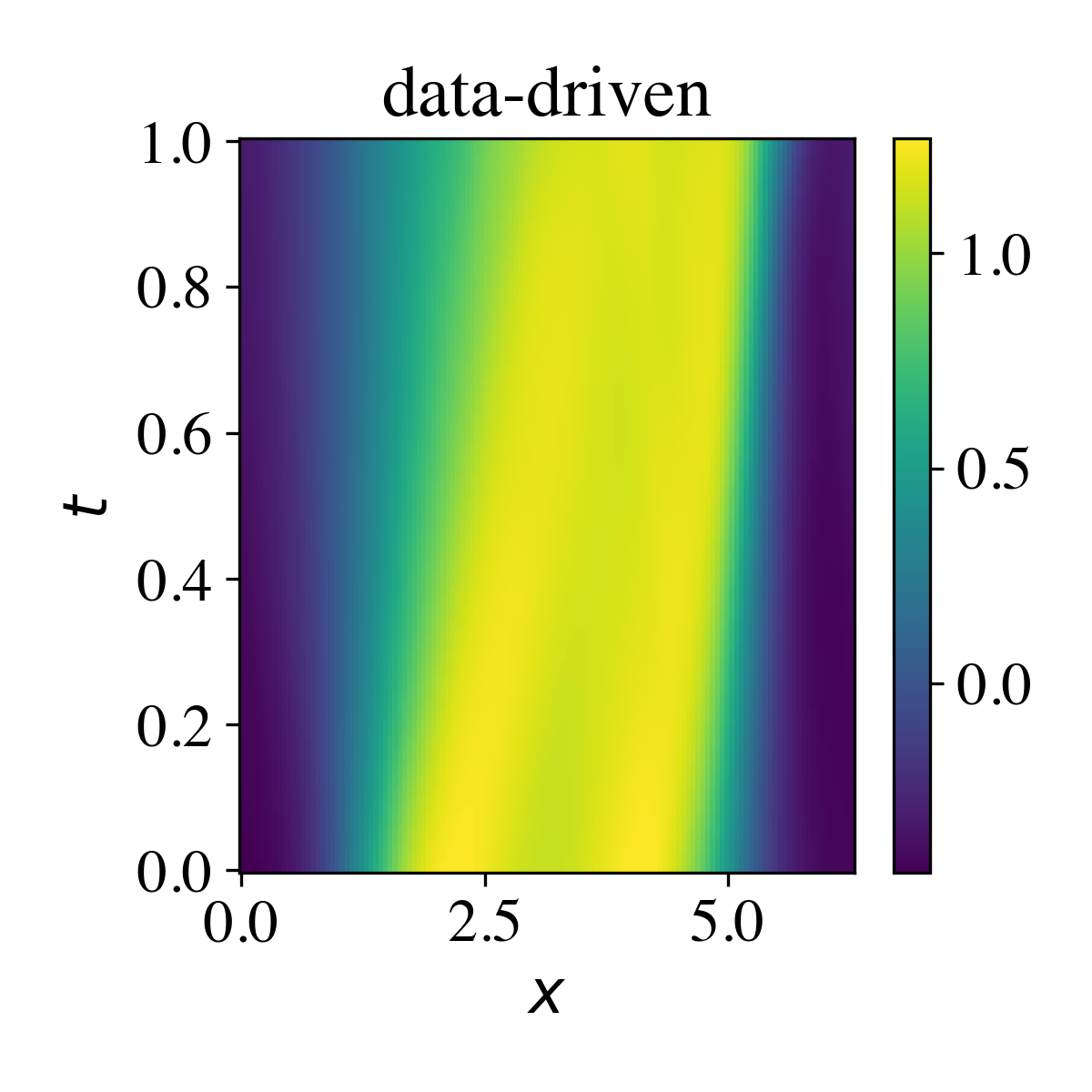}
    \includegraphics[width=0.32\textwidth]{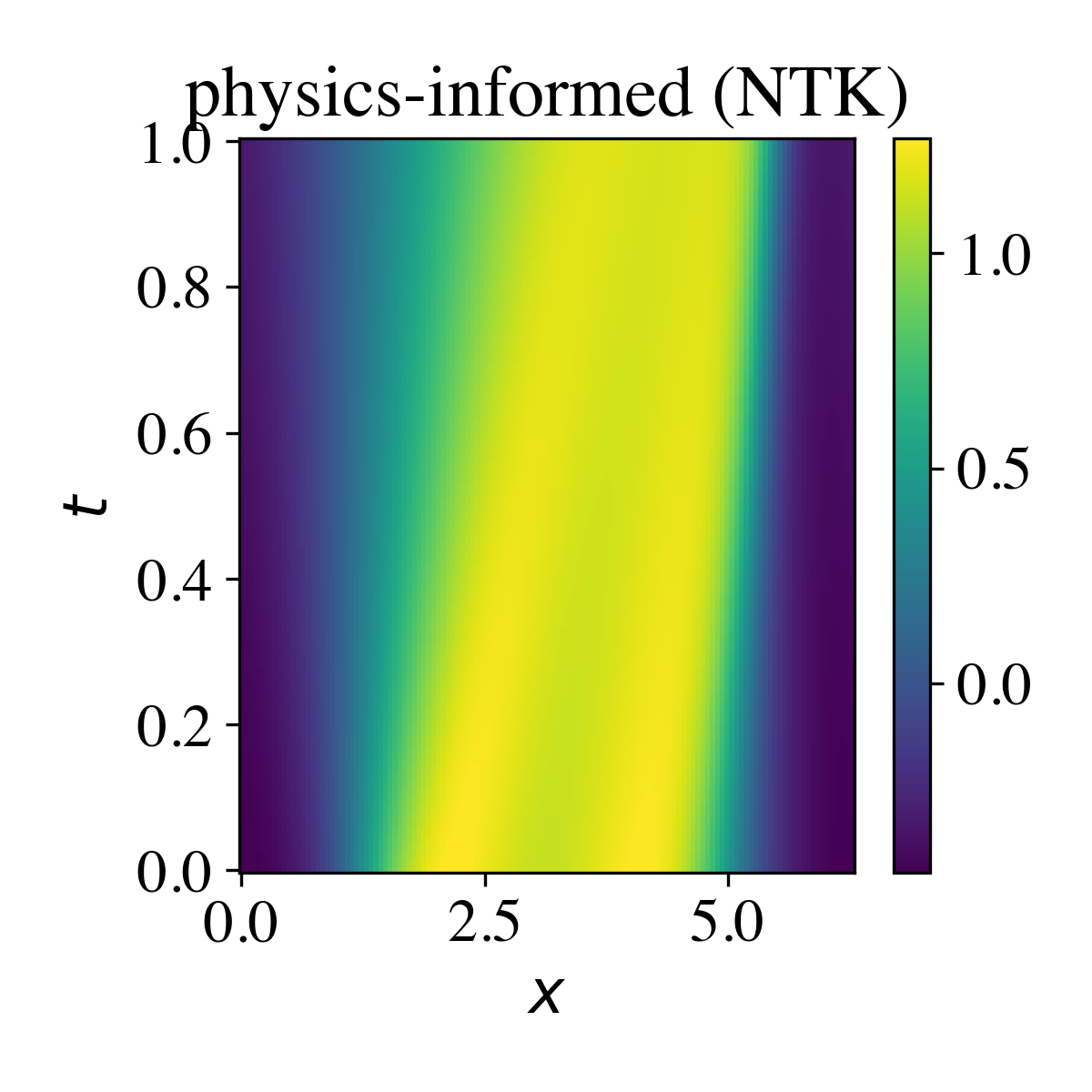}\\
    \hfill
    \includegraphics[width=0.32\textwidth]{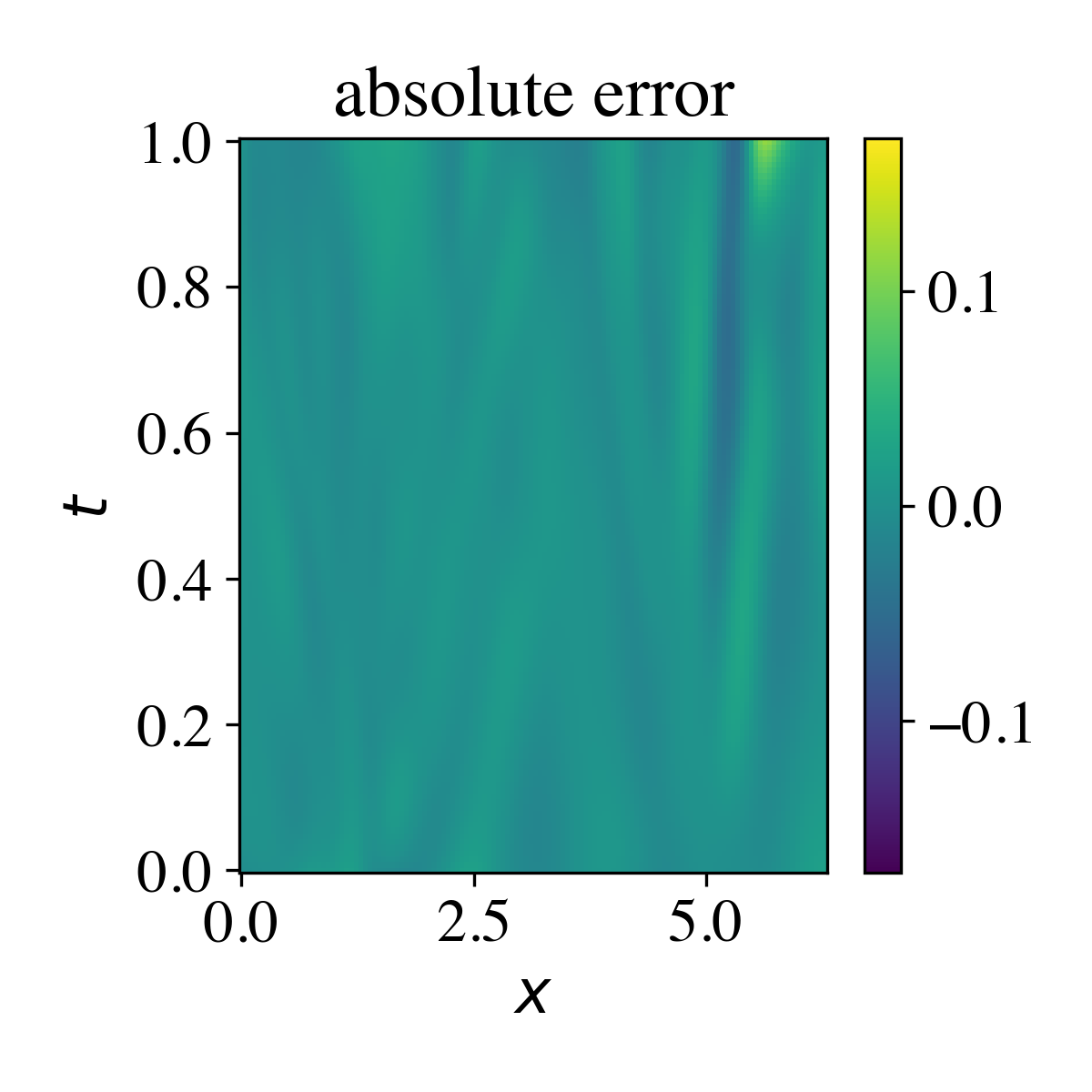}
    \includegraphics[width=0.32\textwidth]{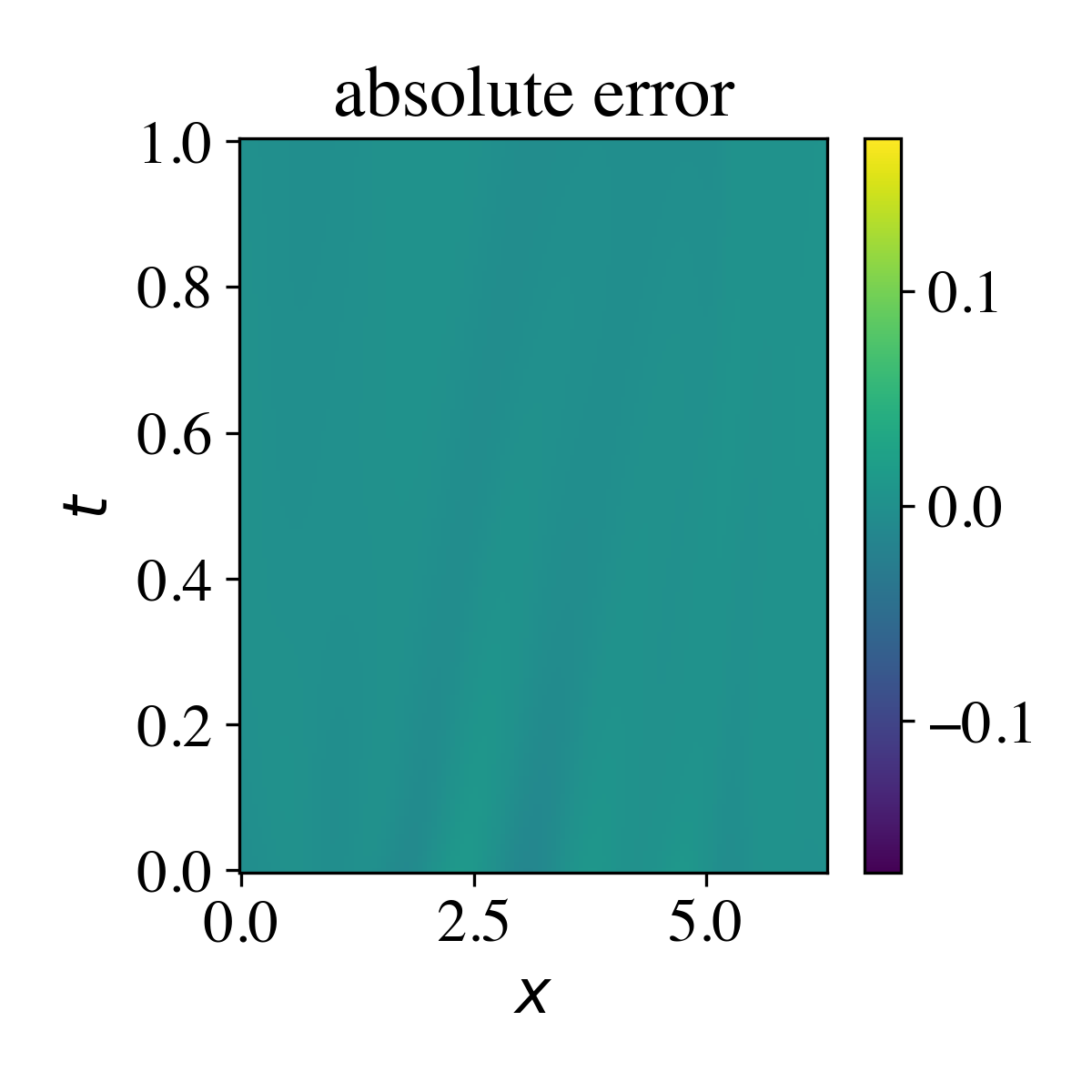}
    \caption{(Left) Fourier reference solution field, (middle) data-driven, and (right) physics-informed (NTK) DeepONet predictions for viscous Burgers with $\nu = 0.1$. The absolute error is calculated as $\hat{s} - s$.}
    \label{fig:viscous-burgers-fourier}
\end{figure}

Consider the viscous Burgers equation 
\begin{align}
    \frac{\partial s}{\partial t} + s \frac{\partial s}{\partial x} - \nu \frac{\partial^2 s}{\partial x^2} = 0
\end{align}
on $(x,t) \in (0,2\pi) \times (0,1)$ with periodic boundary conditions. The viscosity coefficient is $\nu = 0.1$. The input to the branch network is the initial condition $s(x,0) = u(x)$. For data-driven training and for calculating errors, the solution field is computed using a Fourier pseudo-spectral method. The goal is to learn the solution operator $\mathcal{G} : u(x) \mapsto s(x,t)$. The data-driven and physics-informed DeepONets with fixed weights use the MLP architecture as in \cite{meuris_machine-learning-based_2023}. The physics-informed DeepONet trained with moderate local NTK adaptive weights uses the modified DeepONet architecture \cite{wang_improved_2022}. Figure \ref{fig:viscous-burgers-fourier} shows the reference solution field generated by the Fourier method for one independent test sample initial condition and the predictions from the data-driven and physics-informed (NTK) DeepONets. Table \ref{tab:burgers-average-errors} quantifies the average relative $\ell_2$ errors for the data-driven, physics-informed with fixed weights, and physics-informed with moderate local NTK adaptive weights compared with the reference solution. The physics-informed model with the NTK outperforms the physics-informed with fixed weights and the data-driven models.

\begin{table}[!htb]
    \centering
    \begin{tabular}{|c|c|c|}\hline
         & $w$ & Average relative $\ell_2$ error \\ \hline
    data-driven & 128 & 1.98\% $\pm$ 1.22\% \\ \hline
    physics-informed (fixed) & 128 & 4.48\% $\pm$ 3.00\% \\ \hline
    physics-informed (NTK) & 128 & 1.20\% $\pm$ 1.06\% \\ \hline
    \end{tabular}
    \caption{Average errors across 100 test samples for viscous Burgers with $\nu = 0.1$.}
    \label{tab:burgers-average-errors}
\end{table}

\begin{figure}[!htb]
    \centering
    \includegraphics[width=0.45\textwidth]{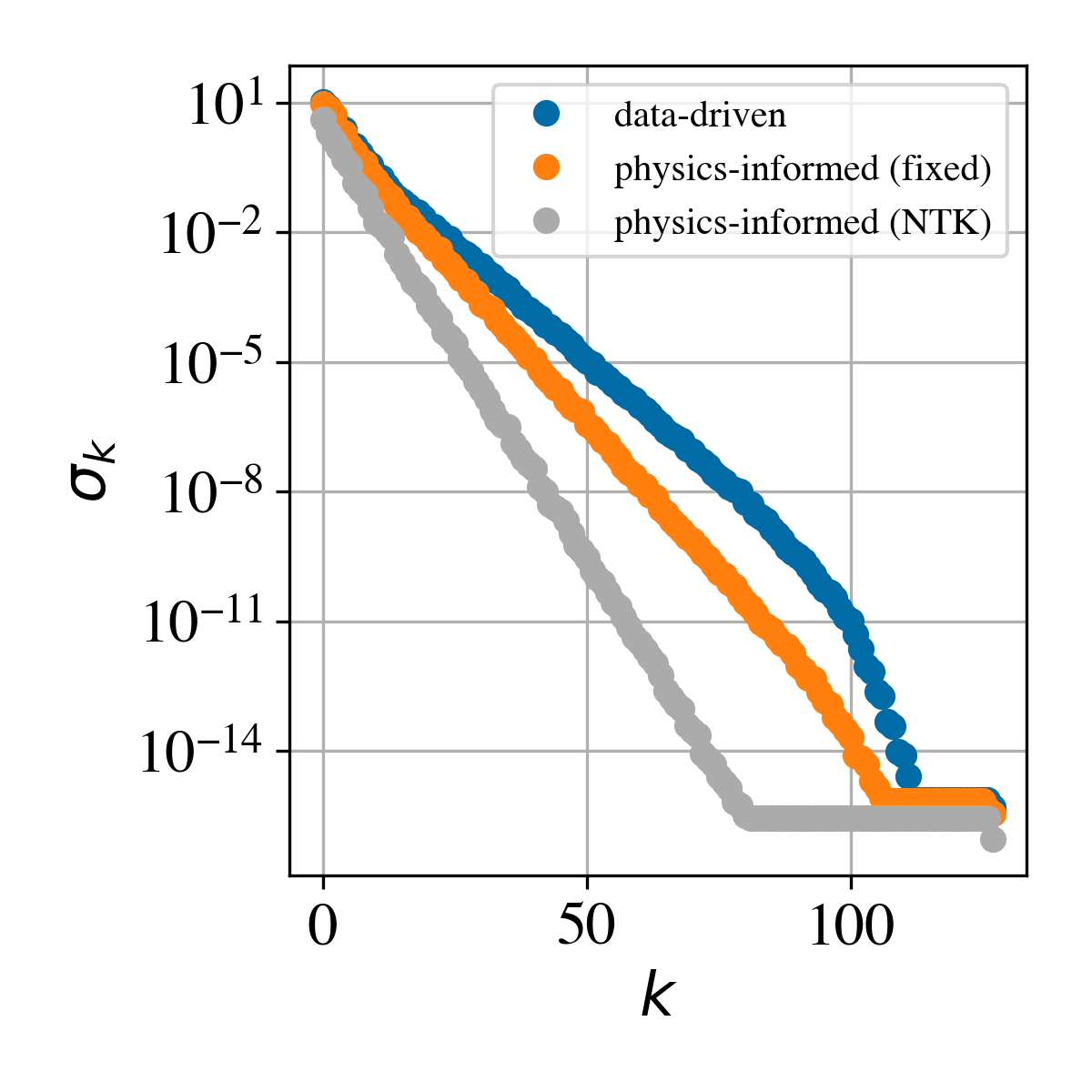}
    \includegraphics[width=0.45\textwidth]{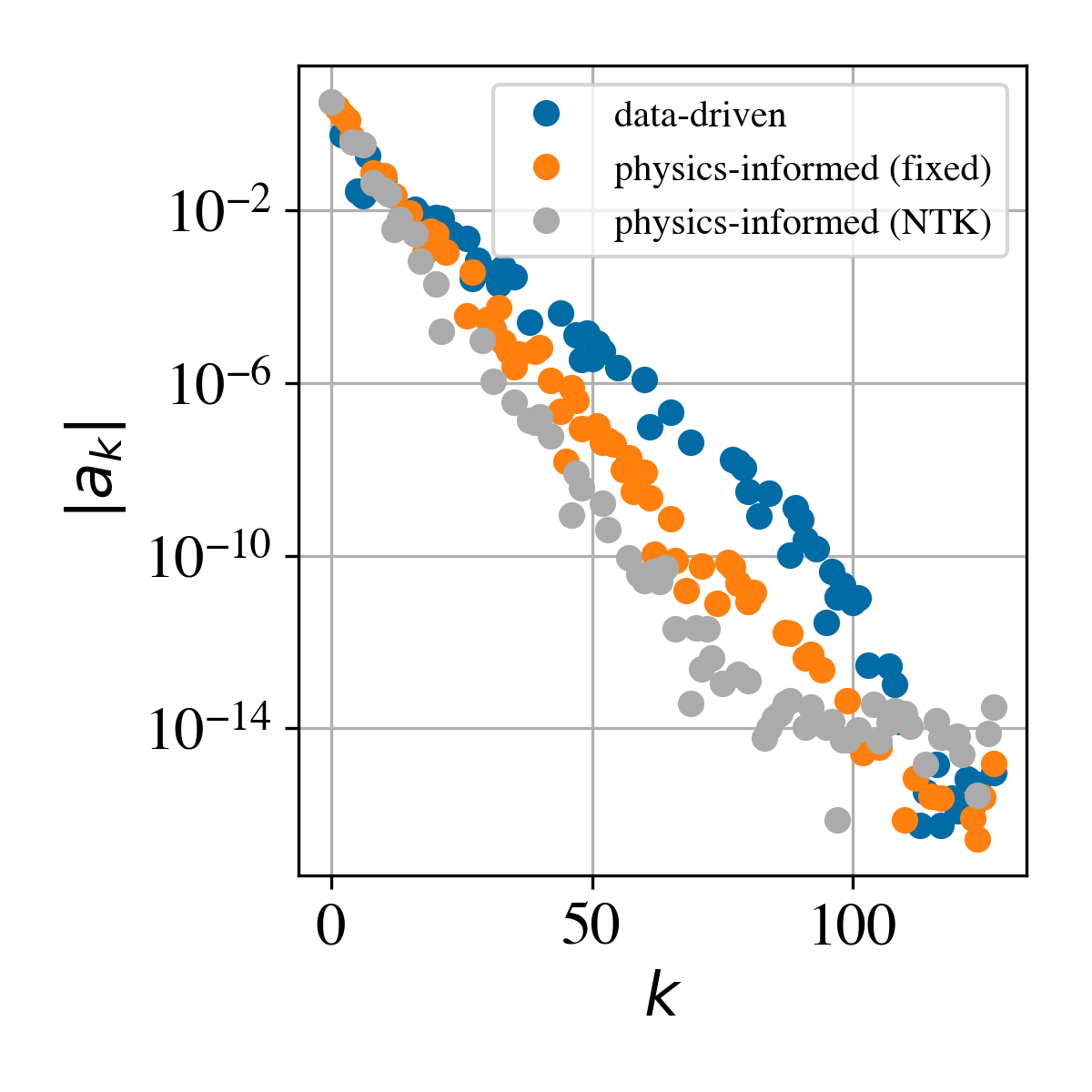}
    \caption{ (Left) Singular values and (right) expansion coefficients $e^{\sin(x)}$ for data-driven and physics-informed DeepONets for viscous Burgers with $\nu = 0.1$.}
    \label{fig:viscous-burgers-B-a-k}
\end{figure}

\begin{figure}[!htb]
    \centering
    \includegraphics[width=\textwidth]{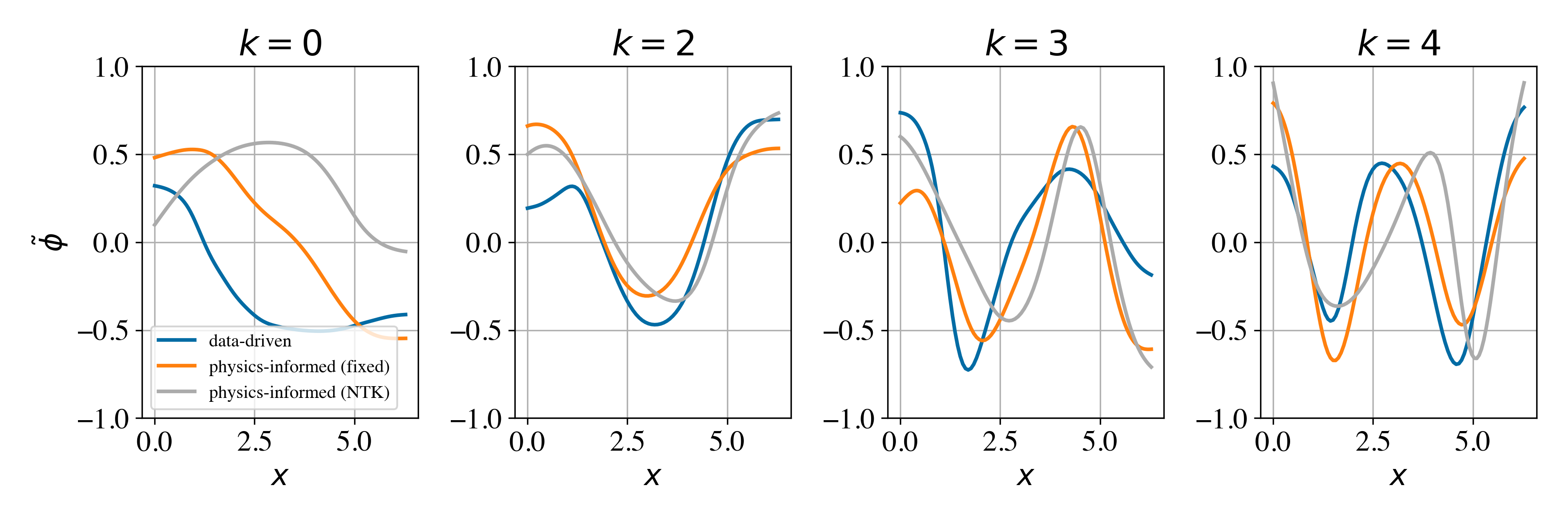}
    \caption{Custom basis functions, plotting with consistent boundaries $\Tilde{\phi}(x = 0 \text{ or } 2\pi)$ for viscous Burgers with $\nu = 0.1$.}
    \label{fig:viscous-burgers-basis}
\end{figure}

Figure \ref{fig:viscous-burgers-B-a-k} shows the decay of the singular values and the expansion coefficients for the DeepONets trained for viscous Burgers with $\nu = 0.1$. The singular values of the basis functions extracted from the physics-informed model trained with NTK adaptive weighting show significantly faster decay than those from the physics-informed DeepONet model with fixed weights and the data-driven DeepONet model. The expansion coefficients of all the trained models decay to machine precision, showing that the models are training effectively. Therefore, the decay of the singular values shows the degrees of freedom that are important in capturing the solution space. A sample of the custom basis functions is shown in Figure \ref{fig:viscous-burgers-basis}. Again, except for $k=0$, the agreement of the basis functions demonstrate the universality in what might be learned by DeepONets. The first 10 custom basis functions can be found in Appendix \ref{sec:burgers-basis-1e-1-2pi}.

\begin{figure}[H]
    \centering
    \includegraphics[width=0.32\textwidth]{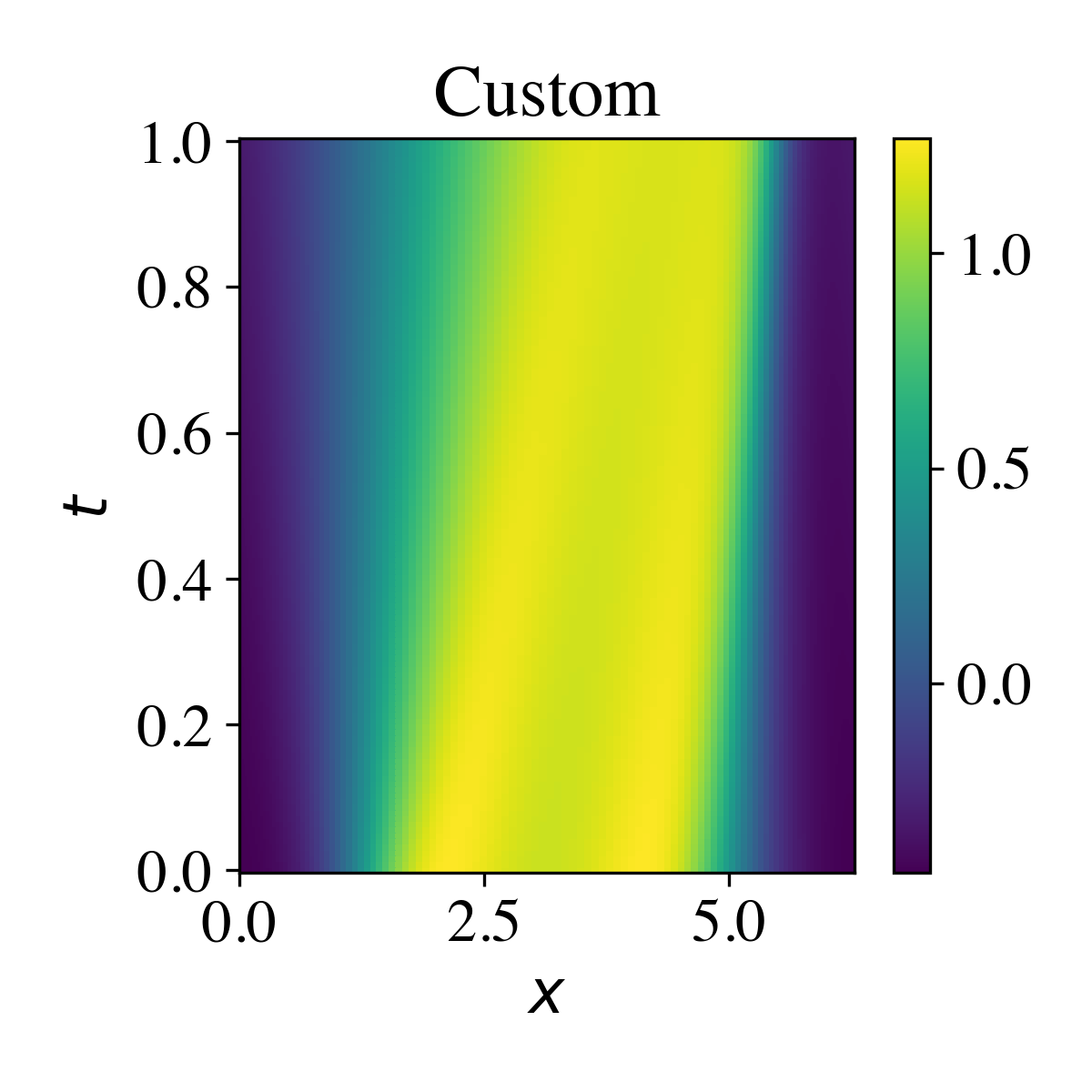}
    \includegraphics[width=0.32\textwidth]{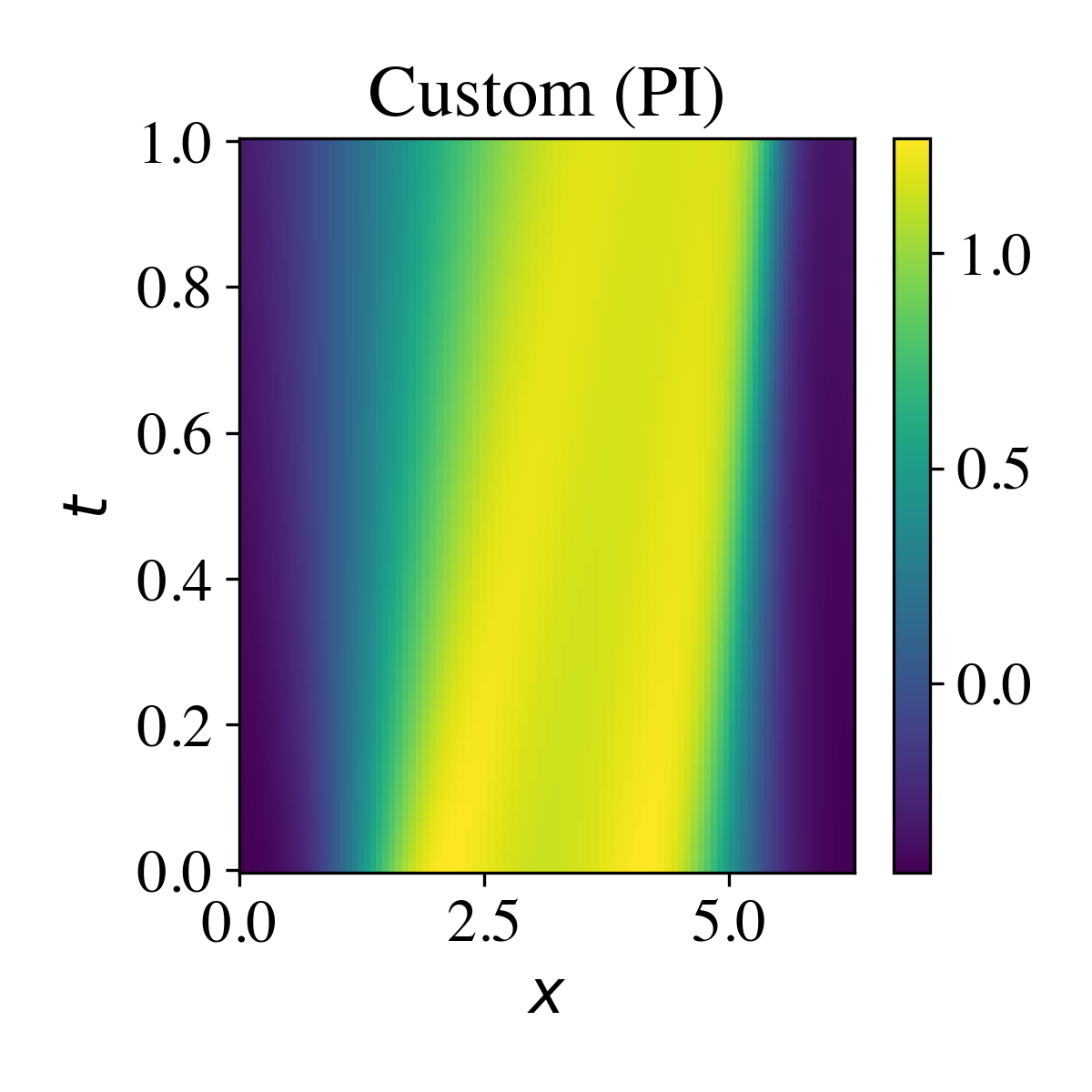}
    \caption{Spectral evolution with (left) data-driven and (right) physics-informed custom basis functions for viscous Burgers with $\nu = 0.1$.}
    \label{fig:viscous-burgers-spectral}
\end{figure}

\begin{figure}[H]
    \centering
    \includegraphics[width=0.45\textwidth]{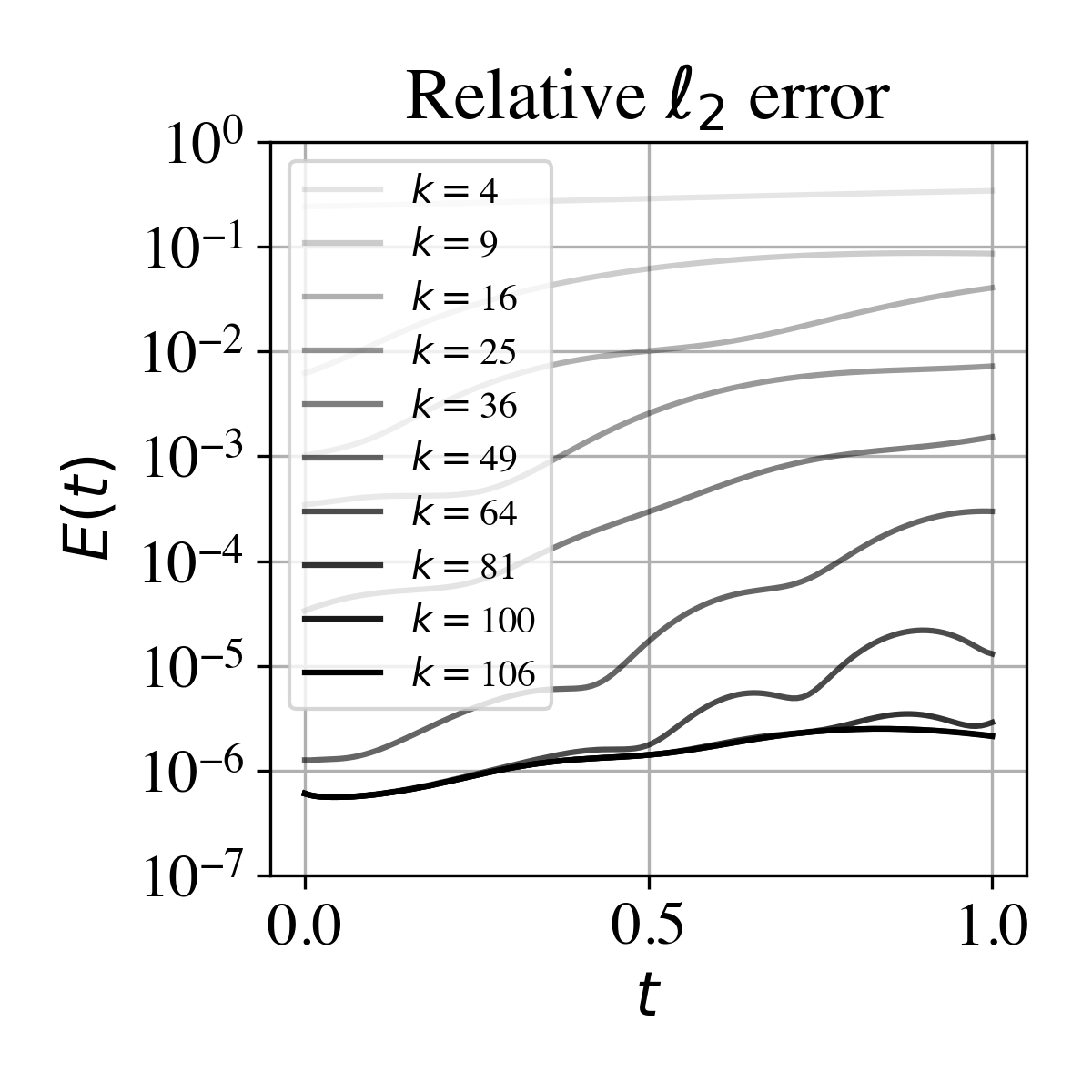}
    \includegraphics[width=0.45\textwidth]{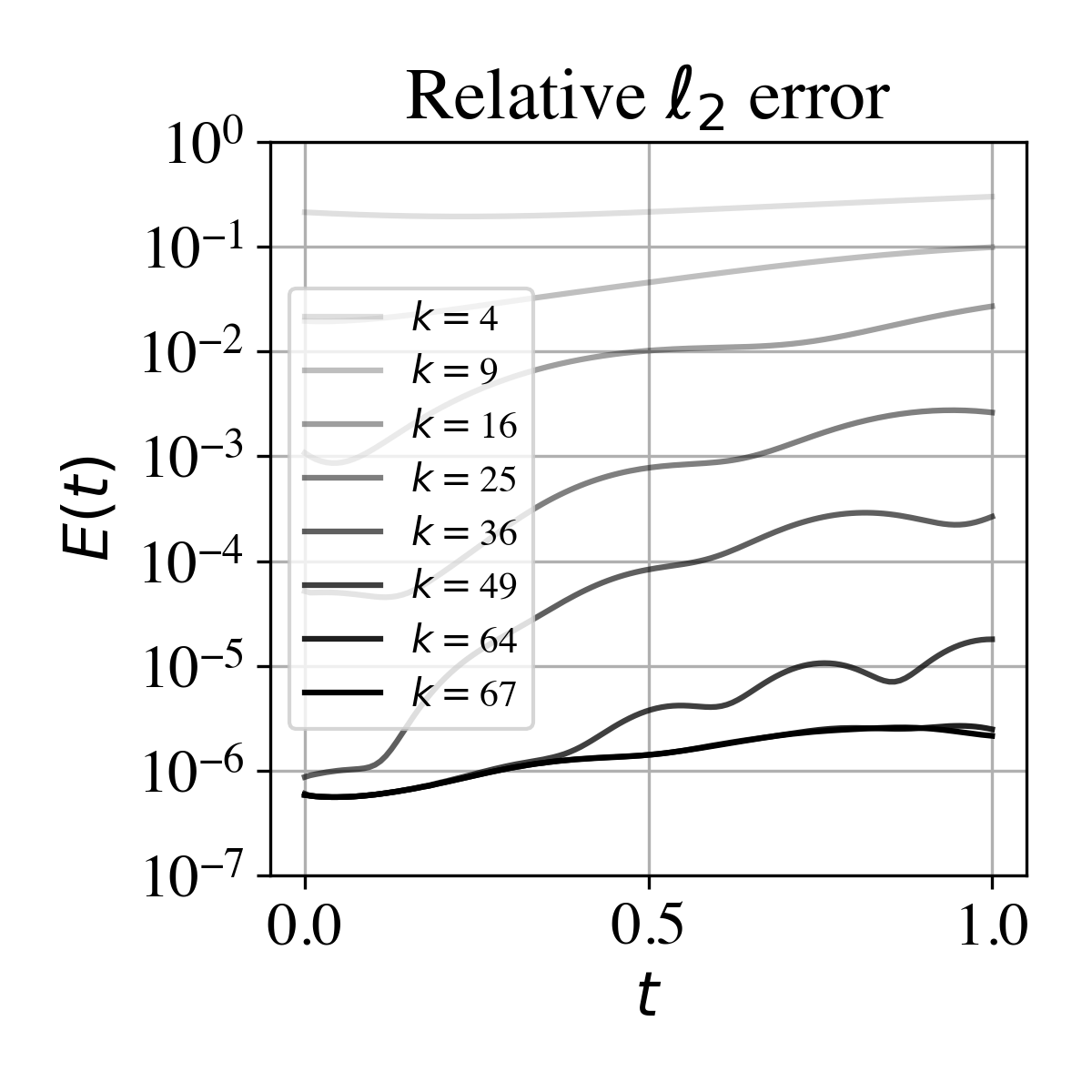}
    \includegraphics[width=0.45\textwidth]{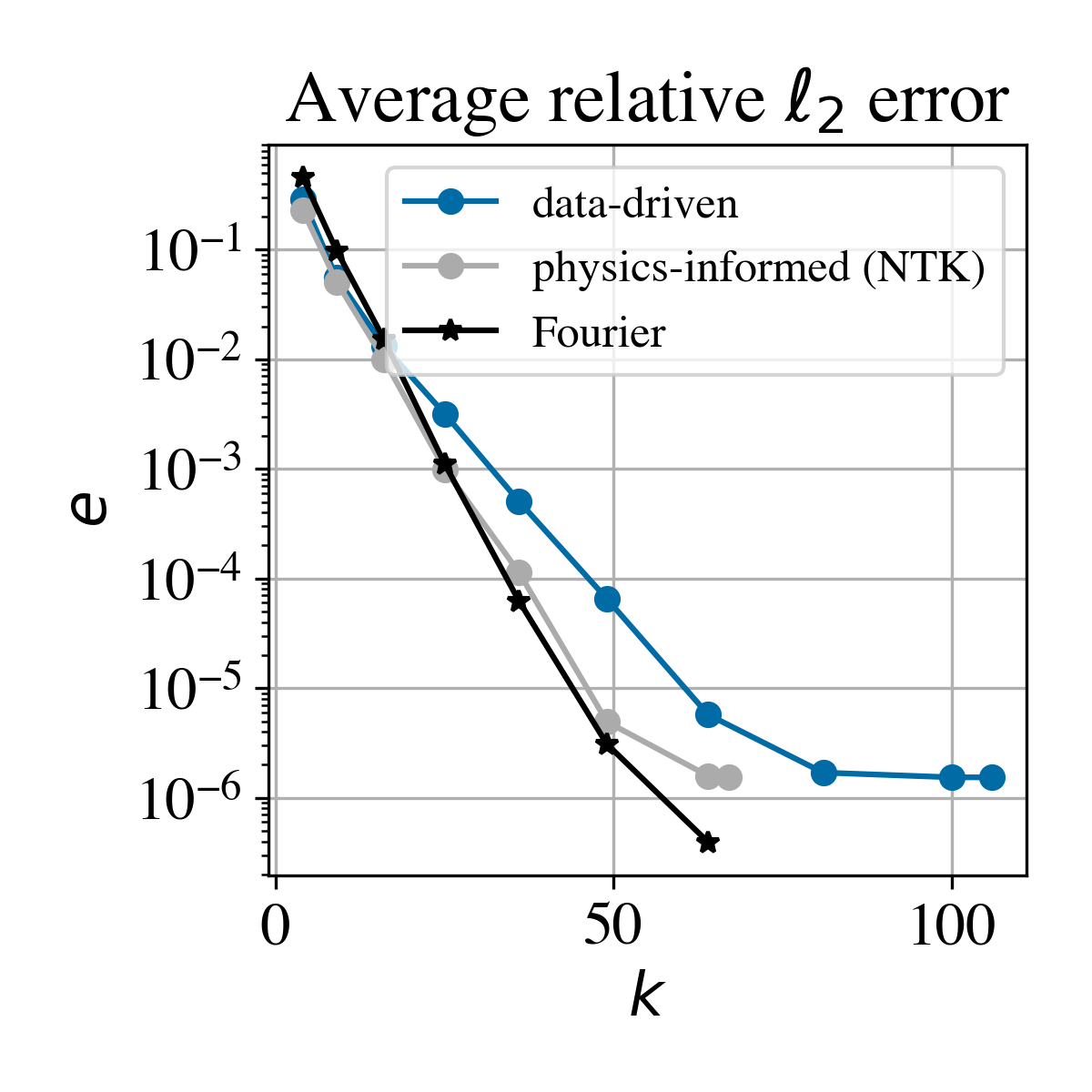}
    \caption{(Top left) Relative data-driven, (top right) relative physics-informed, and (bottom) average relative $\ell_2$ test sample errors for random test sample initial condition with $k$ custom basis functions for viscous Burgers with $\nu = 0.1$.}
    \label{fig:viscous-burgers-error-plots}
\end{figure}

We can use these basis functions in evolving the approximation using a spectral approach, with 106 data-driven custom basis functions and 67 physics-informed custom basis functions. The approximation fields are shown in Figure \ref{fig:viscous-burgers-spectral}. Figure \ref{fig:viscous-burgers-error-plots} shows the relative $\ell_2$ errors and average relative $\ell_2$ error for the data-driven and physics-informed custom basis functions as we approach the cutoff in the singular value decay, where the errors eventually saturate. The decay of the error for the physics-informed model with NTK-guided weights is shown to be comparable to the Fourier method. Shown in Table \ref{tab:viscous-burgers-spectral-error}, we achieve the same magnitude of the average relative $\ell_2$ error of $10^{-6}$ with 106 data-driven custom basis functions and only 67 physics-informed custom basis functions.

\clearpage

These results show that improving physics-informed DeepONets training, such as by using adaptive weighting like that based on NTK, could have model reduction benefits for smooth problems. Moreover, for nonlinear problems, enforcing the physics during training does have a substantial impact. These benefits are evident through assessing the decay of the singular value spectrum as well as the drop in the values of the expansion coefficients. For models that train well and require fewer degrees of freedom to capture the solution space, we can exploit the advantages in constructing a more robust collection of basis functions for use in a spectral method.

\begin{table}[H]
    \centering
    \begin{tabular}{|c|c|c|}\hline
         & $p$ & Average relative $\ell_2$ error \\ \hline\hline
    data-driven & 106 & $1.537 \times 10^{-6}$ \\ \hline
    physics-informed (NTK) & 67 & $1.548 \times 10^{-6}$ \\ \hline
    \end{tabular}
    \caption{Average errors for independent test sample for viscous Burgers with $\nu = 0.1$.}
    \label{tab:viscous-burgers-spectral-error}
\end{table}

\subsubsection{$\nu = 0.001$ and $\nu = 0.0001$}
\label{sec:burgers}

\begin{figure}[!htb]
    \centering
    \includegraphics[width=0.32\textwidth]{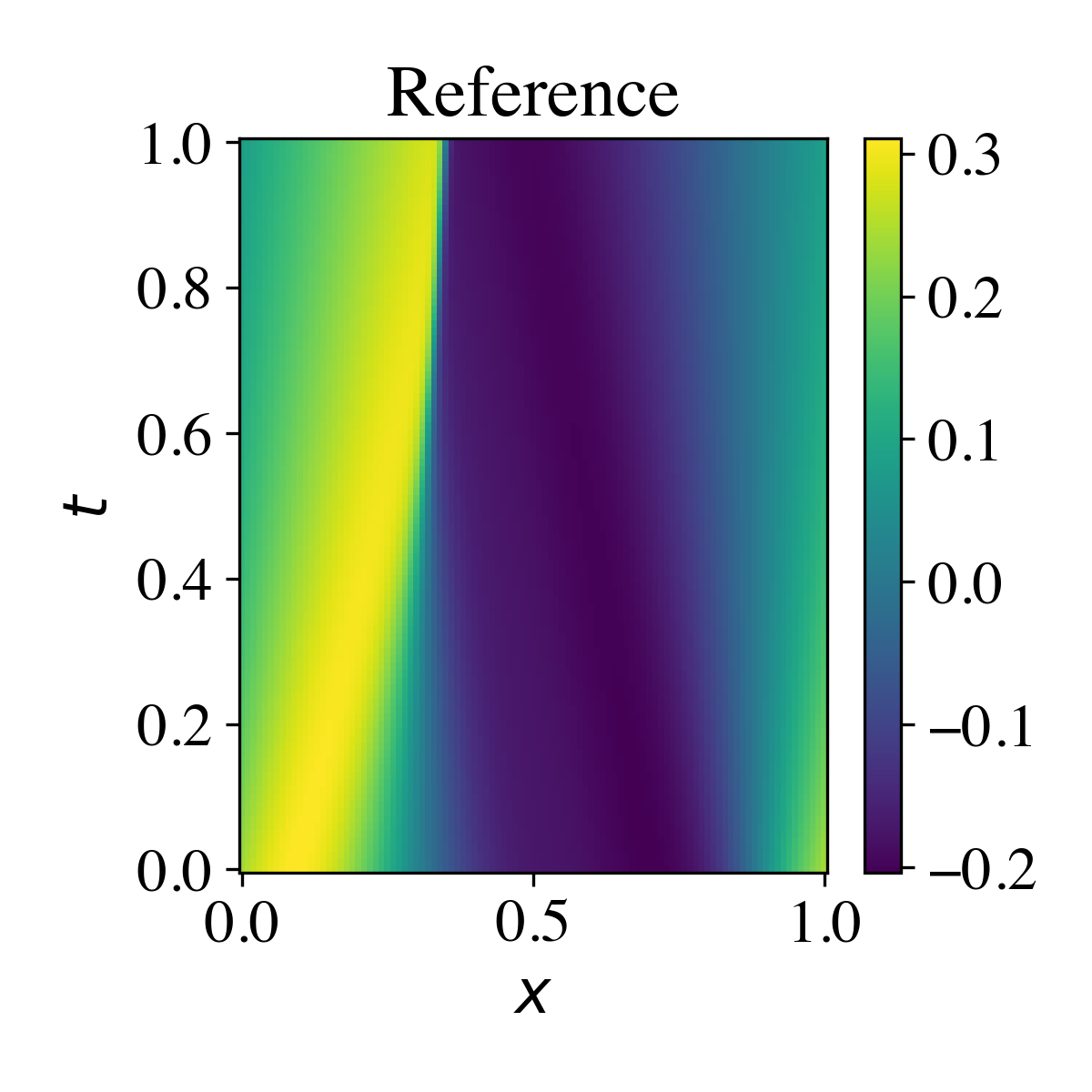}
    \includegraphics[width=0.32\textwidth]{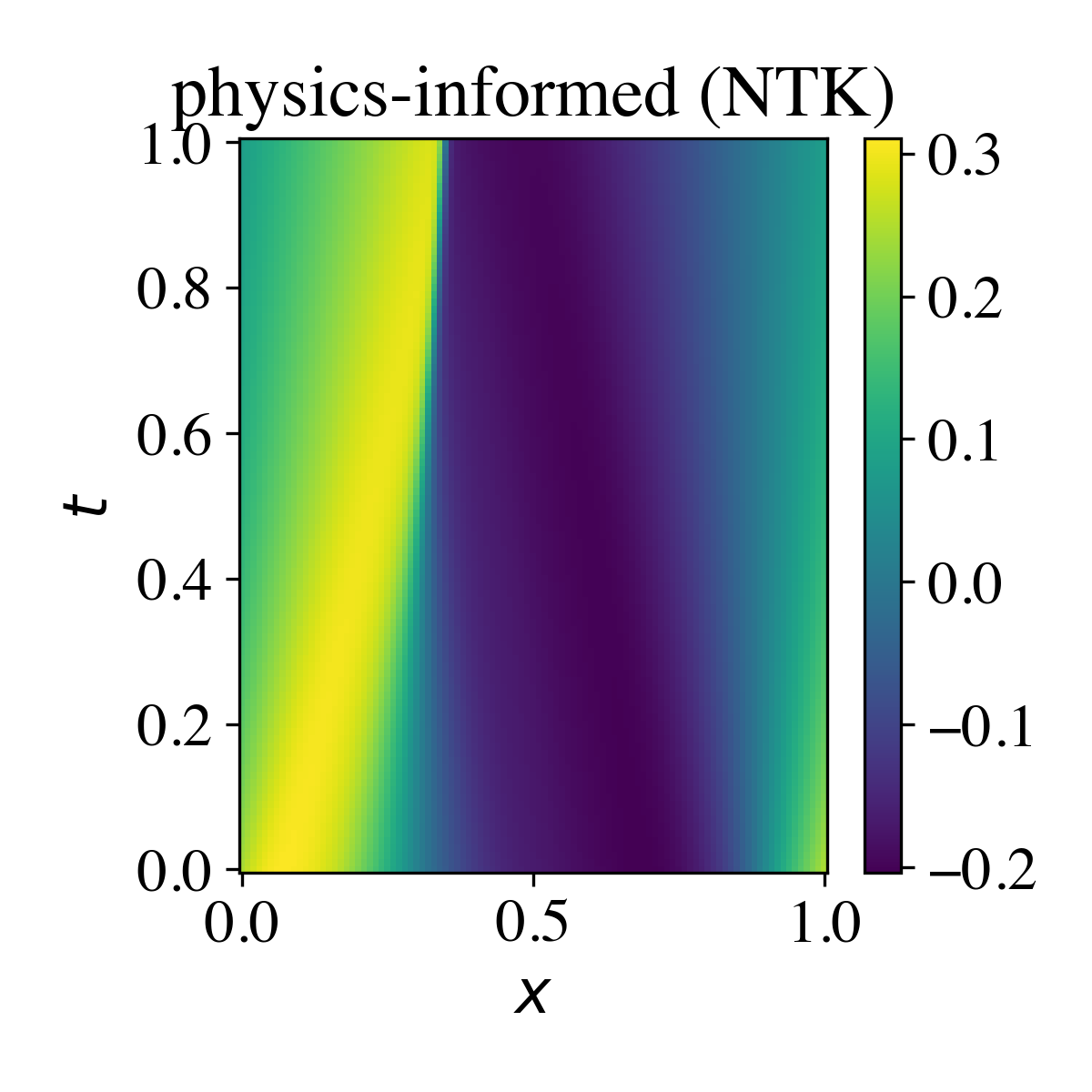}
    \caption{(Left) Numerical reference solution field and (right) physics-informed DeepONet prediction for viscous Burgers with $\nu = 0.001$.}
    \label{fig:viscous-burgers-benchmark-1e-3}
\end{figure}

Now we turn to viscous Burgers with a lower viscosity, where physics-informed DeepONets struggle to train well, particularly without the NTK \cite{wang_improved_2022}. Consider the viscous Burgers benchmark investigated in \cite{li_fourier_2021, wang_learning_2021} with initial conditions generated from a GRF $\sim \mathcal{N}(0, 25^2 (-\Delta + 5^2 I)^{-4}$) on $(x,t) \in (0,1) \times (0,1)$ with periodic boundary conditions. For calculating errors, the testing dataset generated using a Fourier spectral method from \cite{wang_improved_2022} is used. 
The models are trained using moderate local NTK adaptive weights with the modified DeepONet architecture \cite{wang_improved_2022}. First, consider viscosity $\nu = 0.001$. The physics-informed DeepONet prediction for this viscosity results in an average relative $\ell_2$ error of 4.48\% $\pm$ 6.12\% across 100 independent test samples. Figure \ref{fig:viscous-burgers-benchmark-1e-3} shows an independent test sample and the physics-informed DeepONet prediction.

\begin{figure}[!htb]
    \centering
    \includegraphics[width=0.32\textwidth]{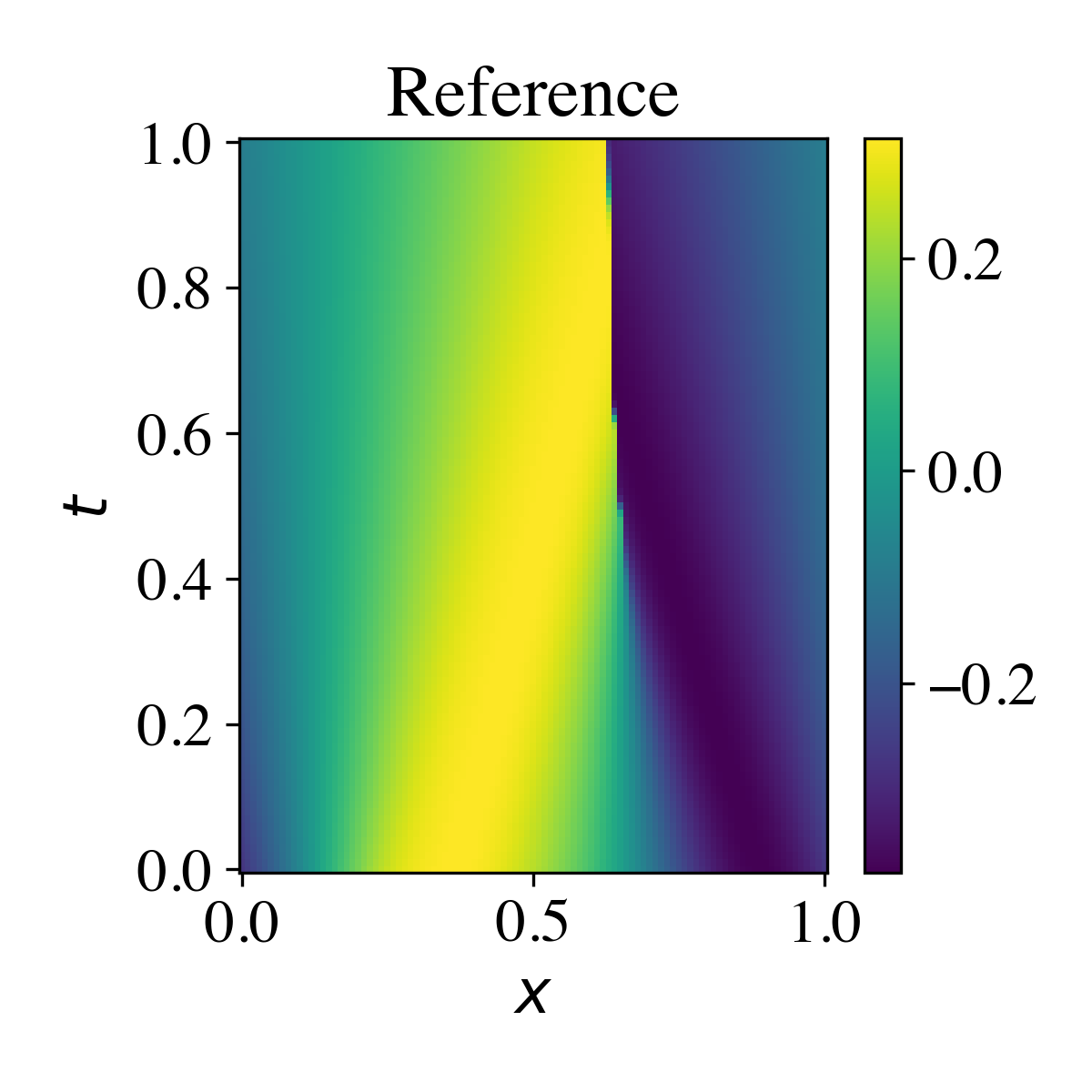}
    \includegraphics[width=0.32\textwidth]{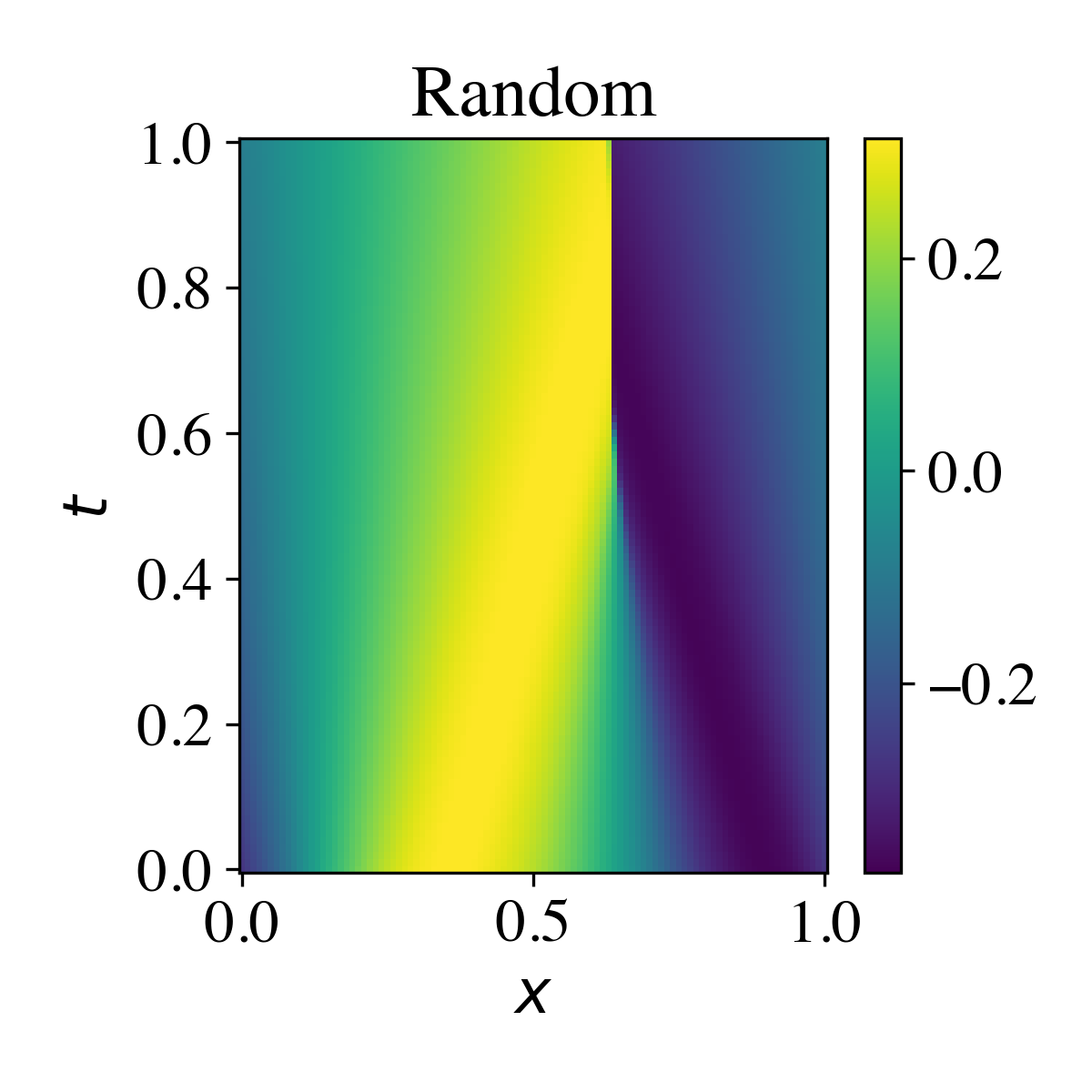}
    \includegraphics[width=0.32\textwidth]{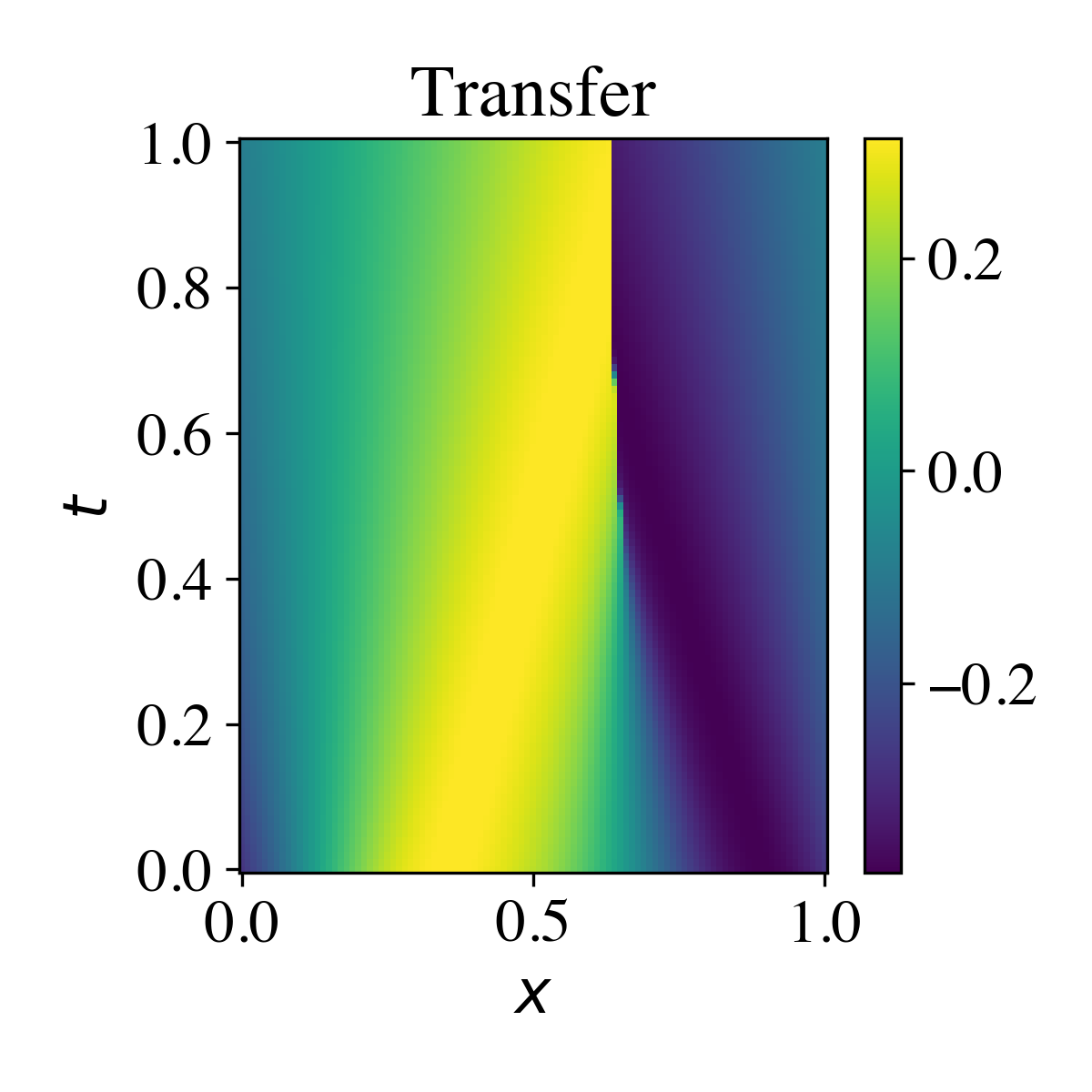}
    \caption{(Left) Numerical reference solution field, (middle) physics-informed using random initialization, and (right) physics-informed using transfer initialization DeepONet predictions for viscous Burgers with $\nu = 0.0001$.}
    \label{fig:viscous-burgers-deeponet}
\end{figure}

We can exploit the well-trained physics-informed DeepONet for $\nu = 0.001$ in initializing a DeepONet model designed to approximate a more difficult problem. Specifically, we effectively fine-tune the trained parameters for viscous Burgers with an even lower viscosity. Consider viscosity $\nu = 0.0001$. Initializing the physics-informed DeepONet from the trained $\nu = 0.001$ parameters (transfer initialization) results in an improved prediction compared to random initialization. Shown in Table \ref{tab:viscous-burgers-average-errors-1e-4-standard}, the average relative $\ell_2$ error from a physics-informed DeepONet initialized randomly for $\nu = 0.0001$ is over 13\%. When initializing the physics-informed DeepONet for solving viscous Burgers with $\nu = 0.0001$ with the trained parameters from $\nu = 0.001$, this error decreases to almost 7\%. Figure \ref{fig:viscous-burgers-deeponet} shows the numerical reference solution field with the two physics-informed DeepONet predictions (random initialization and transfer initialization).

\begin{table}[!htb]
    \centering
    \begin{tabular}{|c|c|c|}\hline
         Initialization & Average relative $\ell_2$ error \\ \hline\hline
    Random (NTK) & 13.67\% $\pm$ 7.28\% \\ \hline
    Transfer (NTK) from $\nu = 0.001$ & 7.03\% $\pm$ 4.94\% \\ \hline
    \end{tabular}
    \caption{Average errors across 100 test samples for viscous Burgers with $\nu = 0.0001$.}
    \label{tab:viscous-burgers-average-errors-1e-4-standard}
\end{table}

\begin{figure}[!htb]
    \centering
    \includegraphics[width=0.45\textwidth]{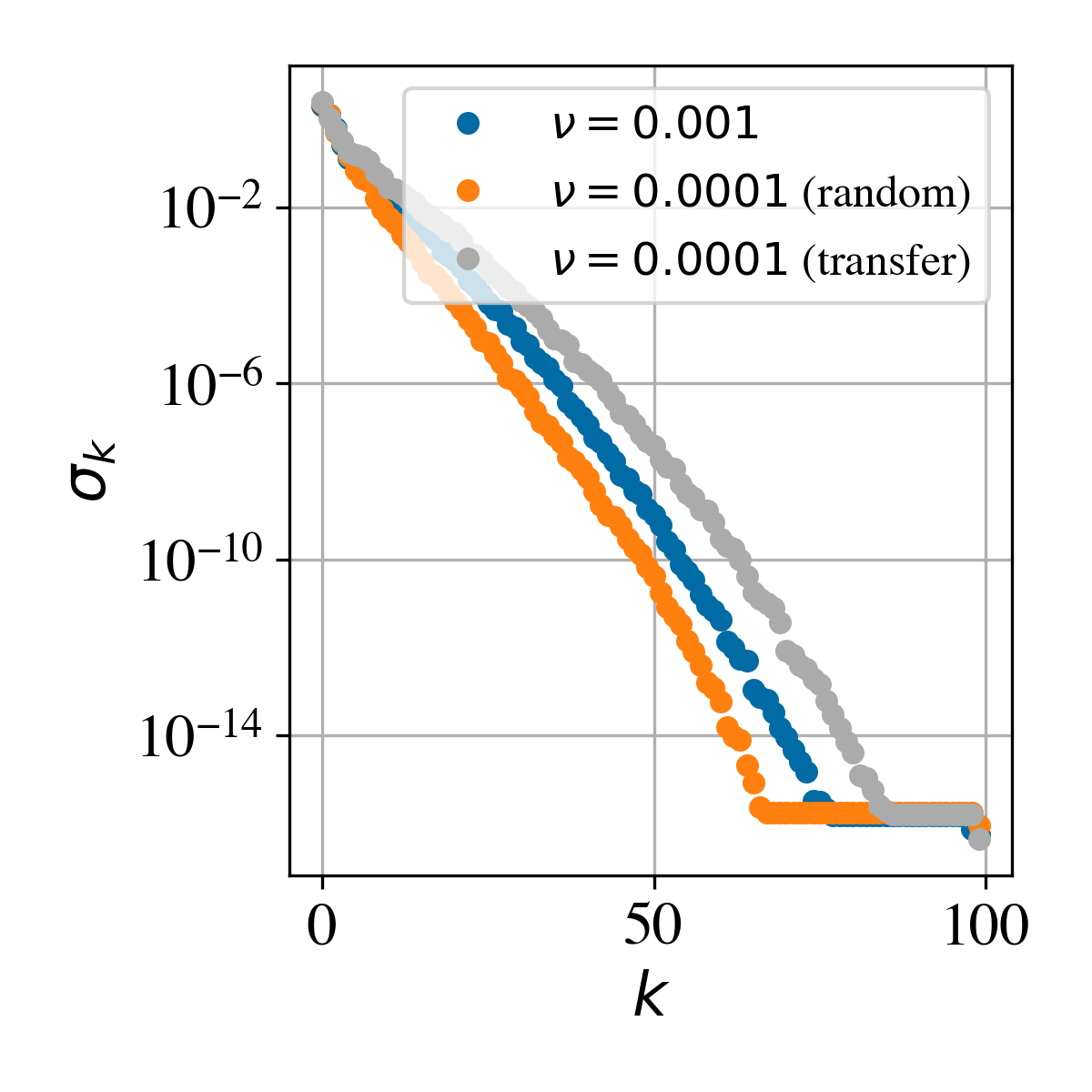}
    \includegraphics[width=0.45\textwidth]{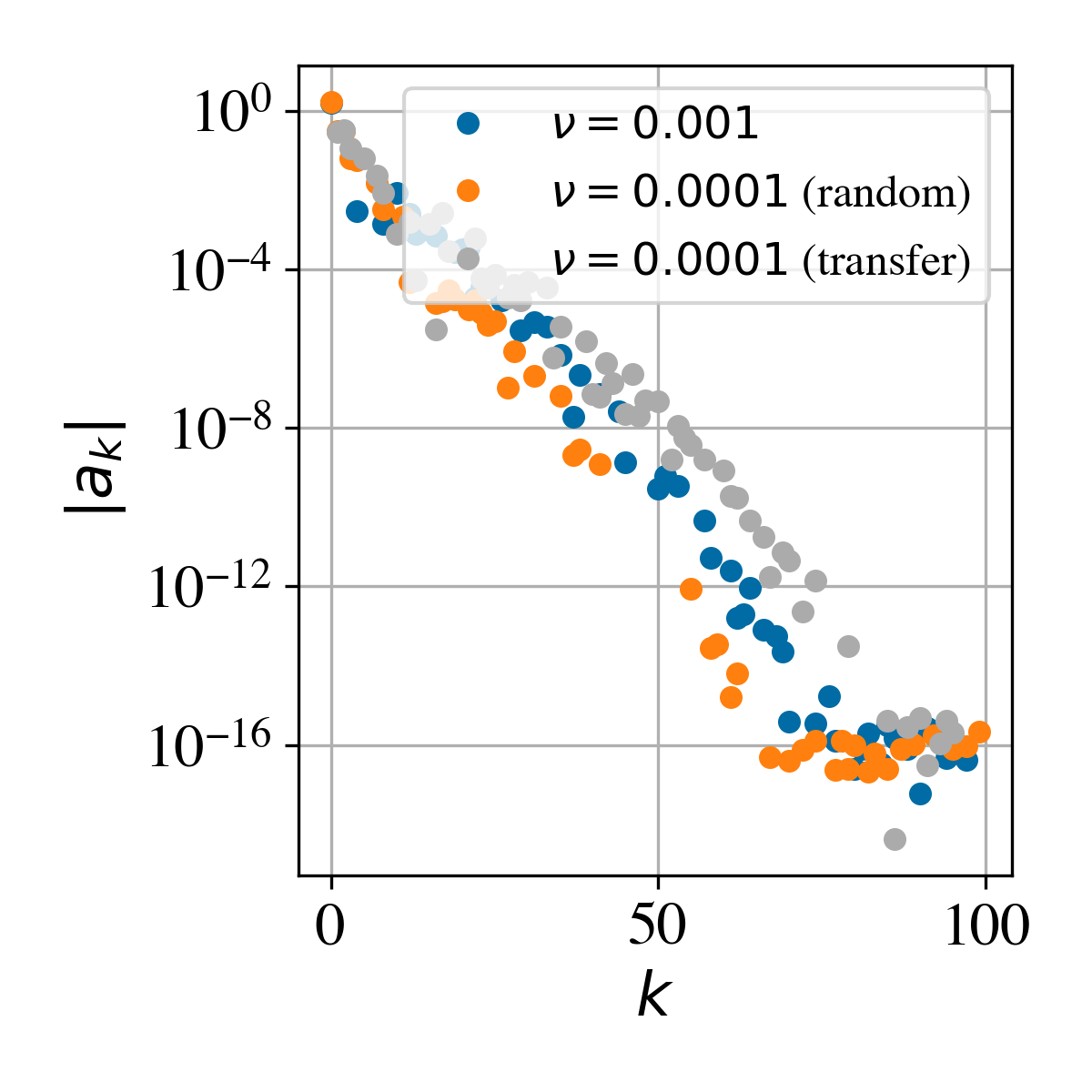}
    \caption{(Left) Singular values and (right) expansion coefficients for physics-informed DeepONets for different viscosities and initializations for $e^{\sin(x)}$ for viscous Burgers on $x \in (0,1)$.}
    \label{fig:viscous-burgers-benchmark-B-a-k}
\end{figure}

Figure \ref{fig:viscous-burgers-benchmark-B-a-k} shows the decay of the singular values and the expansion coefficients for the physics-informed DeepONets, with $\nu = 0.001$ as well as $\nu = 0.0001$ with random and transfer initialization. The expansion coefficients of all the trained models decay to machine precision, again showing that the models are training effectively. The decay of the singular values show the degrees of freedom that are important in capturing the solution space. For non-smooth problems, there is often more information (i.e., more basis functions) needed to learn this solution space. The physics-informed model for $\nu = 0.0001$ trained from the random initialization does not include all the modes required to represent this space, evidenced by the lower rank compared with the $\nu = 0.001$ problem. When the $\nu = 0.0001$ problem is initialized from the trained parameters from the $\nu = 0.001$ problem, we see the decay of the singular values being slower, meaning more basis functions are required to contain the information necessary to represent the solution.

\begin{figure}[!htb]
    \centering
    \includegraphics[width=\textwidth]{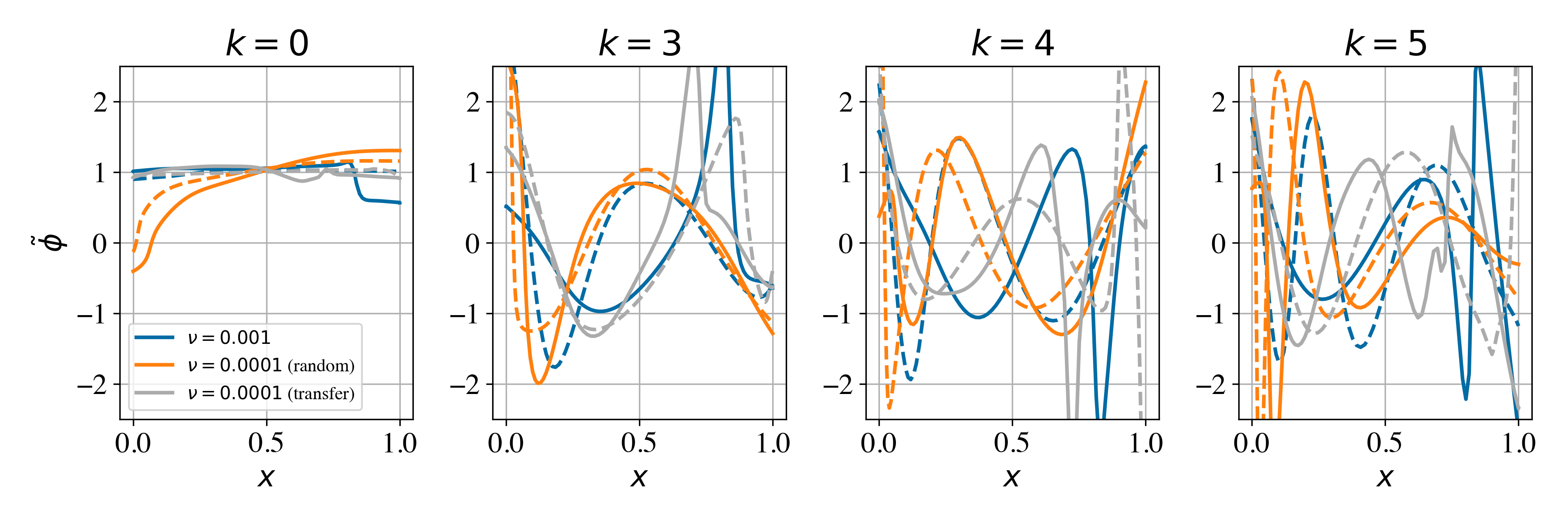}
    \caption{Custom basis functions, plotting with consistent boundaries $\Tilde{\phi}(x = 0 \text{ or } 1)$ (dashed: frozen-in-time at $t=1$) for viscous Burgers on $x \in (0,1)$.}
    \label{fig:viscous-burgers-benchmark-basis}
\end{figure}

A sample of the custom basis functions is shown in Figure \ref{fig:viscous-burgers-benchmark-basis}. Following \cite{meuris_machine-learning-based_2023}, the custom basis functions are constructed at both $t=0$ (solid) and $t=1$ (dashed), i.e., ``frozen-in-time'' at the beginning (initial) and end (final) of the temporal domain. If the physics-informed DeepONet is effective in representing the solution space, we would expect the basis functions to change in time. For the $\nu = 0.001$ model, we see substantial shift between the initial and final basis functions across the spatial domain. For the $\nu = 0.0001$ model that is randomly initialized, the initial and final basis functions do not show significant shift in the spatial domain. However, for the $\nu = 0.0001$ with transfer initialization, there is noticeable shift away from the starting parameters from $\nu = 0.001$, toward functions that better represent the harder problem of $\nu = 0.0001$. This shows the underlying fine-tuning process performed by the DeepONet in refining the parameters for the $\nu = 0.0001$ problem. The first 10 custom basis functions can be found in Appendix \ref{sec:burgers-basis-1e-4}.

\subsection{Korteweg-de Vries}

\begin{figure}[!htb]
    \centering
    \includegraphics[width=0.32\textwidth]{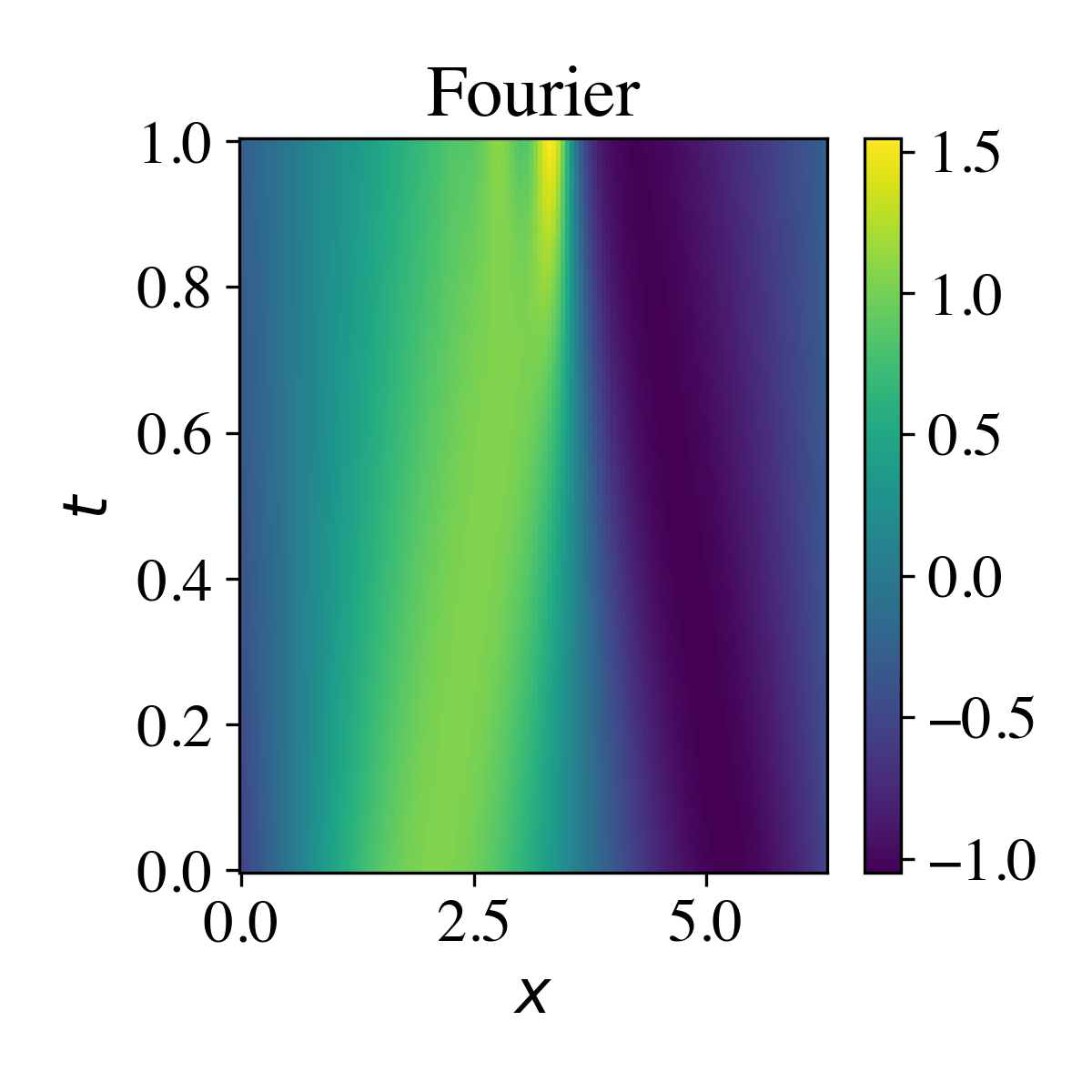}
    \includegraphics[width=0.32\textwidth]{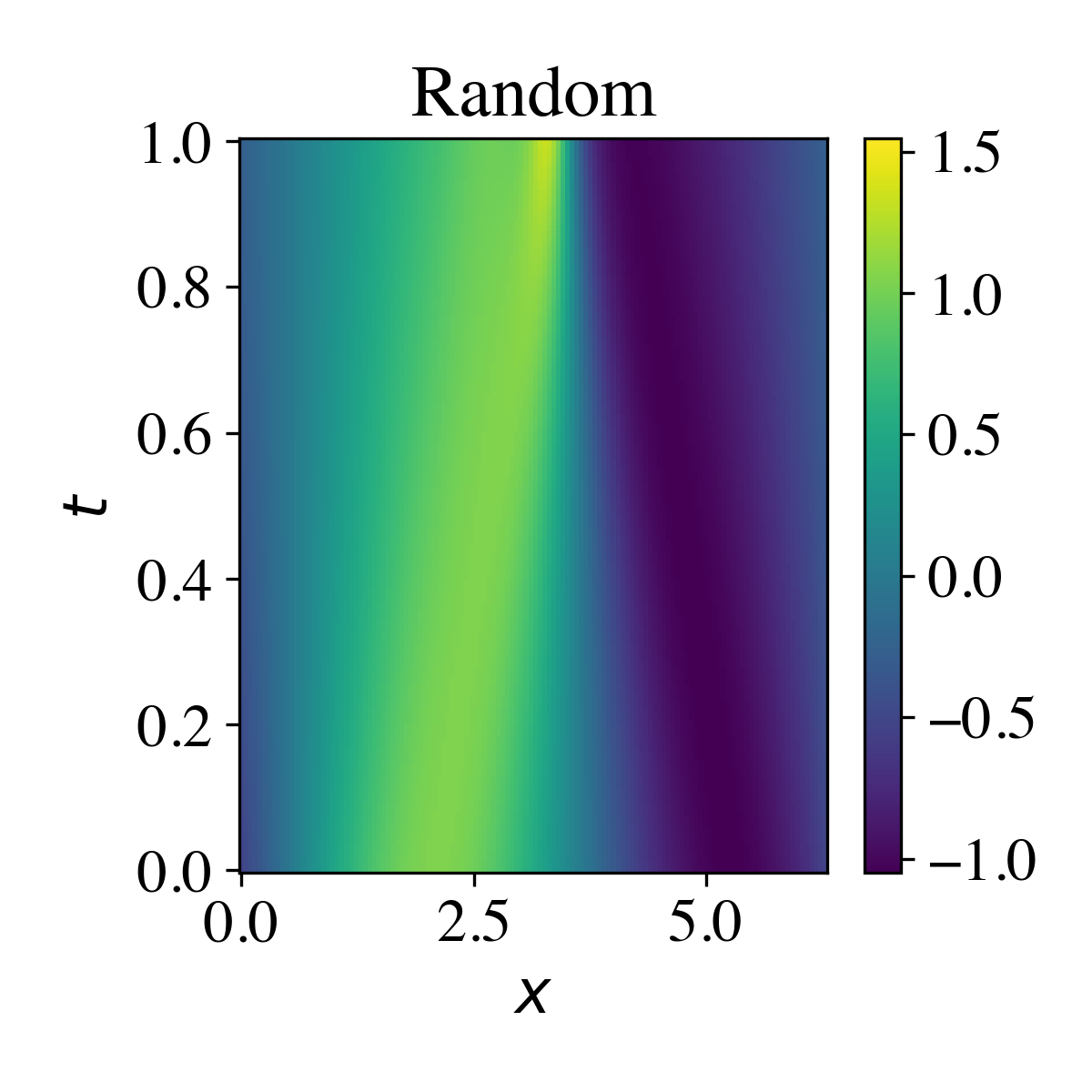}
    \includegraphics[width=0.32\textwidth]{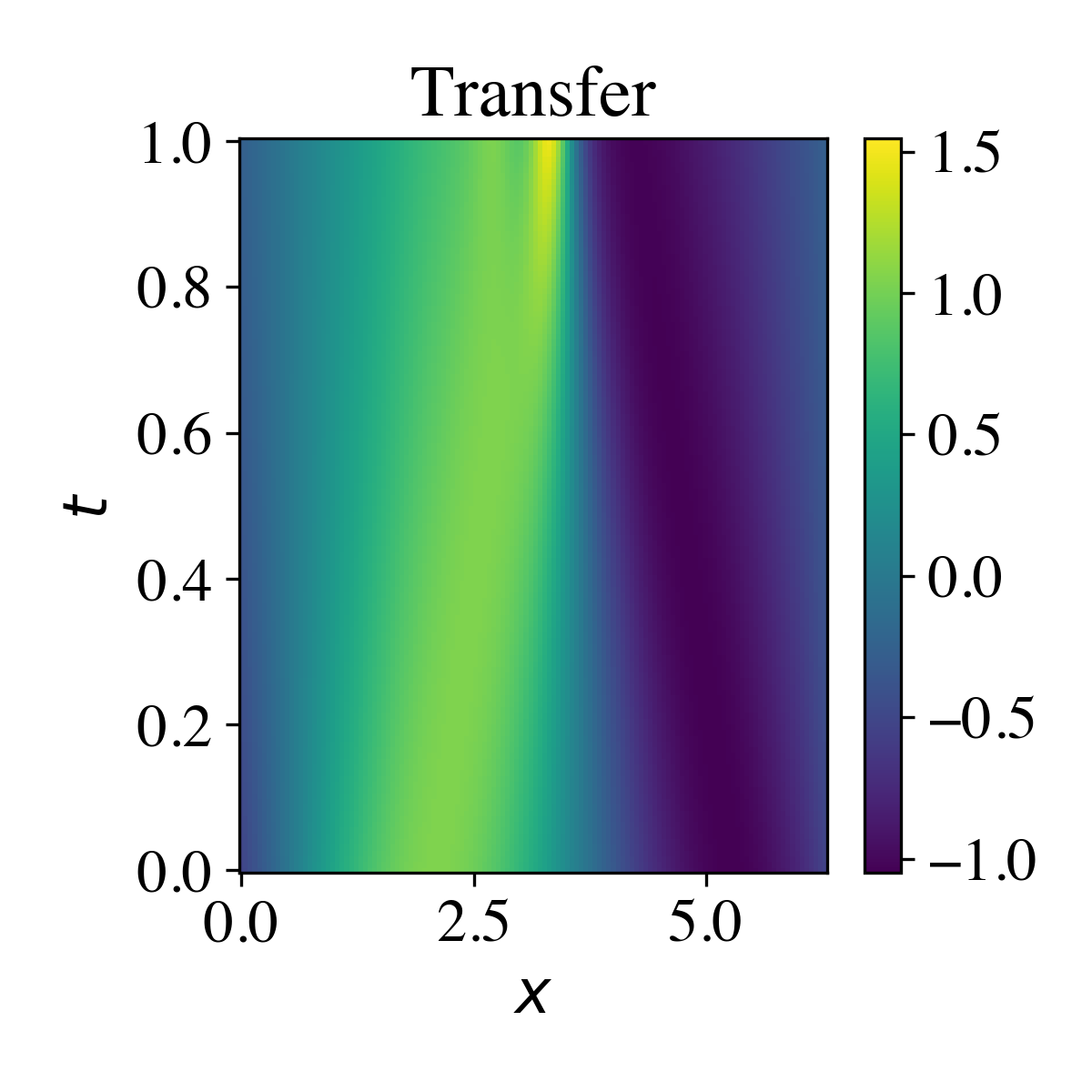}
    \caption{(Left) Fourier reference solution field, (middle) physics-informed using random initialization, and (right) physics-informed using transfer initialization DeepONet predictions for Korteweg-de Vries.}
    \label{fig:kdv-fourier}
\end{figure}

Lastly, we use transfer learning for initializing physics-informed DeepONets across different, but related, PDEs. Consider the Korteweg-de Vries equation
\begin{align}
    \frac{\partial s}{\partial t} + s \frac{\partial s}{\partial x} + \delta^2 \frac{\partial^3 s}{\partial x^3} = 0 
\end{align}
on $(x,t) \in (0,2\pi) \times (0,1)$ with periodic boundary conditions. The input to the branch network is the initial condition $s(x,0) = u(x)$. The solution field is computed using a Fourier pseudo-spectral method. The strength of dispersion coefficient is $\delta = 0.1$. The data-driven and physics-informed DeepONets use the modified DeepONet architecture \cite{wang_improved_2022}. The physics-informed DeepONet can be initialized from the trained physics-informed parameters from viscous Burgers (transfer) on the same domain. Note that this domain is different than in Sec. \ref{sec:burgers}. The domain in Section \ref{sec:burgers} was chosen to provide a comparison with existing papers \cite{wang_improved_2022, wang_learning_2021}. The relationship between viscous Burgers and Korteweg-de Vries is in the strengths of the viscous term and dispersive term. The physics-informed DeepONets use local CK adaptive weights. For this problem, CK weights demonstrated better performance compared to NTK weights. Figure \ref{fig:kdv-fourier} shows the reference solution field generated by the Fourier method for one independent test sample initial condition and the predictions from the physics-informed DeepONets initialized randomly and initialized with the trained parameters from the viscous Burgers $\nu = 0.0001$ problem. The viscous Burgers parameters for $\nu = 0.0001$ are themselves transfer initialized from the $\nu = 0.001$ problem with local CK adaptive weights. Shown in Table \ref{tab:average-kdv-errors}, the average relative $\ell_2$ error when randomly initialized is 3.92\%. When initializing the physics-informed DeepONet for solving Korteweg-de Vries with the trained parameters from viscous Burgers, this error decreases to 3.29\% across 100 test samples. The larger standard deviations point to increased variability in the predictions from the physics-informed model.

\begin{table}[!htb]
    \centering
    \begin{tabular}{|c|c|c|}\hline
       Initialization  & Average relative $\ell_2$ error \\ \hline\hline
    Random (CK) & 3.92\% $\pm$ 6.80\% \\ \hline
    Transfer (CK) from viscous Burgers with $\nu = 0.0001$ & 3.29\% $\pm$ 5.94\% \\ \hline
    \end{tabular}
    \caption{Average errors across 100 test samples for Korteweg-de Vries from transfer initialized viscous Burgers with $\nu = 0.0001$.}
    \label{tab:average-kdv-errors}
\end{table}

\begin{figure}[!htb]
    \centering
    \includegraphics[width=0.45\textwidth]{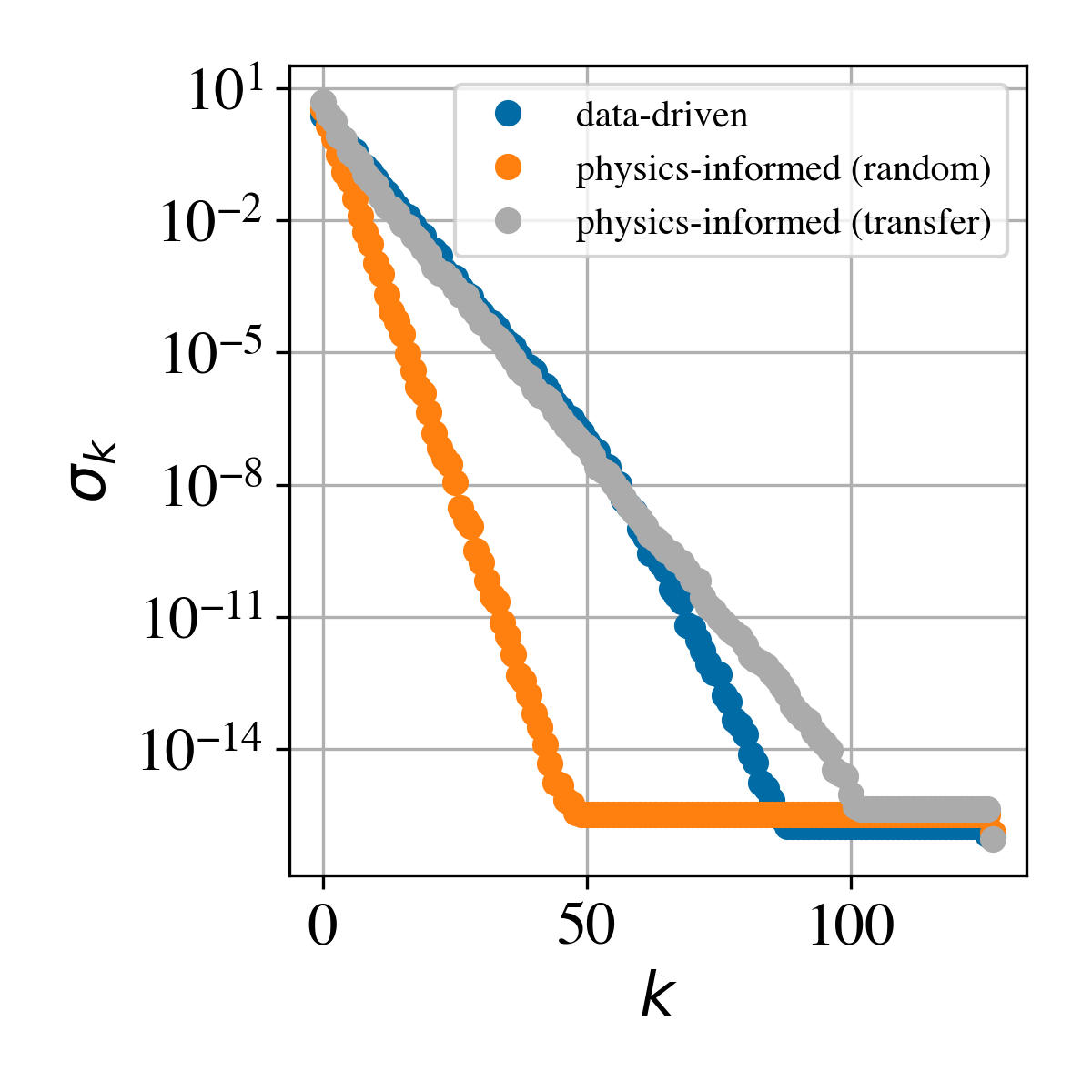}
    \includegraphics[width=0.45\textwidth]{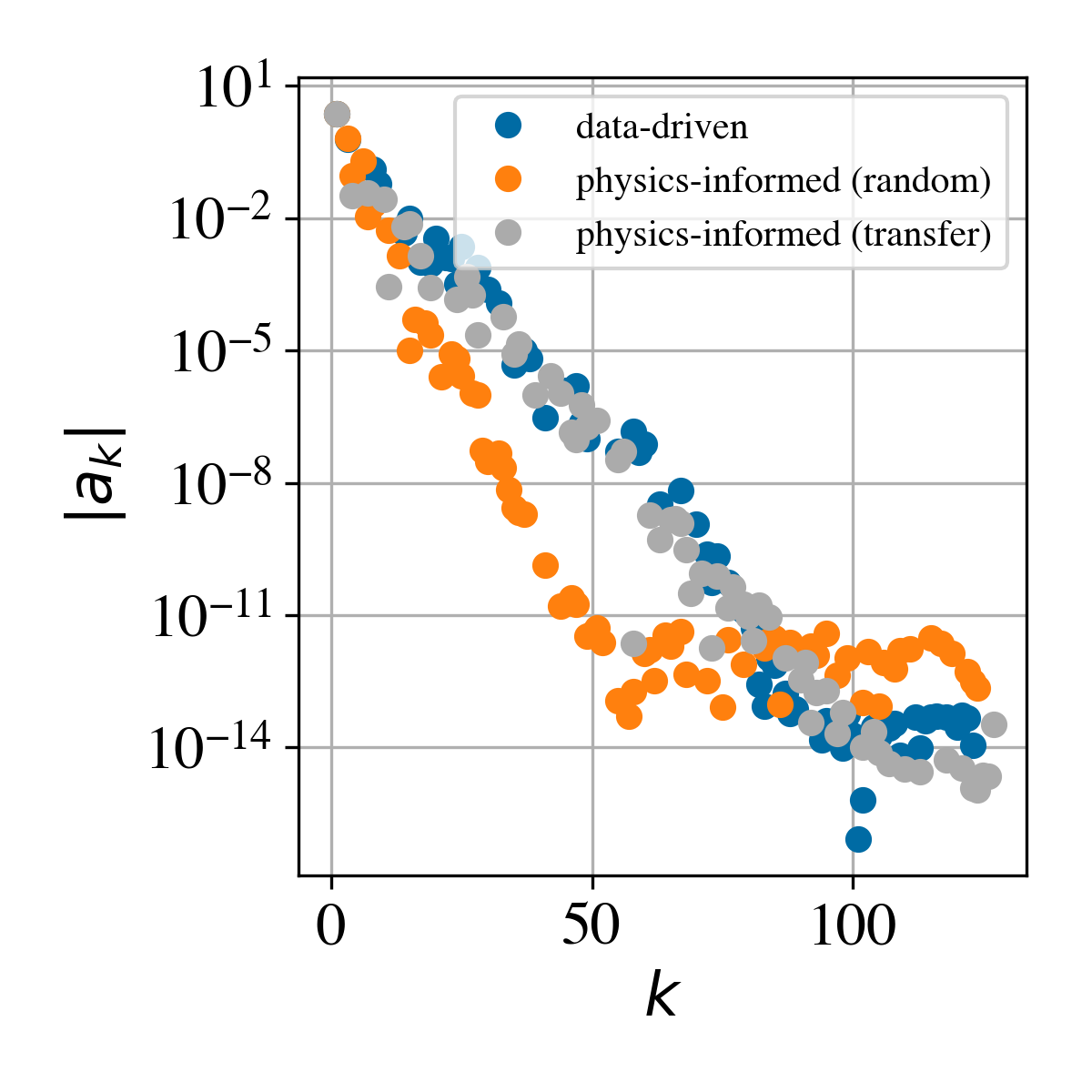}
    \caption{(Left) Singular values and (right) expansion coefficients for data-driven and physics-informed DeepONets for different initializations for $e^{\sin(x)}$ for Korteweg-de Vries.}
    \label{fig:kdv-B-a-k}
\end{figure}

Figure \ref{fig:kdv-B-a-k} shows the decay of the singular values and the expansion coefficients for the data-driven and physics-informed DeepONets, with random and transfer initialization. The expansion coefficients of the data-driven model and physics-informed model using the transfer initialization decay to machine precision, again showing that the models are training effectively. However, the expansion coefficients of the physics-informed model initialized randomly do not drop all the way to machine precision, meaning that it is not training effectively. The decay of the singular values show the degrees of freedom that are important in capturing the solution space. The singular values of the randomly initialized physics-informed model decays much faster, showing that it is not including all the modes necessary to represent the solution space. When initialized from the trained parameters from the viscous Burgers problem, we see the decay of the singular values being significantly slower, meaning more basis functions are required to capture the information necessary to represent the solution, and that the model is learning effectively.

\begin{figure}[!htb]
    \centering
    \includegraphics[width=\textwidth]{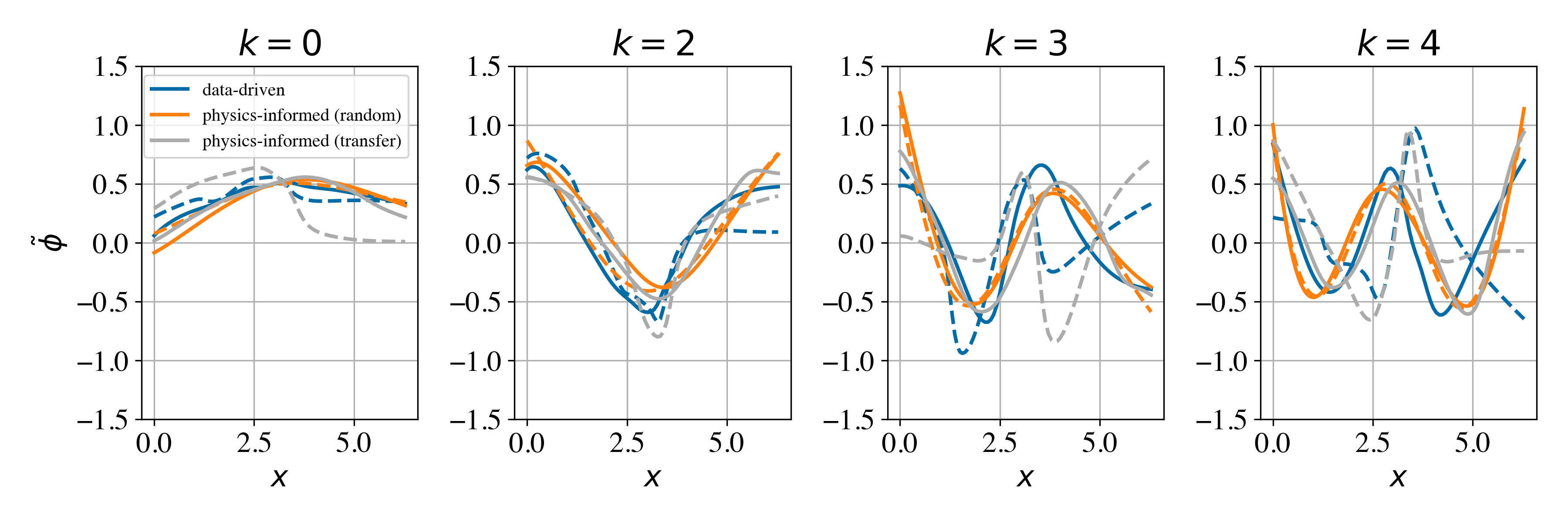}
    \caption{Custom basis functions, plotting with consistent boundaries $\Tilde{\phi}(x = 0 \text{ or } 2\pi)$ (dashed: frozen-in-time at $t=1$) for Korteweg-de Vries.}
    \label{fig:kdv-basis}
\end{figure}

A sample of the custom basis functions is shown in Figure \ref{fig:kdv-basis}. We expect the basis functions to change in time. For the data-driven custom basis functions, we see substantial shift between the initial and final basis functions across the spatial domain. For physics-informed model that is randomly initialized, the initial and final basis functions show almost no change in time. However, for the physics-informed model with transfer initialization, there is a substantial shift toward functions that better represent the Korteweg-de Vries problem, evidenced by the better agreement with the data-driven custom basis functions. This demonstrates the underlying fine-tuning process performed by the DeepONet across different PDEs. The first 10 custom basis functions can be found in Appendix \ref{sec:kdv-basis}.

For Korteweg-de Vries, randomly initializing the DeepONet parameters is shown to result in less effective basis functions in time and space. The decay of the singular values as well as the decay of the expansion coefficients needs to be considered when assessing how well a physics-informed DeepONet is learning. Finally, there are benefits to using transfer learning to initialize a DeepONet for a more difficult PDE from a related PDE, such as learning more effective basis functions that can be used in a spectral method.

\section{Conclusions}

In this work, we evaluated quantities of interest (e.g., singular values, expansion coefficients) toward assessing the performance of a physics-informed DeepONet. Further, we extracted basis functions from the trained models to develop further understanding toward the universality of the learning process of DeepONets. The basis functions extracted from the improved physics-informed DeepONets are shown to provide potential model reduction benefits when used to evolve the approximation using a spectral approach for advection-diffusion and viscous Burgers with $\nu = 0.1$. Lastly, we proposed a transfer learning approach for improving training for physics-informed DeepONets between parameters of the same PDE as well as across different, but related, PDEs. This approach was demonstrated on problems where physics-informed DeepONets struggle to train, including viscous Burgers with $\nu = 0.0001$, and resulted in significant reduction in error and more expressive learned basis functions. These results contribute to the promising methodology of discovering the solutions of more difficult problems starting from the solutions of easier, but related, problems \cite{howard_stacked_2024}. Future work includes further investigation toward the performance of physics-informed DeepONets and the learned basis functions using different adaptive weighting schemes.

\section*{Acknowledgements}

This project was completed with support from the U.S. Department of Energy, Advanced Scientific Computing Research program, under the Scalable, Efficient and Accelerated Causal Reasoning Operators, Graphs and Spikes for Earth and Embedded Systems (SEA-CROGS) project (Project No. 80278). The computational work was performed using PNNL Institutional Computing at Pacific Northwest National Laboratory. Pacific Northwest National Laboratory (PNNL) is a multi-program national laboratory operated for the U.S. Department of Energy (DOE) by Battelle Memorial Institute under Contract No. DE-AC05-76RL01830.\\

\noindent
This material is based upon work supported by the U.S. Department of Energy, Office of Science, Office of Advanced Scientific Computing Research, Department of Energy Computational Science Graduate Fellowship under Award Number DE-SC0023112. This report was prepared as an account of work sponsored by an agency of the United States Government. Neither the United States Government nor any agency thereof, nor any of their employees, makes any warranty, express or implied, or assumes any legal liability or responsibility for the accuracy, completeness, or usefulness of any information, apparatus, product, or process disclosed, or represents that its use would not infringe privately owned rights. Reference herein to any specific commercial product, process, or service by trade name, trademark, manufacturer, or otherwise does not necessarily constitute or imply its endorsement, recommendation, or favoring by the United States Government or any agency thereof. The views and opinions of authors expressed herein do not necessarily state or reflect those of the United States Government or any agency thereof.

\newpage
\appendix

\section{Training data and network parameters}
\label{sec:parameters}

\begin{figure}[H]
    \centering 
    \includegraphics[width=0.5\textwidth]{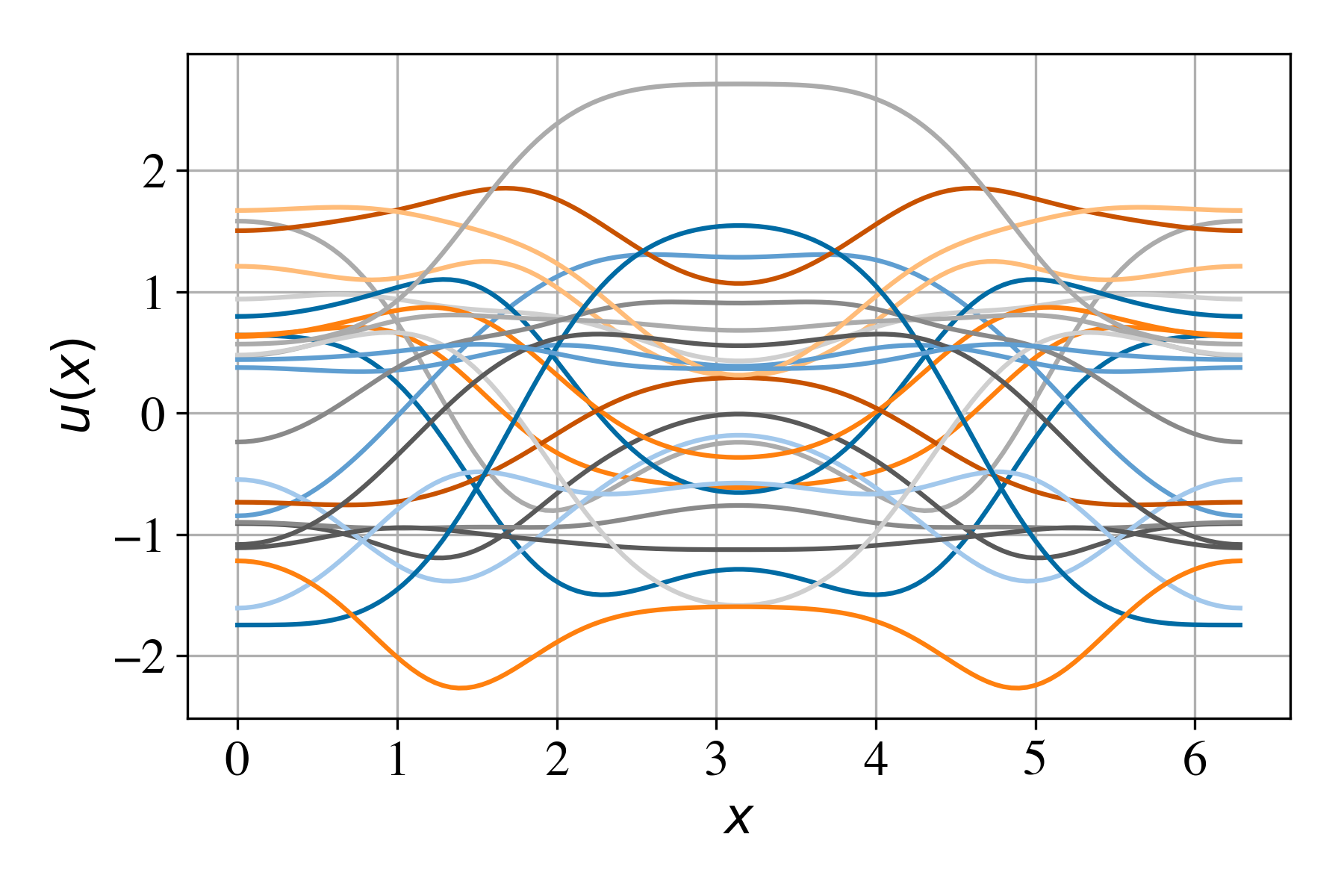}
    \caption{25 example functions sampled from the GRF, $f(\sin^2 (x/2))$ and $l = 0.5$.}
    \label{fig:grf}
\end{figure}

Physics-informed DeepONets are trained with the hyperbolic tangent activation function. The Adam optimizer with learning rate 0.001 is used with an exponential decay scheduler. Table \ref{tab:inputs} shows the input function spaces and domains for the selected test cases. For advection-diffusion, some function $f(x)$ is sampled from a mean zero Gaussian random field (GRF) with covariance kernel
\begin{align}
    \kappa_l (f(x_1), f(x_2)) = \exp\left( \frac{-||f(x_1) - f(x_2)||^2}{2l^2} \right)
\end{align}
for some length scale $l$ to generate the random training and testing inputs, as in \cite{meuris_machine-learning-based_2023}. Some sample inputs are plotted in Figure \ref{fig:grf}. Table \ref{tab:parameters} specifies the training hyperparameters for each test case, and Table \ref{tab:arch} shows the network sizes.

\begin{table}[H]
    \centering
    \begin{tabular}{|c|c|c|}\hline
          & Function/Kernel & Domain  \\ \hline\hline
        {Advection-diffusion}  & $f(\sin^2 (x/2)),\,\, l = 0.5$ & $x \in (0,2\pi)$ \\ \hline
        {Viscous Burgers}      & $\mathcal{N}(0, 25^2 (-\Delta + 5^2 I)^{-4})$ & $x \in (0,1)$ \\ \hline
        \multirow{3}{*}{Korteweg-de Vries} & \multirow{3}{*}{$c ( -a \sin(x) + b \cos(x) )$} & $a, b \in \mathcal{U}[0, 1)$ \\
                                           & & $c \in \mathcal{U}[-1,1)$ \\
                                           & & $x \in (0,2\pi)$\\ \hline
    \end{tabular}
    \caption{Input function space $u(x)$ \cite{meuris_machine-learning-based_2023, wang_respecting_2024}.}
    \label{tab:inputs}
\end{table}

\newpage
\begin{table}[H]
    \centering
    \begin{tabular}{|c|c|c|c|c|c|c|c|c|}\hline
         & $m$ & Train & Test & $P$ & Batch size & Epochs \\ \hline\hline
        {Advection-diffusion} & 128 & 500 & 100 & 128 & 1000 & 200000 \\ \hline
        {Viscous Burgers}  & 101 & 500 & 100 & 101 & 14000 & 200000 \\ \hline
        Korteweg-de Vries  & 128 & 500 & 100 & 128 & 16384 & 200000 \\ \hline
    \end{tabular}
    \caption{Hyper-parameter settings.}
    \label{tab:parameters}
\end{table}

\begin{table}[H]
    \centering
    \begin{tabular}{|c|c|c|c|c|}\hline
         & Trunk width & Trunk depth & Branch width & Branch depth \\ \hline\hline
        Advection-diffusion & 128 & 4 & 128 & 3 \\ \hline
        Viscous Burgers & 100 & 7 & 100 & 7 \\ \hline
        Korteweg-de Vries & 128 & 6 & 128 & 5 \\ \hline
    \end{tabular}
    \caption{Network sizes.}
    \label{tab:arch}
\end{table}

\section{Viscous Burgers $x \in (0,2\pi)$}
\label{sec:burgers_2pi}

\begin{figure}[!htb]
    \centering
    \includegraphics[width=0.32\textwidth]{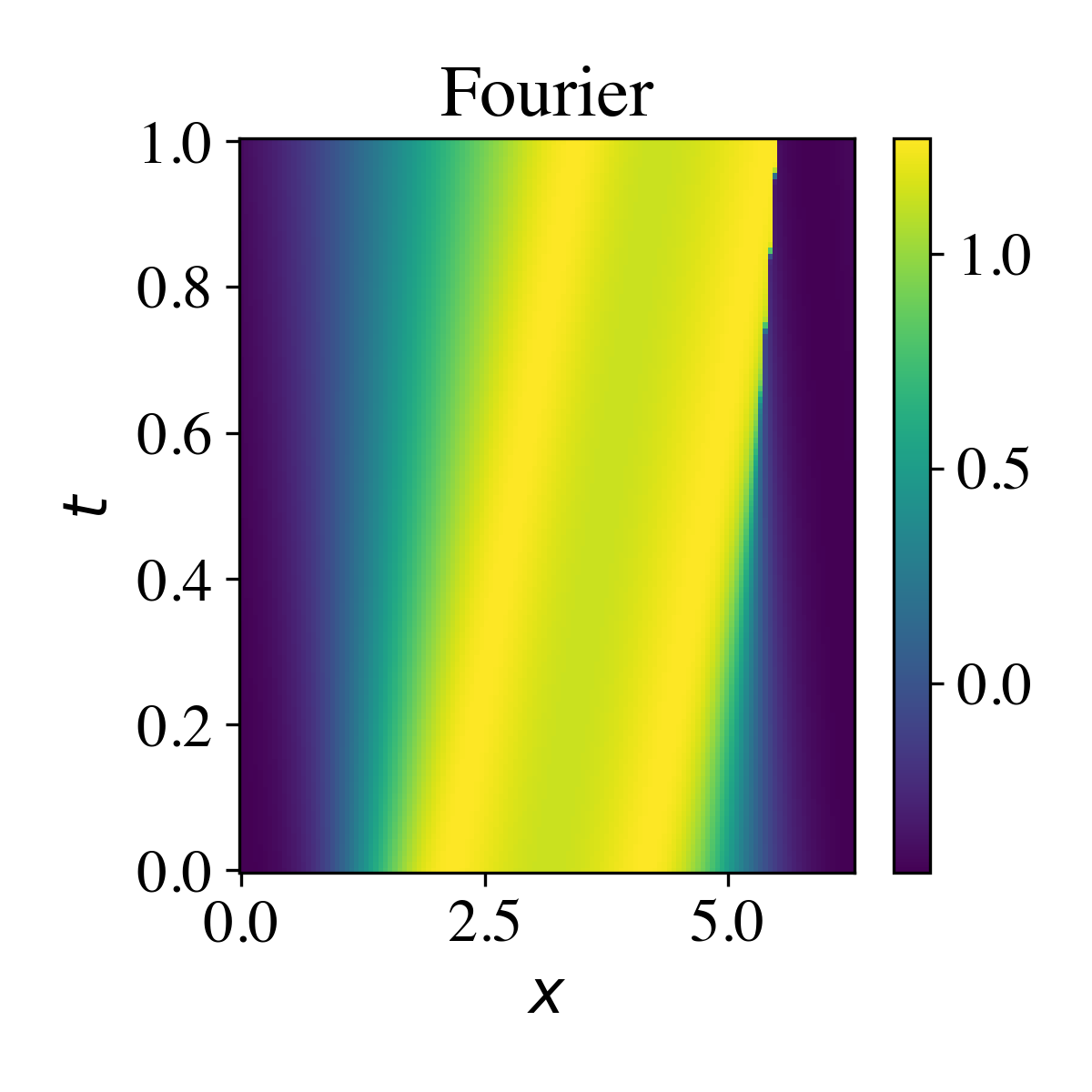}
    \includegraphics[width=0.32\textwidth]{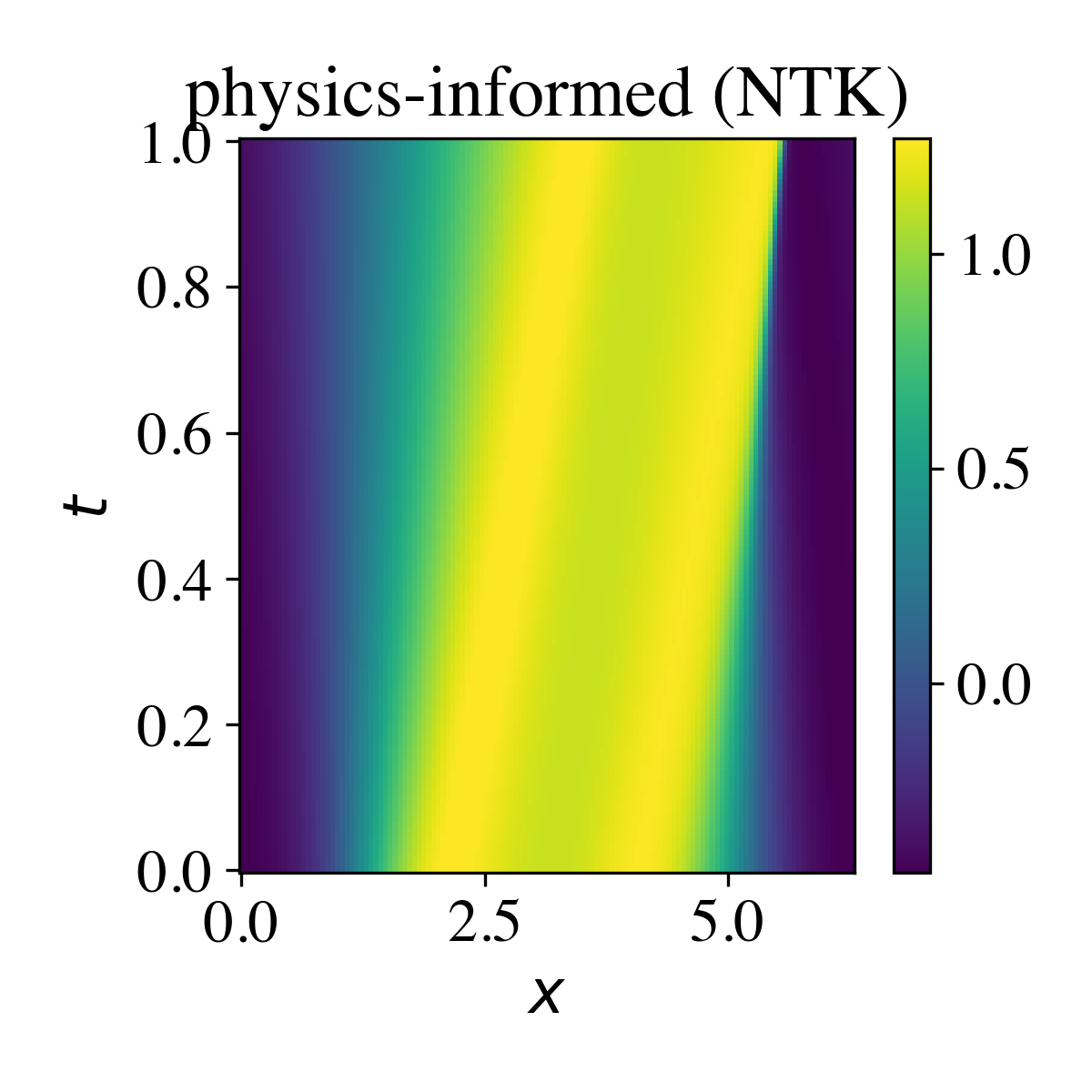}
    \caption{(Left) Fourier reference solution field and (right) physics-informed DeepONet prediction for $\nu = 0.001$.}
    \label{fig:viscous-burgers-1e-3}
\end{figure}

Consider viscous Burgers on the domain $x \in (0,2\pi)$. The initial conditions are generated from the GRF in Appendix \ref{sec:parameters} used for advection-diffusion. First, consider viscosity $\nu = 0.001$. The physics-informed DeepONet is trained with local NTK adaptive weights and results in an average relative $\ell_2$ error of $3.84\% \pm 3.34\%$ across 100 independent test samples computed using a Fourier pseudo-spectral method. Figure \ref{fig:viscous-burgers-1e-3} shows an independent test sample and the physics-informed DeepONet prediction. The widths of the trunk and branch nets are 128. The depths of the trunk and branch nets are 6 and 5, respectively, to enable transfer initialization to the Korteweg-de Vries problem.

\begin{figure}[!htb]
    \centering
    \includegraphics[width=0.4\textwidth]{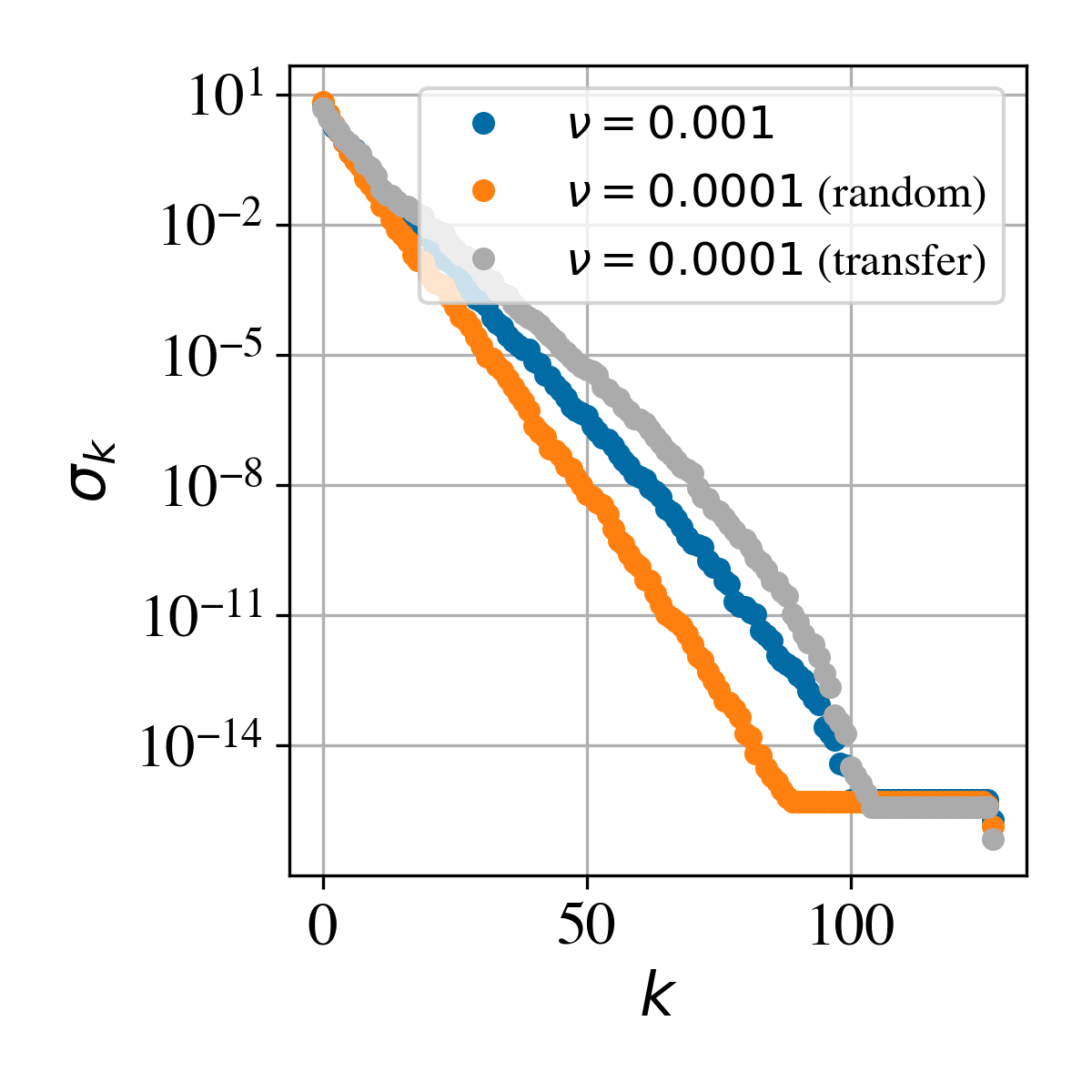}
    \includegraphics[width=0.4\textwidth]{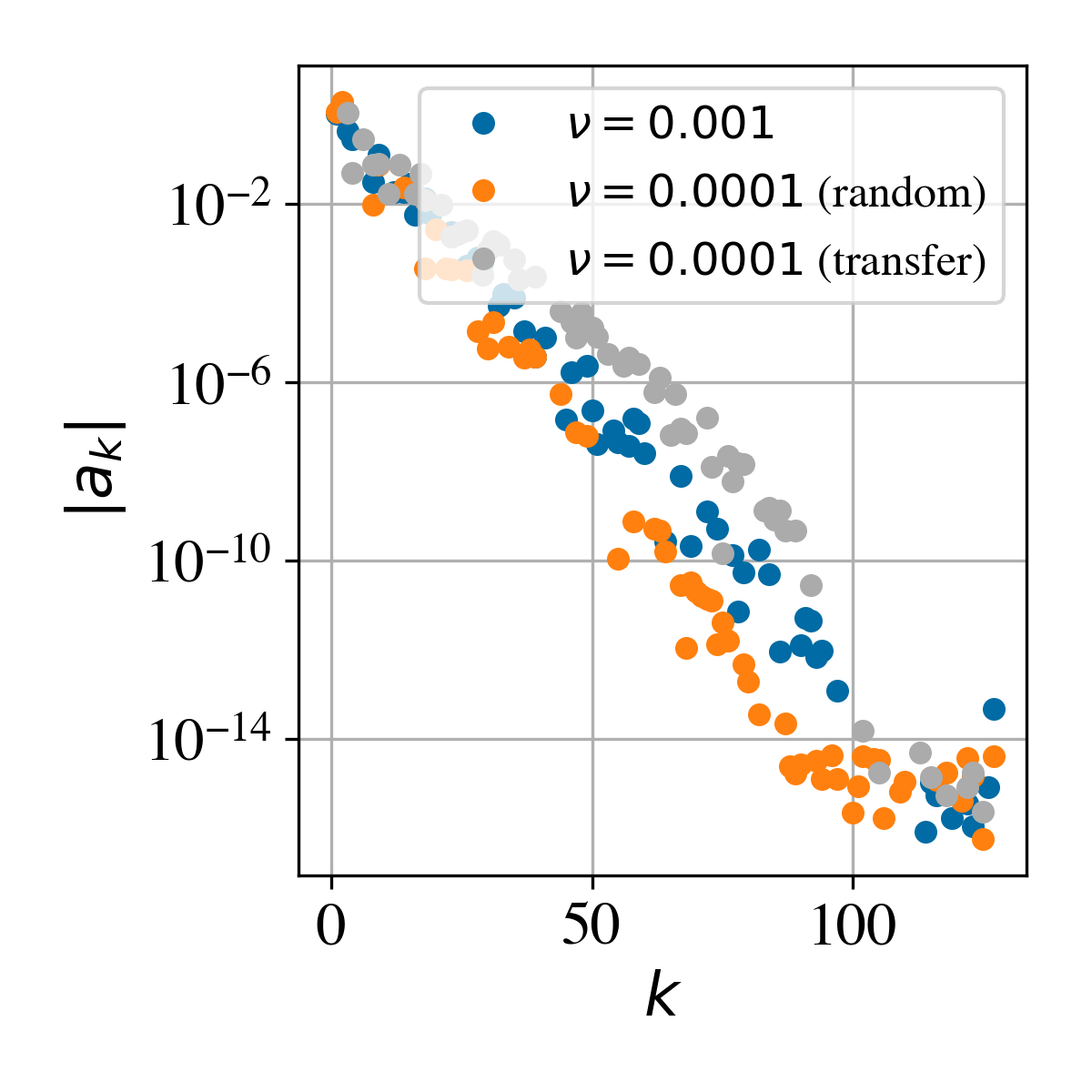}
    \caption{(Left) Singular values and (right) expansion coefficients for physics-informed DeepONets for different viscosities and initializations for $e^{\sin(x)}$ for viscous Burgers on $x \in (0,2\pi)$.}
    \label{fig:viscous-burgers-1e-4-B-a-k}
\end{figure}

\begin{figure}[!htb]
    \centering
    \includegraphics[width=\textwidth]{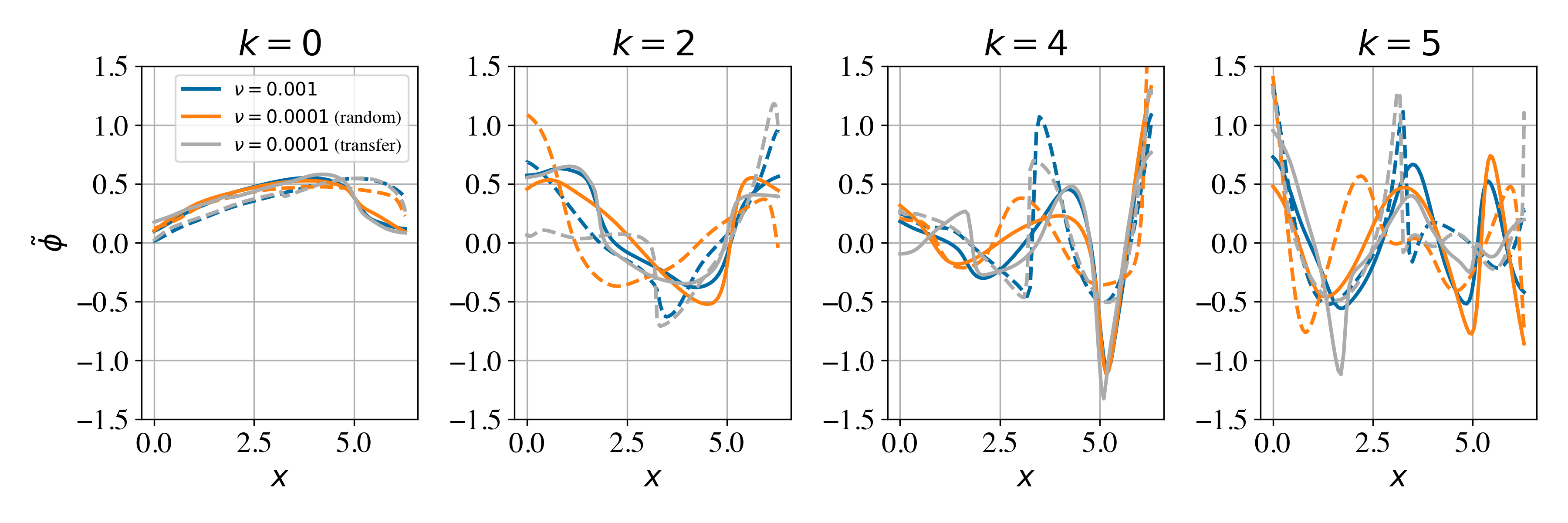}
    \caption{Custom basis functions, plotting with consistent boundaries $\Tilde{\phi}(x = 0 \text{ or } 2\pi)$ (dashed: frozen-in-time at $t=1$) for viscous Burgers on $x \in (0,2\pi)$.}
    \label{fig:viscous-burgers-1e-4-basis}
\end{figure}

Now, consider viscosity $\nu = 0.0001$. Initializing the physics-informed DeepONet with local NTK adaptive weights from the trained $\nu = 0.001$ parameters (transfer initialization) results in an improved prediction compared to random initialization. Shown in Table \ref{tab:viscous-burgers-average-errors-1e-4}, the average relative $\ell_2$ error from a physics-informed DeepONet initialized randomly for $\nu = 0.0001$ is 3.15\%. When initializing the physics-informed DeepONet for solving viscous Burgers with $\nu = 0.0001$ with the trained parameters from $\nu = 0.001$, this error decreases to 2.24\%. Figure \ref{fig:viscous-burgers-1e-4-B-a-k} shows the decay of the singular values and the expansion coefficients for the physics-informed DeepONets. The expansion coefficients decay to machine precision, meaning that they are training effectively. However, the transfer model utilizes more modes to represent the solution space, compared to the randomly initialized model that has a lower rank than the less difficult problem of $\nu = 0.001$, meaning that it is not including all the necessary modes. A sample of the custom basis functions is shown in Figure \ref{fig:viscous-burgers-1e-4-basis}, ``frozen-in-time'' at both $t=0$ and $t=1$. Qualitatively, the solution fields for viscous Burgers with $\nu = 0.001$ and $\nu = 0.0001$ look similar, which is reflected in the basis functions, particularly in capturing the location of the shock. The parameters are given a ``warm start'' in moving toward basis functions that are more representative of the solution space, rather than starting from scratch with a random initialization. The first 10 custom basis functions can be found in Appendix \ref{sec:burgers-basis-1e-4-2pi}.

\begin{table}[!htb]
    \centering
    \begin{tabular}{|c|c|c|}\hline
         Initialization & Average relative $\ell_2$ error \\ \hline\hline
    Random (NTK) & 3.15\% $\pm$ 2.35\% \\ \hline
    Transfer (NTK) from $\nu = 0.001$ & 2.24\% $\pm$ 1.63\% \\ \hline
    \end{tabular}
    \caption{Average errors across 5 test samples for viscous Burgers with $\nu = 0.0001$.}
    \label{tab:viscous-burgers-average-errors-1e-4}
\end{table}

\section{Basis functions}

\subsection{Advection-diffusion}
\label{sec:adv-diff-basis}

\begin{figure}[H]
    \centering
    \includegraphics[width=0.7\textwidth]{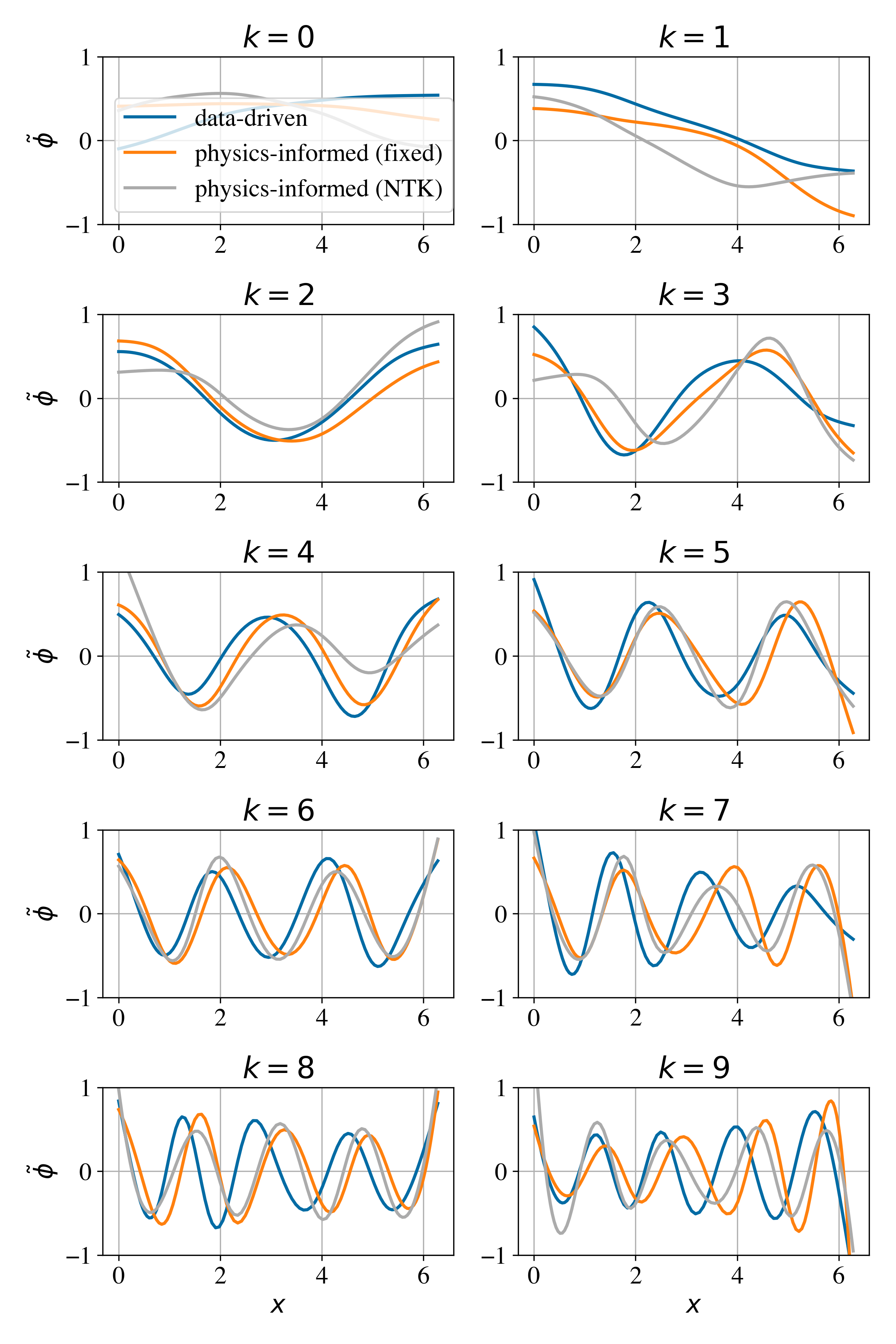}
    \caption{First 10 custom basis functions, plotting with consistent boundaries $\Tilde{\phi}(x = 0 \text{ or } 2\pi)$ for advection-diffusion.}
\end{figure}

\subsection{Viscous Burgers}

\subsubsection{$\nu=0.1$ on $x \in (0,2\pi)$}
\label{sec:burgers-basis-1e-1-2pi}

\begin{figure}[H]
    \centering
    \includegraphics[width=0.7\textwidth]{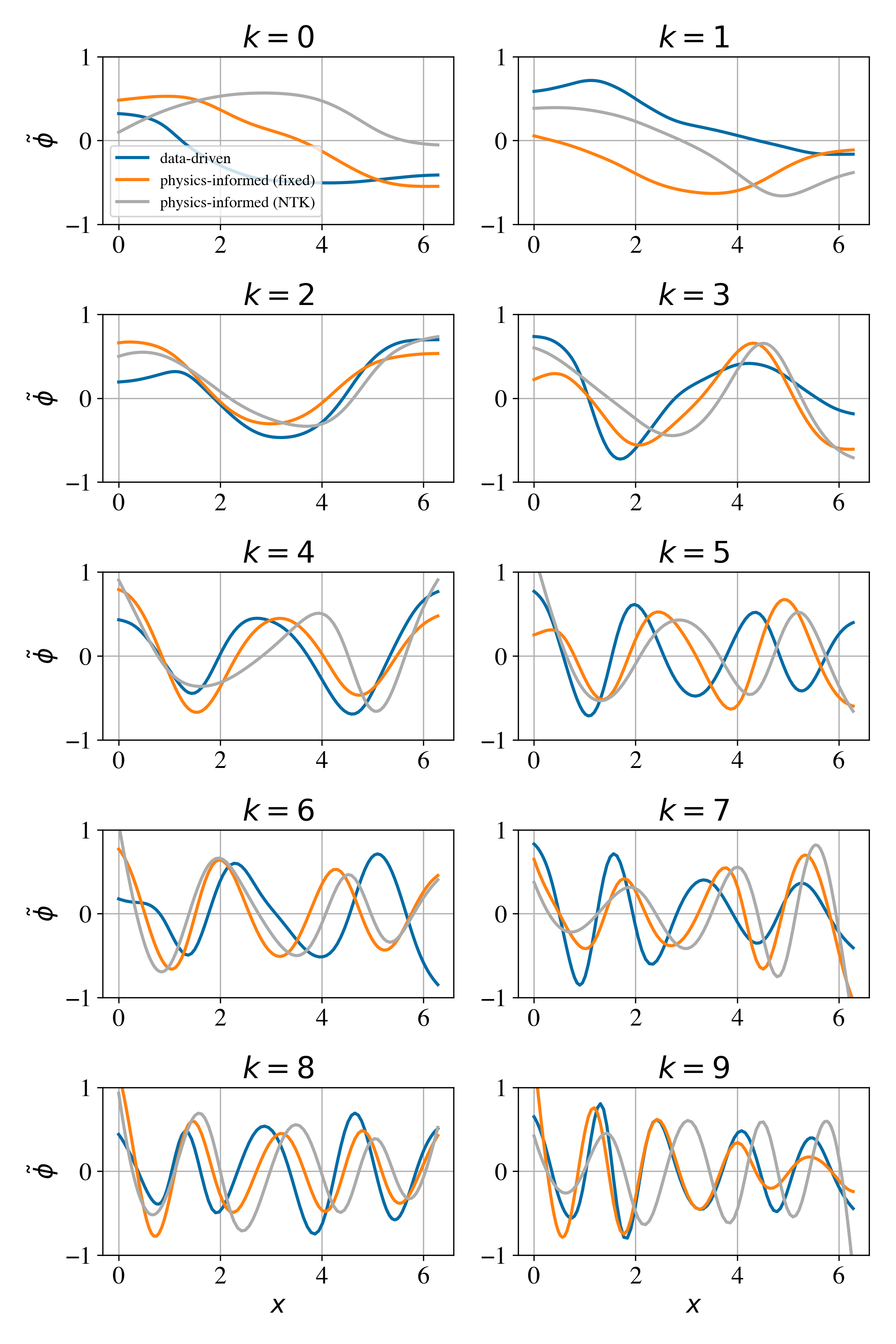}
    \caption{First 10 custom basis functions, plotting with consistent boundaries $\Tilde{\phi}(x = 0 \text{ or } 2\pi)$ for viscous Burgers with $\nu = 0.1$.}
\end{figure}

\subsubsection{$\nu=0.001$ and $\nu=0.0001$ on $x \in (0,2\pi)$}
\label{sec:burgers-basis-1e-4-2pi}

\begin{figure}[H]
    \centering
    \includegraphics[width=0.7\textwidth]{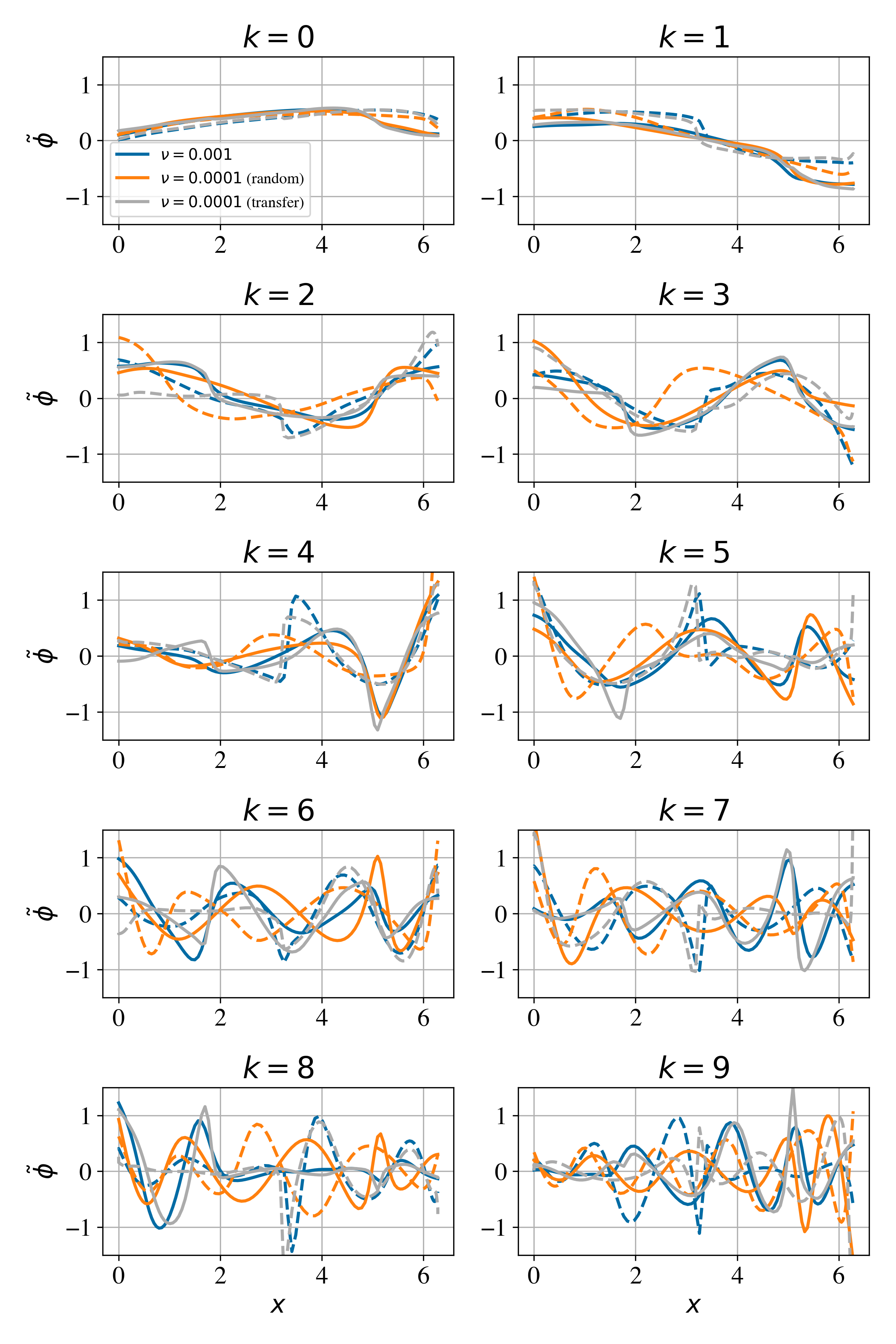}
    \caption{First 10 custom basis functions, plotting with consistent boundaries $\Tilde{\phi}(x = 0 \text{ or } 2\pi)$ (dashed: ``frozen-in-time'' at $t = 1$) for viscous Burgers with $\nu = 0.001$ and $\nu = 0.0001$.}
\end{figure}

\subsubsection{$\nu=0.001$ and $\nu=0.0001$ on $x \in (0,1)$}
\label{sec:burgers-basis-1e-4}

\begin{figure}[H]
    \centering
    \includegraphics[width=0.7\textwidth]{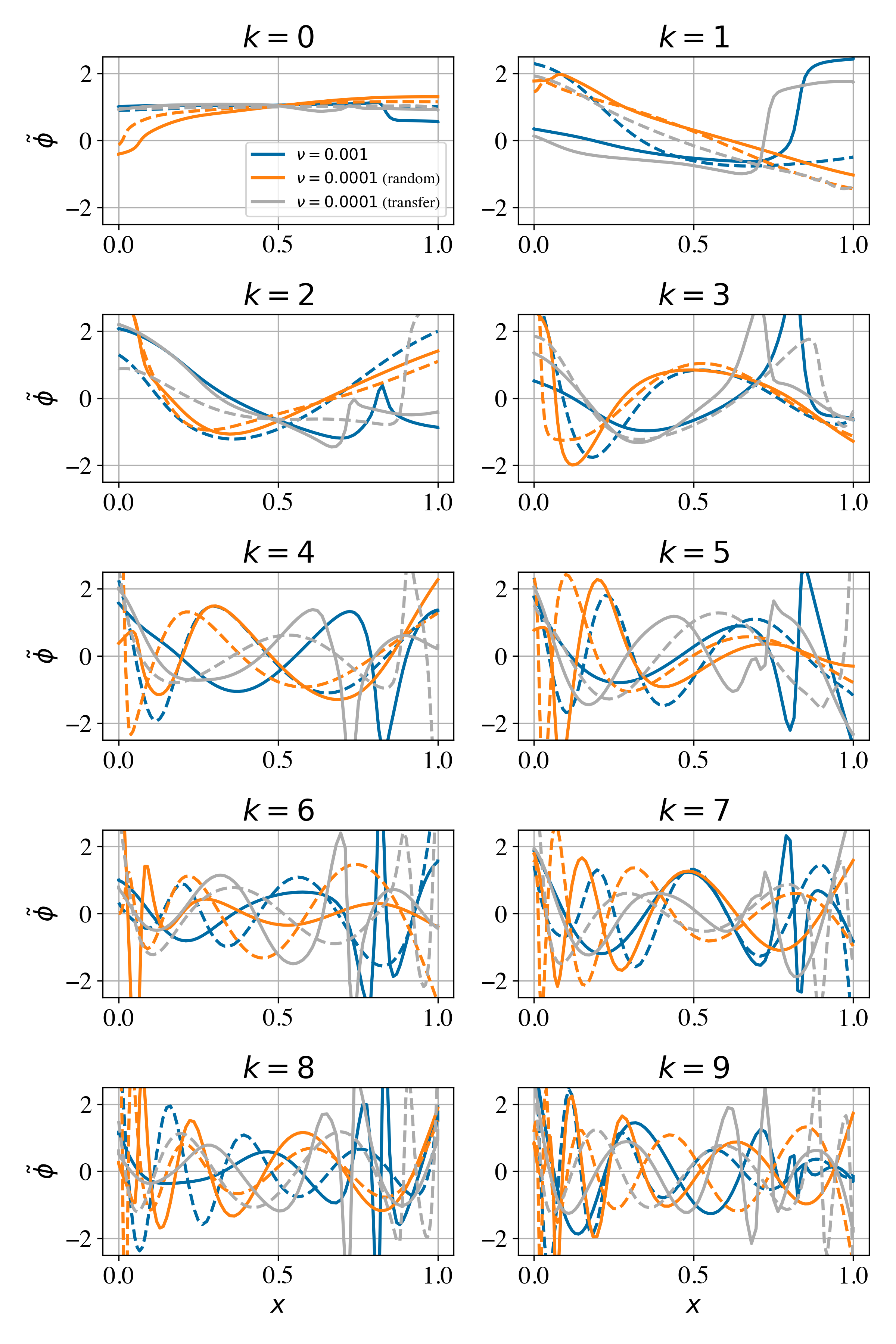}
    \caption{First 10 custom basis functions, plotting with consistent boundaries $\Tilde{\phi}(x = 0 \text{ or } 1)$ (dashed: ``frozen-in-time'' at $t = 1$) for viscous Burgers with $\nu = 0.001$ and $\nu = 0.0001$.}
\end{figure}

\subsection{Korteweg-de Vries}
\label{sec:kdv-basis}

\subsubsection{Random initialization}

\begin{figure}[H]
    \centering
    \includegraphics[width=0.7\textwidth]{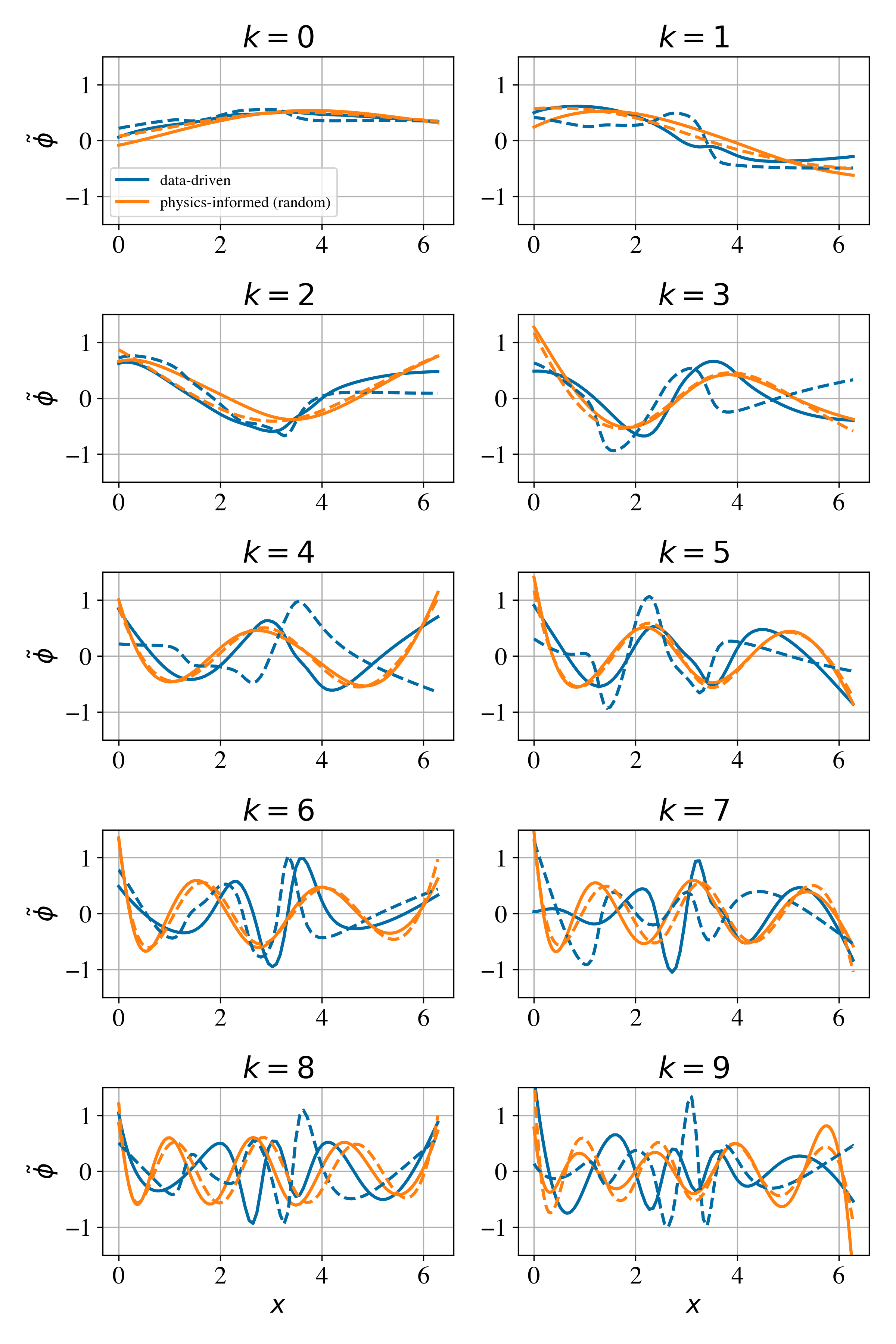}
    \caption{First 10 custom basis functions from random initialization, plotting with consistent boundaries $\Tilde{\phi}(x = 0 \text{ or } 2\pi)$ (dashed: ``frozen-in-time'' at $t = 1$) for Korteweg-de Vries.}
\end{figure}

\subsubsection{Transfer initialization}

\begin{figure}[H]
    \centering
    \includegraphics[width=0.7\textwidth]{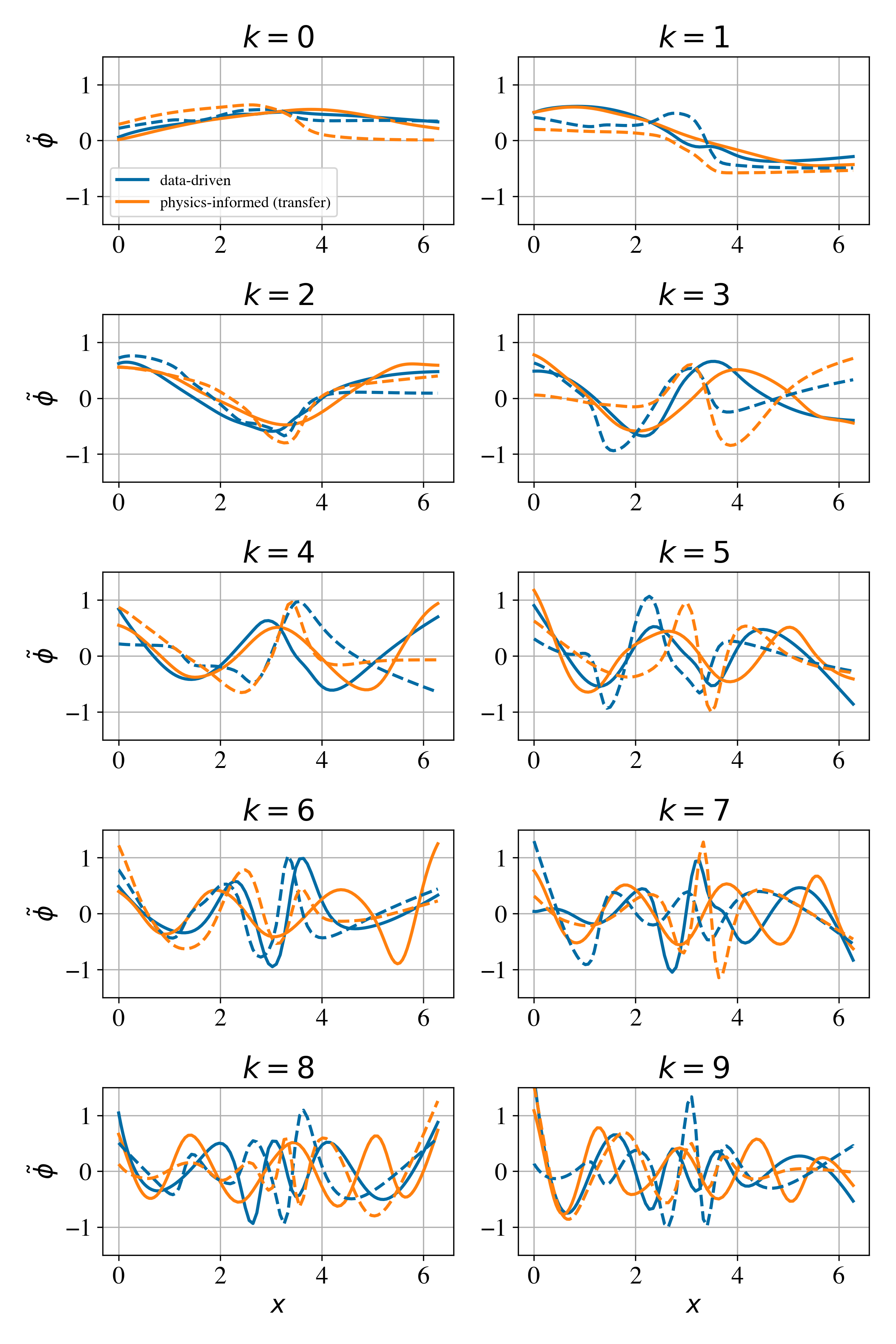}
    \caption{First 10 custom basis functions from transfer initialization with viscous Burgers, plotting with consistent boundaries $\Tilde{\phi}(x = 0 \text{ or } 2\pi)$ (dashed: ``frozen-in-time'' at $t = 1$) for Korteweg-de Vries.}
\end{figure}

\section{Training cost}
\label{sec:cost}

Computational costs for training physics-informed DeepONets with moderate local adaptive weights for the viscous Burgers benchmark case with $\nu = 0.001$ from Section \ref{sec:burgers} are included in Table \ref{tab:cost}. All models were trained on 1 NVIDIA P100 GPU on the PNNL Institutional Computing Cluster (Marianas). Training times are averaged over 5 independent runs.

\begin{table}[!htb]
    \centering
    \begin{tabular}{|c|c|c|}\hline
          & Training time & Average relative $\ell_2$ error \\ \hline\hline
        Neural tangent kernel (NTK) & 7 hr 56 min & 4.48\% $\pm$ 6.12\% \\ \hline
        Conjugate kernel (CK) & 3 hr 11 min & 7.56\% $\pm$ 7.63\% \\ \hline
    \end{tabular}
    \caption{Training costs for physics-informed DeepONet for viscous Burgers with $\nu = 0.001$.}
    \label{tab:cost}
\end{table}

\section{Error definitions}
\label{sec:error}

In the error computations, $\hat{s}$ is the approximation and $s$ is the ``ground truth'' solution.

\subsection{Relative $\ell_2$ error}
\begin{align}
    E(t_n) = \frac{||\hat{s}(x,t_n) - s(x,t_n)||_2}{||s(x,t_n)||_2}
\end{align}

\subsection{Average error}
\begin{align}
    e = \frac{1}{T} \int_0^T E(t) \,dt
\end{align}
where the integral is approximated using the trapezoidal rule \cite{meuris_machine-learning-based_2023}.

\newpage
\printbibliography

\end{document}